\documentclass[10pt,journal,compsoc]{IEEEtran}
\usepackage{amsmath,amsfonts,amssymb}
\usepackage{algorithmic}
\usepackage{algorithm}
\usepackage{array}
\usepackage{bbm}
\usepackage[caption=false,font=normalsize,labelfont=sf,textfont=sf]{subfig}
\usepackage{textcomp}
\usepackage{stfloats}
\usepackage{url}
\usepackage{verbatim}
\usepackage{graphicx}
\usepackage{booktabs}
\usepackage{color}
\usepackage[table]{xcolor}
\usepackage{diagbox}
\usepackage{multirow}
\usepackage{cite}
\usepackage{flushend}
\usepackage{threeparttable}
\usepackage[pdfstartview=FitH,
            CJKbookmarks=true,
            bookmarksnumbered=true,
            bookmarksopen=true,
            colorlinks,
            linkcolor=red,
            anchorcolor=blue,
            citecolor=green
            ]{hyperref}
\newcounter{RNum}
\renewcommand{\theRNum}{\arabic{RNum}}
\newcommand{\Remark}{\noindent\textit{\textbf{Remark}~\refstepcounter{RNum}\textbf{\theRNum}}: }
\begin{document}
\title{Siamese Object Tracking for Unmanned Aerial Vehicle: A Review and Comprehensive Analysis}

\author{ \vskip 1em
 Changhong Fu\textsuperscript{1,}*,
 Kunhan Lu\textsuperscript{1},
 Guangze Zheng\textsuperscript{1},
 Junjie Ye\textsuperscript{1},
 Ziang Cao\textsuperscript{2},
 Bowen Li\textsuperscript{1},
 and Geng Lu\textsuperscript{3}
\IEEEcompsocitemizethanks{
\IEEEcompsocthanksitem
* Corresponding author

\IEEEcompsocthanksitem
\textsuperscript{1} C. Fu, K. Lu, G. Zheng, J. Ye, and B. Li are with the School of Mechanical Engineering, Tongji University, Shanghai 201804, China.
\itshape{Email: changhongfu@tongji.edu.cn}

\IEEEcompsocthanksitem
\textsuperscript{2} Z. Cao is with the School of Automotive Studies, Tongji University, Shanghai 201804, China.

\IEEEcompsocthanksitem
\textsuperscript{3} G. Lu is with the Department of Automation, Tsinghua University, Beijing 100084, China.}
}

\markboth{Journal of \LaTeX\ Class Files, VOL. XX, NO. X, MONTH XXXX}%
{Shell \MakeLowercase{\textit{et al.}}: Bare Demo of IEEEtran.cls for Computer Society Journals}

\IEEEtitleabstractindextext{%
\begin{abstract}
Unmanned aerial vehicle (UAV)-based visual object tracking has enabled a wide range of applications and attracted increasing attention in the field of intelligent transportation systems because of its versatility and effectiveness.
As an emerging force in the revolutionary trend of deep learning, Siamese networks shine in UAV-based object tracking with their promising balance of accuracy, robustness, and speed.
Thanks to the development of embedded processors and the gradual optimization of deep neural networks, Siamese trackers receive extensive research and realize preliminary combinations with UAVs.
However, due to the UAV's limited onboard computational resources and the complex real-world circumstances, aerial tracking with Siamese networks still faces severe obstacles in many aspects.
To further explore the deployment of Siamese networks in UAV-based tracking, this work presents a comprehensive review of leading-edge Siamese trackers, along with an exhaustive UAV-specific analysis based on the evaluation using a typical UAV onboard processor.
Then, the onboard tests are conducted to validate the feasibility and efficacy of representative Siamese trackers in real-world UAV deployment.
Furthermore, to better promote the development of the tracking community, this work analyzes the limitations of existing Siamese trackers and conducts additional experiments represented by low-illumination evaluations.
In the end, prospects for the development of Siamese tracking for UAV-based intelligent transportation systems are deeply discussed.
The unified framework of leading-edge Siamese trackers, \emph{i.e.}, code library, and the results of their experimental evaluations are available at \url{https://github.com/vision4robotics/SiameseTracking4UAV}.
\end{abstract}

}

\maketitle

\IEEEdisplaynontitleabstractindextext

\IEEEpeerreviewmaketitle

\section{Introduction}\label{sec:Introduction}

\IEEEPARstart{V}{isual} object tracking has evolved as one of the most promising domains in intelligent transportation systems, especially in unmanned aerial vehicles (UAVs)~\cite{Fu2022CorrelationFF,Muffert2013ASB,Tian2018TrackingOW,Liu2017VisualOT,Chen2018RealtimeOT,Zheng2018RobustAL,Lin2021RecfER,Liu2021MultistreamSA,Li2022LearningAD,Fu2020ObjectSD,Fu2021DisruptorAI}.
Given an initial state in the first aerial image frame, UAV-based object tracking aims to infer and anticipate the location and scale of an arbitrary object in consecutive aerial image frames.
Benefiting from the small volume, flexible mobility, and convenient operation, UAVs can better implement object tracking applications, especially in intelligent transportation systems, such as road tracking~\cite{Zhou2015EfficientRD}, target recognition~\cite{Wang2020DevelopmentOU}, traffic
surveillance~\cite{Ke2017RealtimeBT}, security patrolling~\cite{Li2016CornerDB}, visual localization~\cite{Ye2022MultiRC}, and so forth.
Recently, following the introduction of deep learning (DL) in visual object tracking, there have been more breakthroughs for UAVs~\cite{Fu2021OnboardRA,Fu2021SiameseAP,Cao2021SiamAPNSA}.

A great variety of state-of-the-art (SOTA) tracking approaches have emerged in recent years, which can be divided into two types: discriminative correlation filter (DCF)-based methods~\cite{Li2021ADTrackTA,Li2020AutoTrackTH,Huang2019LearningAR,Bhat2019LearningDM,Danelljan2020ProbabilisticRF,Danelljan2019ATOMAT} and convolutional neural network (CNN)-based approaches~\cite{Held2016LearningTT,Huang2017LearningPF,Valmadre2017EndtoEndRL,Tao2016SiameseIS,Bertinetto2016FullyConvolutionalSN,Li2018HighPV}.
Due to efficient calculation, the DCF-based trackers can usually achieve real-time on a single CPU\cite{Huang2019LearningAR,Li2020AutoTrackTH,Li2021ADTrackTA}.
As a result of the restricted computational capacity and power consumption of UAVs, DCF-based methods are previously employed for aerial tracking~\cite{Fu2022CorrelationFF}.
Nonetheless, because of complex optimization strategies and the use of handcrafted features, DCF-based trackers lack robustness and generalization in dynamic complex environments.
On the other hand, based on the deep CNNs~\cite{Ondraovi2021SiameseVO}, Siamese trackers achieve a good balance between accuracy and efficiency.
Since the potential of the Siamese networks for object tracking being generally observed, they are regarded as the current promising architectures and lead emerging trends~\cite{MarvastiZadeh2022DeepLF,Ondraovi2021SiameseVO}.
After the in-depth research in recent years, more efficient Siamese trackers~\cite{Fu2021SiameseAP,Cao2021SiamAPNSA} with higher performance and faster running speed have been designed.
Besides, the rapid development of UAVs and embedded processors in recent years has made lightweight GPUs viable on UAVs.
Therefore, Siamese trackers can get a significant efficiency boost by utilizing lightweight GPUs, allowing them to satisfy the real-time requirements of aerial tracking.
As a result, Siamese trackers become a remarkable option for the deployment of UAV-based tracking~\cite{Fu2021SiameseAP,Cao2021SiamAPNSA}.

Despite their competitive performance, Siamese trackers still have a large room for improvement.
In contrast to the general tracking, during the real-world flight, the complicated and changeable natural environments, as well as the aggressive pose changes of the UAV itself, bring aerial tracking a lot of challenges~\cite{Fu2022CorrelationFF}: low resolution (LR), occlusion (OCC), illumination variation (IV), viewpoint change (VC), and fast motion (FM).
When it comes to the actual deployment of UAV-based tracking, there are still various realistic challenges for Siamese trackers due to resource limitations and extreme complex environments.
Hence, in addition to the above five UAV tracking challenges, the following extra harsh and inevitable reality aspects also provide significant obstacles to UAV-based tracking:
\begin{itemize}
\item{\textbf{Limited computational resources:} 
Considering the payload and size of UAVs, carrying sufficient computation and power resources becomes a difficult problem.
To some extent, the exploitation of embedded processors, such as NVIDIA Jetson AGX Xavier\footnote{https://www.nvidia.com/en-us/autonomous-machines/\\embedded-systems/jetson-agx-xavier/ .}, which can own more computing capacity with a smaller size and less power consumption than high-performance servers, has assisted the deployment of Siamese trackers onboard UAVs.
As a commonly-used onboard processor on UAVs, an NVIDIA Jetson AGX Xavier can deliver up to 32 tera operations per second (TOPS) of AI performance.
However, compared with high-performance computers, such as NVIDIA A100\footnote{https://www.nvidia.com/en-us/data-center/a100/ .} with a peak rate of 2,496 TOPS, the amount of computing power available is still limited.
The ability to track objects in real-time is critical for performing aerial tracking tasks, hence one of the main goals for Siamese network-based UAV trackers is to achieve better computation efficiency that can approach real-time on embedded processors.
Thus, while seeking to optimize performance to overcome UAV tracking challenges, the real-time capability is essential for object tracking onboard UAV~\cite{Fu2021SiameseAP,Cao2021SiamAPNSA}.}
\item{\textbf{Low-illumination:}
The operating environment of UAVs is not only restricted to the daytime, but also includes the outdoors at night and other scenarios with insufficient illumination conditions~\cite{Li2022AllDayOT,Ye2022TrackerMN,Ye2021DarklighterLU}.
As the aim of designing the deep CNN, it is designed to mimic the human vision.
In a low-illumination environment, lighting is frequently inconsistent and insufficient, posing significant challenges for both the operator and the UAV trackers.
Furthermore, the target object is difficult to discern from the UAV perspective due to the presence of point-like light sources and intricate shadows.
Hence, accurately distinguishing the foreground and background in low-illumination settings is critical to the trackers.}
\end{itemize}

To enlighten further researches on Siamese UAV tracking, in this work, the fundamental idea of Siamese networks is expounded first.
Besides, the core framework of representative Siamese trackers are presented and summarized in detail according to their respective innovations and advantages.
Then, for a more task-oriented comparison of their performance, comprehensive and quantitative experiments are performed on a typical UAV onboard processor with six authoritative and public UAV tracking benchmarks, \emph{i.e.}, UAV123@10fps~\cite{Mueller2016ABA}, UAV20L~\cite{Mueller2016ABA}, DTB70~\cite{Li2017VisualOT}, UAVDT~\cite{Du2018TheUA}, VisDrone-SOT2020-test~\cite{Fan2020VisDroneSOT2020TV}, and UAVTrack112~\cite{Fu2021OnboardRA}.
In addition to the overall evaluation and attribute-based analysis, this work conducts onboard tests to validate the feasibility of the representative trackers in real-world UAV deployment.
In particular, the restrictions faced by UAV tracking are discussed including a low-illumination evaluation on three low-illumination UAV benchmarks, \emph{i.e.}, UAVDark135~\cite{Li2022AllDayOT}, DarkTrack2021~\cite{Ye2022TrackerMN} and NAT2021~\cite{Ye2022UnsupervisedDA}.
As a conclusion to this work, potential research directions and prospects are discussed in light of the existing problems.

The contributions of this work lie five-fold:
\begin{enumerate}
\item{Comprehensive review: This work presents a comprehensive introduction to Siamese object tracking, enumerates a range of SOTA Siamese trackers, and highlights the discussion about the application of Siamese trackers from the UAV-specific perspective.}
\item{Code library\footnote{The integrated code library is available at \url{https://github.com/vision4robotics/SiameseTracking4UAV} .}: The majority of publicly accessible Siamese trackers are merged into a single code library in this work. Furthermore, the experimental results are also supplied for convenient reference.}
\item{Experimental evaluation\footnote{The raw results of the experimental evaluation are available at \url{https://github.com/vision4robotics/SiameseTracking4UAV} .}: This work contains exhaustive evaluation experiments of Siamese trackers on six authoritative UAV benchmarks, \emph{i.e.}, UAV123@10fps~\cite{Mueller2016ABA}, UAV20L~\cite{Mueller2016ABA}, DTB70~\cite{Li2017VisualOT}, UAVDT~\cite{Du2018TheUA}, VisDrone-SOT2020-test~\cite{Fan2020VisDroneSOT2020TV}, and UAVTrack112~\cite{Fu2021OnboardRA}. This work also includes evaluations to illustrate the limitations of SOTA Siamese network-based UAV tracking approaches.}
\item{Onboard tests: This work deploys representative Siamese trackers on a typical UAV onboard processor, \emph{i.e.}, an NVIDIA Jetson AGX Xavier, to realize real-world UAV tracking tests, where their real-time capabilities and robustness in severe realistic environments are verified.}
\item{Prospect directions: This work analyzes the existing problems and discusses the potential improvement directions of Siamese trackers. Considering the depolyment of Siamese trackers onboard UAVs, the directions of tracker design and optimization are clarified, \emph{e.g.}, the trade-off between the processing speed and performance against UAV tracking challenges, the feasible solutions to deal with adverse environments.}
\end{enumerate}

\Remark To the best of our knowledge, this work is the first to extensively compare the performance of Siamese trackers on the typical UAV onboard processor, aiming to achieve comprehensive Siamese object tracking analysis for the UAV-based intelligent transportation systems.
In addition, it is also the first to qualitatively and quantitatively compare SOTA Siamese trackers for UAV-based intelligent transportation systems.
Moreover, with the results drawn from the aforementioned summaries and comparisons, this work first thoroughly analyzes the existing problems and discusses the potential improvement directions of promising Siamese trackers.

\section{Related Works}\label{sec:Related Works}
In an early survey, A. Yilmaz \emph{et al.}~\cite{Yilmaz2006ObjectTA} detailedly categorized the SOTA tracking methods at that time, and summarized them based on the use of object representations.
They underlined the relationship between object detection and object tracking as well as analyzed the tracking issues involving the utilization of relevant image features, motion model selection, and object detection.
In other surveys~\cite{Smeulders2014VisualTA,Yang2011RecentAA,Li2013ASO}, a large scale of trackers appearing before DL has been introduced and examined.
Compared with DL-based trackers, these traditional methods may lack robustness in facing complex and diverse tracking conditions.
Although nowadays, these early tracking strategies tend to be outdated and inapplicable, these previous works have undeniably inspired subsequent research to some extent.

This section contains a large number of relevant literatures in the field of visual object tracking.
Before getting into specific Siamese tracking surveys, this section first covers some recent works on DL-based tracking methods.
A variety of surveys on the topic of UAV-based tracking are also presented.
Finally, embedded processors suitable for Siamese UAV tracking are introduced.

\subsection{DL-based tracking}\label{sec:DL-based tracking}
The emergence of DL has brought a leap forward in the field of visual object tracking~\cite{Krebs2017ASO}.
DL-based trackers show better robustness, especially for real-world scenarios.
Based on DL, the CNNs shine in tracking and their performance has improved much further.
\cite{Fiaz2018HandcraftedAD,Li2018DeepVT} classify and analyze these tracking methods, highlighting the advantages of DL-based trackers.
From their experimental results, the great prospects of CNNs in the field of visual object tracking have been noticed.
Moreover, some works have been devoted to the problems caused by the limitation of training data.
In order to exploit enormous datasets properly, S. Zagoruyko and N. Komodakis~\cite{Zagoruyko2017DeepCA} propose an approach to learn a generic similarity function for patches and analyses several neural network architectures.
The conclusion emphasizes the conspicuous performance of Siamese network-based architectures, demonstrating that they are exceptionally powerful, offering a large performance improvement, and confirming the value of multi-resolution information when comparing patches.
For most DL-based online trackers, due to continuous online learning, the calculation speed is greatly affected, and it is difficult to achieve promising real-time performance~\cite{Bhat2019LearningDM,Danelljan2020ProbabilisticRF,Danelljan2019ATOMAT}.
In~\cite{Abbass2020ASO}, experiments and analyses on trackers' performance show that online learning is difficult to balance accuracy and speed.
Recently, S. M. Marvasti-Zadeh \emph{et al.}~\cite{MarvastiZadeh2022DeepLF} comprehensively list many different types of trackers in the past decade and roughly classify them into two main categories based on the application of DL.
Their work has constructed a complete taxonomy of these trackers based on their respective unique attributes and a range of extensive experimental evaluations on numerous benchmarks.
In their survey, some problems faced by trackers have been raised, and corresponding solutions have been proposed.
Although~\cite{MarvastiZadeh2022DeepLF} recognizes Siamese trackers' enormous potential and holistically generalizes them, there are relatively few detailed analyses and extensive experimental evaluations of Siamese trackers, particularly from the UAV-specific perspective.

\subsection{Siamese tracking}\label{sec:Siamese tracking}
Siamese models are first applied to signature verification~\cite{Bromley1993SignatureVU}, and has since been gradually extended to visual detection and tracking.
A multiple pedestrian tracking method~\cite{LealTaix2016LearningBT} proposes a discussion about the association between detection and tracking based on Siamese neural networks.
Most surveys tend to emphasize the powerful performance of CNNs and see the great potential of Siamese neural networks.
With the rise of research into Siamese neural networks, some surveys those specifically aimed at the Siamese trackers have appeared.
A previous survey~\cite{Pflugfelder2017AnIA} provides a series of analyses based on the early Siamese trackers at the time when the Siamese network was first used for visual object tracking.
Their research focuses on the outstanding performance and promising future of Siamese trackers and then presents a detailed comparison of them based on data collected from other literature.
Based on previous works, 
M. Ondra{\v s}ovi{\v c} \emph{et al.}~\cite{Ondraovi2021SiameseVO} propose a more extensive and in-depth survey concentrating on the Siamese trackers.
Their work involves most of the current Siamese trackers, analyzes them from their structure and properties, and then discusses the existing problems based on the experimental results identified from other papers.
S. Javed \emph{et al.}~\cite{Javed2021VisualOT} investigate and compare existing DCF and Siamese trackers.
The core formulations of DCF and Siamese trackers are analyzed respectively.
For these trackers, a variety of comparisons are conducted to evaluate tracking performance and speed based on the quantitative results from respective papers or other papers.
In addition, the characteristics and structure of these trackers are presented and explored, as well as the future development direction.
Although the Siamese trackers have been evaluated and described in the previous studies~\cite{Pflugfelder2017AnIA,Ondraovi2021SiameseVO,Javed2021VisualOT}, it is difficult to make an intuitive and fair comparison based solely on data directly gathered from various literature, since the performance of the tracker will be affected to some extent by the evaluation platform.
Furthermore, while they have evaluated the accuracy and efficiency of Siamese trackers, they have not considered the method's application to UAVs and the influence of the tracker's speed on UAV tracking deployment.
Besides, the prospects of combining trackers with UAVs have not been mentioned.

\subsection{UAV-based tracking}\label{sec:UAV-based tracking}
The use of UAVs, although, is getting more prevalent, and visual object tracking is growing more prominent, existing reviews those address the combination of UAVs with visual tracking, especially Siamese tracking, are still insufficient.
\cite{Hao2018ARO,Chen2019TheRO} present a range of discussions about object tracking based on UAV.
According to the conclusions, traditional tracking methods are being eclipsed by DCF-based and DL-based trackers due to their superior performance.
Although DL-based trackers may achieve improved accuracy and robustness, part of them is facing the challenge of limited onboard resources.
S. You \emph{et al.}~\cite{You2019ARO} provide a similar conclusion that the DL-based trackers or the trackers based on DCF with convolutional features outperform the DCF-based trackers that use handcrafted features.
Furthermore, due to the small size and low energy supply of UAVs, deploying server dedicated efficient GPUs on UAVs is difficult.
Thus, the computational speed of DL-based methods on the CPU is inherently slow.
After all, real-time is the most basic requirement for aerial tracking.
A comprehensive review~\cite{Fu2022CorrelationFF} about DCFs for UAV-based tracking summarizes the remarkable DCF-based trackers in recent years and presents a series of exhaustive experimental evaluations.
\cite{Fu2022CorrelationFF} includes a series of real-flight onboard tests that involves deploying trackers on a conventional CPU-based onboard processor to validate the trackers' real-time performance.
However, the DCF-based methods necessitate the use of handcrafted features, which not only complicates tracker design but also makes it more difficult for the tracker to be more robust owing to excessive intervention.
The DL-based UAV object detection and tracking methods in recent years are summarized by a survey~\cite{Wu2022DeepLF}.
Focusing on three typical research topics, \emph{i.e.}, static object detection, video object detection, and multiple object tracking, the development status of UAV object detection and tracking is described.
In particular, several datasets those are commonly used for UAV detection and tracking are presented.
Although there have been some survey works in the field of UAV tracking, few have been considered in combination with SOTA Siamese trackers and novel embedded high-performance GPUs are also rarely noticed.
Especially with the development of embedded devices as well as the continuous optimization and improvement of Siamese trackers, the possibility of Siamese UAV tracking is realized.

\subsection{Tracking on the embedded processor}\label{sec:Tracking on the embedded processor}
As previously mentioned, DL-based trackers offer great accuracy and robustness, but they frequently require a lot of computational capacity.
The tremendous performance of DL-based tracking methods, including Siamese networks, will be further unlocked if GPUs can be utilized for computations.
NVIDIA Jetson series, which is developed for more efficient DL on embedded systems, has a tiny size and a relatively powerful computational capacity, which is exactly what is required for UAV-based tracking deployment.
With a CPU-GPU heterogeneous architecture that allows the CPU to boot the OS and the CUDA-capable GPU to easily implement sophisticated machine-learning algorithms, NVIDIA Jetson is the most potential and the most commonly-used processor for the inference phase of machine learning~\cite{Mittal2019ASO}.
NVIDIA Jetson AGX Xavier, which was released in recent years as the most eye-catching newer product in the series, has been regarded as the most promising option for implementing CNN-based object tracking.
The deployment of the Siamese tracker on the UAVs can be accomplished since the GPU can be utilized on the UAV onboard processor~\cite{Fu2021SiameseAP,Cao2021SiamAPNSA}.
Since the embedded processors deployed on UAVs are different from the high-performance computers commonly used for evaluation, the performance of the trackers will vary on the UAV onboard processor.
Nevertheless, the performance evaluation of Siamese trackers on UAV onboard processor is still lacking.
This work utilizes a typical UAV onboard processor to investigate the performance of Siamese trackers in UAV deployment.

\vspace{1em}
Even though much research has been accomplished on Siamese trackers, little attention has been given to the potential of integrating them with UAVs and few existing reviews focus on both Siamese trackers and their UAV deployments.
Being at the forefront of emerging trends, the potential of Siamese tracking in UAV-based intelligent transportation systems is worthy of further exploration.
Therefore, this work aims to generalize Siamese trackers and consider practical applications in combination with UAVs.
This work also focuses on experimental evaluation for UAV deployment.
The unified framework of leading-edge Siamese trackers, \emph{i.e.}, code library, and the results of their experimental evaluations are made public.
This work also involves a series of onboard tests.
To better develop Siamese tracking in UAV-based intelligent transportation systems, the prospect is discussed.

\section{Siamese Trackers}\label{sec:Siamese Trackers}
With the rapid development of DL in recent years, Siamese neural network can be considered the most promising tracking method.
Actually, Siamese models are originally used for signature verification~\cite{Bromley1993SignatureVU}.
Since it was utilized in visual object tracking~\cite{Tao2016SiameseIS,Bertinetto2016FullyConvolutionalSN}, it has drawn a lot of interest and developed swiftly due to its simple and effective dual-branch structure, which fits the tracking task.
Siamese neural network calculates the similarity between giving template and search region by using a generic similarity function.
Based on CNNs, the architecture of Siamese networks is usually simple and efficient.
The simple structure means fast computation, which is crucial for the deployment of trackers on UAVs.
In feature extraction, CNNs usually have strong and robust capabilities~\cite{Krizhevsky2012ImageNetCW}.
Different from ordinary image classification or detection, which contains only one branch for feature extraction, a typical characteristic of the Siamese neural network is the Siamese backbone branches.
To get as much feature information from both the template and target as possible, two identical backbone networks have been exploited.
By sharing weights, the two branches own the same mapping capability, so that the networks can get discriminative representation from the two patches.
Therefore, the networks possess the ability to distinguish the foreground and the background via applying the similarity function, which is helpful for accuracy and robustness.

\begin{figure*}[ht]
\centering
\includegraphics[scale=0.42]{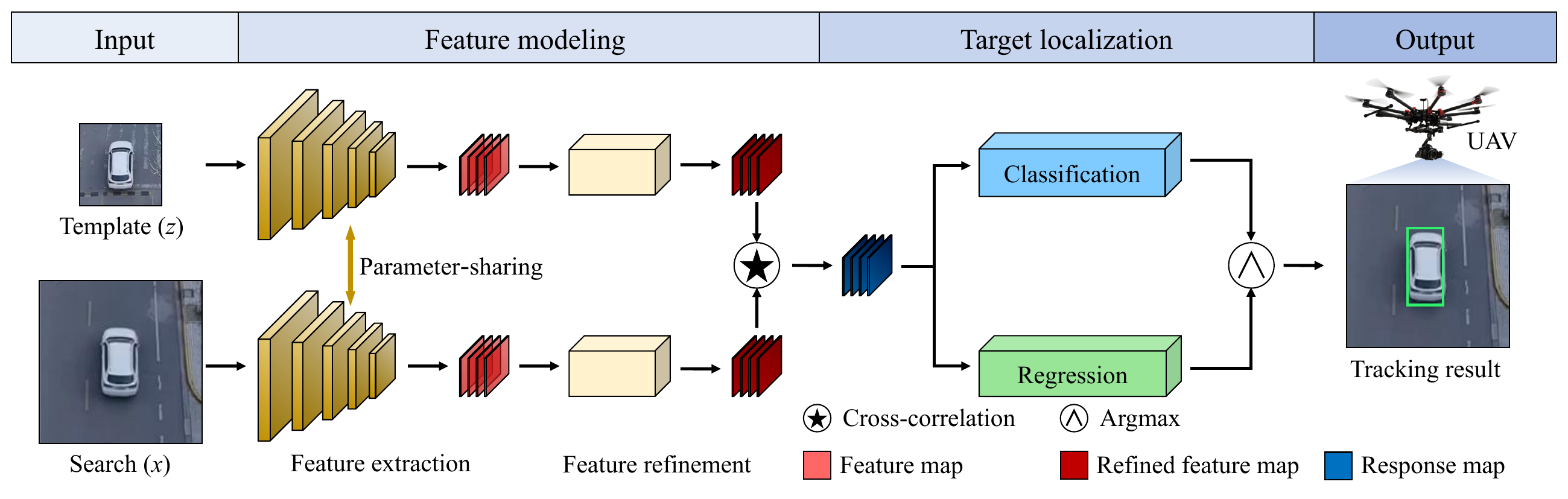}
\caption{Generic architecture of Siamese trackers.
It mainly consists of \emph{feature modeling} and \emph{target localization} stages.
(Sequence courtesy of benchmark UAVTrack112~\cite{Fu2021OnboardRA}.)
}
\label{fig:main}
\end{figure*}

\begin{figure*}[ht]
	\centering
	\includegraphics[scale=0.6]{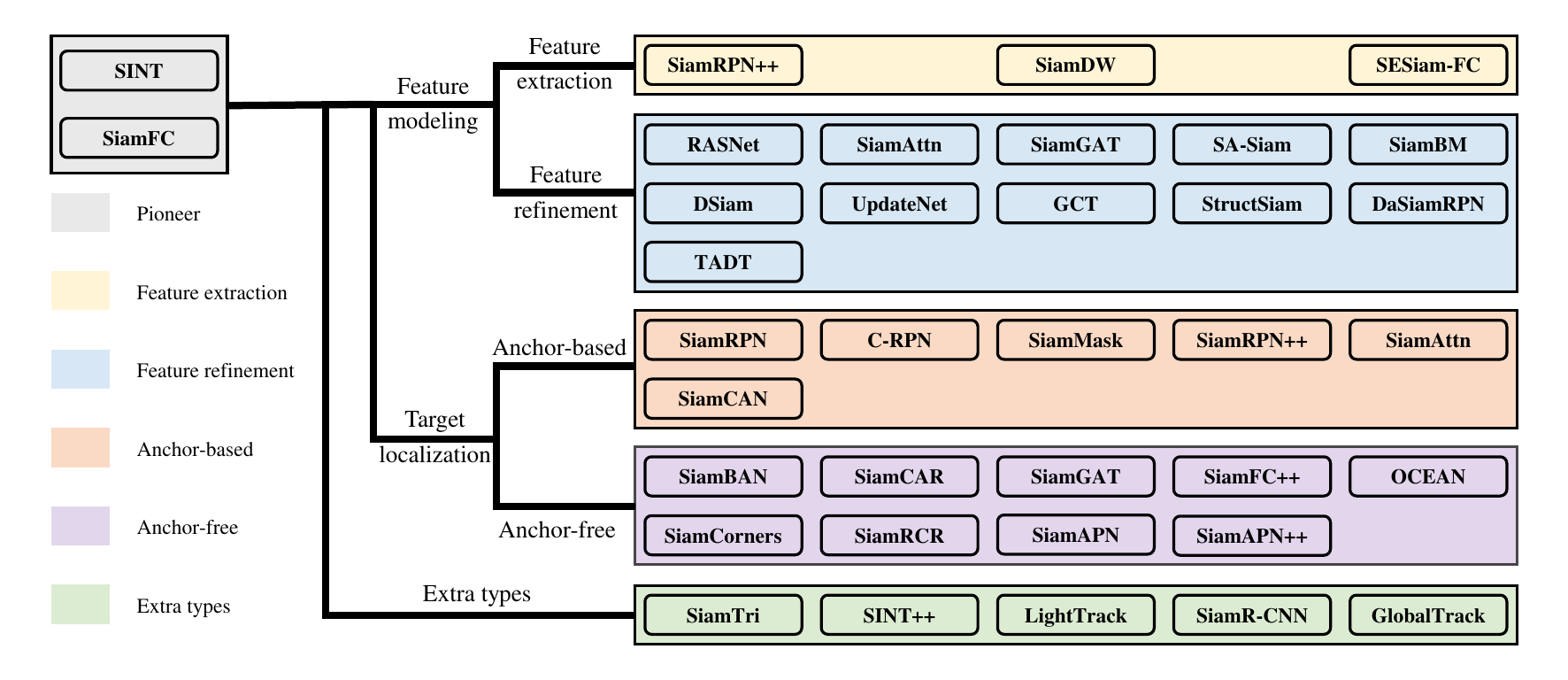}
	\caption{The category of leading-edge Siamese trackers over the years.
		They are all introduced in detail in Sec.~\ref{sec:Siamese Trackers}, based on their most remarkable contributions and innovations.
		Different colors represent different categories.
	}
	\label{fig:tree}
\end{figure*}

This work describes a generic architecture to cover leading-edge Siamese trackers according to their common characteristics, as shown in Fig.~\ref{fig:main}.
A general Siamese tracking framework includes feature modeling and target localization.
Feature modeling is the process of generating feature maps in an embedding space for both the given template and the search region images.
It can be further subdivided into two stages: feature extraction and feature refinement.
Feature extraction is performed on the input images using a parameter-sharing Siamese backbone to create feature maps; on this basis, feature refinement is used to refine the generated feature maps so that the feature information can be used more efficiently.
Subsequently, the cross-correlation operation constructs a similarity map based on the feature maps from the template and the search region to indicate the similarity.
In order to increase the accuracy of tracking results, target localization applies similarity maps to estimate the bounding box of the tracking target through classification and regression.
Target localization can be separated into two forms, \emph{i.e.}, anchor-based and anchor-free, depending on whether the pre-defined anchors are utilized for classification and regression.
As shown in Fig.~\ref{fig:tree}, all trackers are divided into corresponding subcategories based on their main contribution and innovation.
TABLE~\ref{tab:trackers} shows the SOTA Siamese trackers in their categories, including their venues, characteristics, backbones, and loss functions.

\Remark Note that some trackers may have innovations and contributions at different phases, with multiple category labels, \emph{e.g.}, SiamRPN++ has both \emph{Feature Extraction} and \emph{Anchor-Based} labels, which will be discussed respectively in the corresponding subsections.

This section comprehensively summarizes the most advanced Siamese trackers, introduces the innovations in their structures and methodologies, as well as profoundly analyzes them from the standpoint of UAV tracking deployment.
In this section, Siamese trackers are introduced starting with the pioneers who are the first to employ the Siamese structure for object tracking and have a profound influence on the follow-up.
Then, Sec.~\ref{sec:Feature Modeling} analyzes the process of feature modeling and introduces the trackers those have innovation and enhancement on feature extraction and feature refinement.
Following that, anchor-based and anchor-free trackers are introduced respectively in Sec.~\ref{sec:Target Localization}, based on different target localization approaches.
Finally, this work goes through several extra type trackers in Sec.~\ref{sec:Extra Types}.

\begin{table*}\scriptsize
\caption{Siamese trackers sorted by publication time and their characteristics.
The typical SOTA trackers listed in this table have been detailed in Sec.~\ref{sec:Siamese Trackers}.
}
\centering
\renewcommand{\arraystretch}{1.3}
\newcommand{\tabincell}[2]{\begin{tabular}{@{}#1@{}}#2\end{tabular}}
\renewcommand\arraystretch{1.2}
\begin{threeparttable}
\begin{tabular}{l l l l l l}
\toprule
\textbf{Tracker} & \textbf{Venue} & \textbf{Category} & \textbf{Characteristic} & \textbf{Backbone} & \textbf{Loss} \\
\midrule
SINT~\cite{Tao2016SiameseIS} & CVPR2016 & Pioneer & RoI pooling, margin contrastive loss & \tabincell{l}{AlexNet \\ VGG} & margin contrastive loss \\
\hline
SiamFC~\cite{Bertinetto2016FullyConvolutionalSN} & ECCV2016 & Pioneer & fully-convolutional, cross-correlation & AlexNet & logistic loss \\
\hline
DSiam~\cite{Guo2017LearningDS} & ICCV2017 & Feature refinement & \tabincell{l}{appearance variation transformation, \\ background suppression transformation, \\ element-wise multi-layer fusion} & AlexNet & logistic loss \\
\hline
SiamRPN~\cite{Li2018HighPV} & CVPR2018 & Anchor-based & RPN & AlexNet & \tabincell{l}{softmax loss, \\ cross-entropy loss, \\ smooth $L1$ loss} \\
\hline
RASNet~\cite{Wang2018LearningAR} & CVPR2018 & Feature refinement & dual attention, channel attention & MatConvNet & logistic loss \\
\hline
SA-Siam~\cite{He2018ATS} & CVPR2018 & Feature refinement & \tabincell{l}{appearance branch, semantic branch, \\ channel attention} & AlexNet & logistic loss \\
\hline
SINT++~\cite{Wang2018SINTRV} & CVPR2018 & Extra types & PSGN, HPTN & AlexNet & Bernoulli cross-entropy loss \\
\hline
DaSiamRPN~\cite{Zhu2018DistractorawareSN} & ECCV2018 & Feature refinement & distractor-aware & AlexNet & - \\
\hline
SiamTri~\cite{Dong2018TripletLI} & ECCV2018 & Extra types & triplet loss & - & triplet loss \\
\hline
StructSiam~\cite{Zhang2018StructuredSN} & ECCV2018 & Feature refinement & \tabincell{l}{local pattern detector, \\ context modeling module, \\ integration module, CRF} & AlexNet & logistic loss \\
\hline
SiamBM~\cite{He2018TowardsAB} & ECCV2018 & Feature refinement & \tabincell{l}{angle estimation, spatial masking, \\ template update} & AlexNet & logistic loss \\
\hline
SiamRPN++~\cite{Li2019SiamRPNEO} & CVPR2019 & \tabincell{l}{Feature extraction \\ Anchor-based} & \tabincell{l}{deeper backbone, \\ depthwise cross-correlation} & \tabincell{l}{AlexNet \\ ResNet \\ MobileNetV2} & \tabincell{l}{cross-entropy loss, \\ smooth $L1$ loss} \\
\hline
SiamDW~\cite{Zhang2019DeeperAW} & CVPR2019 & Feature extraction & CIR units & CIR & - \\
\hline
SiamMask~\cite{Wang2019FastOO} & CVPR2019 & Anchor-based & class-agnostic binary segmentation & ResNet & binary logistic loss \\
\hline
C-RPN~\cite{Fan2019SiameseCR} & CVPR2019 & Anchor-based & cascaded RPN & AlexNet & \tabincell{l}{cross-entropy loss, \\ smooth $L1$ loss} \\
\hline
TADT~\cite{Li2019TargetAwareDT} & CVPR2019 & Feature refinement & \tabincell{l}{target-aware, \\ ridge regression loss, ranking loss} & AlexNet & \tabincell{l}{ridge regression loss, \\ ranking loss} \\
\hline
GCT~\cite{Gao2019GraphCT} & CVPR2019 & Feature refinement & ST GCN, CT GCN & AlexNet & logistic loss \\
\hline
UpdateNet~\cite{Zhang2019LearningTM} & ICCV2019 & Feature refinement & template update & AlexNet & $L2$ loss \\
\hline
SiamAttn~\cite{Yu2020DeformableSA} & CVPR2020 & \tabincell{l}{Feature refinement \\ Anchor-based} & \tabincell{l}{DSA, self-attention, cross-attention, \\ region refinement} & ResNet & \tabincell{l}{negative log-likelihood loss, \\ smooth $L1$ loss, BCE loss} \\
\hline
SiamBAN~\cite{Chen2020SiameseBA} & CVPR2020 & Anchor-free & box adaptive & ResNet & cross-entropy loss, IoU loss \\
\hline
SiamCAR~\cite{Guo2020SiamCARSF} & CVPR2020 & Anchor-free & center-ness & ResNet & \tabincell{l}{cross-entropy loss, IoU loss, \\ center-ness loss} \\
\hline
SiamR-CNN~\cite{Voigtlaender2020SiamRV} & CVPR2020 & Extra types & \tabincell{l}{category-agnostic RPN, re-detection, \\ RoI Align, cascade, TDPA, Box2Seg, \\ hard example mining strategy} & ResNet & - \\
\hline
Ocean~\cite{Zhang2020OceanOA} & ECCV2020 & Anchor-free & \tabincell{l}{Anchor-free regression, \\ object-aware classification, \\ online update} & ResNet & BCE loss, IoU loss \\
\hline
SiamFC++~\cite{Xu2020SiamFCTR} & AAAI2020 & Anchor-free & quality assessment & \tabincell{l}{AlexNet \\ GoogLeNet} & focal loss, BCE loss, IoU loss \\
\hline
GlobalTrack~\cite{Huang2020GlobalTrackAS} & AAAI2020 & Extra types & \tabincell{l}{QG RPN, QG RCNN, \\ cross-query loss} & ResNet & \tabincell{l}{BCE loss, smooth $L1$ loss, \\ cross-query loss} \\
\hline
SiamGAT~\cite{Guo2021GraphAT} & CVPR2021 & \tabincell{l}{Feature refinement \\ Anchor-free} & graph attention module, bipartite graph & GoogLeNet & \tabincell{l}{cross-entropy loss, IoU loss, \\ center-ness loss} \\
\hline
LightTrack~\cite{Yan2021LightTrackFL} & CVPR2021 & Extra types & NAS, DSConv, MBConv & NAS & IoU loss, BCE loss \\
\hline
SiamAPN~\cite{Fu2021SiameseAP} & ICRA2021 & Anchor-free & APN, feature fusion, multi-classification & AlexNet & \tabincell{l}{$L1$ loss, smooth $L1$ loss, \\ cross-entropy loss, BCE loss, \\ IoU loss} \\
\hline
SiamAPN++~\cite{Cao2021SiamAPNSA} & IROS2021 & Anchor-free & \tabincell{l}{AAN, self-attention, cross-attention, \\ APN-DF} & AlexNet & \tabincell{l}{$L1$ loss, smooth $L1$ loss, \\ cross-entropy loss, BCE loss, \\ IoU loss} \\
\hline
SiamCAN~\cite{Zhou2021SiamCANRV} & TIP2021 & Anchor-based & \tabincell{l}{localization, global context module, \\ atrous spatial pyramid module, \\ multi-scale learnable attention module} & ResNet & cross-entropy loss, $L1$ loss \\
\hline
SiamConers~\cite{Yang2022SiamCornersSC} & TMM2021 & Anchor-free & corner pooling & ResNet & focal loss, smooth $L1$ Loss \\
\hline
SiamRCR~\cite{Peng2021SiamRCRRC} & IJCAI2021 & Anchor-free & dynamically re-weighted loss & ResNet & \tabincell{l}{focal loss, BCE loss, IoU loss, \\ dynamically re-weighted loss} \\
\hline
SE-SiamFC~\cite{Sosnovik2021ScaleEI} & WACV2021 & Feature extraction & scale-equivariant & SE-ResNet & - \\
\bottomrule
\end{tabular}
\begin{tablenotes}
\footnotesize
\item{*} - indicates that the corresponding content is not mentioned in the original paper as an emphasis.
\end{tablenotes}
\end{threeparttable}
\label{tab:trackers}
\end{table*}

\subsection{Pioneers}\label{sec:Pioneers}
SINT~\cite{Tao2016SiameseIS} is credited with being the originator of the Siamese tracking method, as it was the first to apply Siamese structure to object tracking.
Due to the far-reaching effect on follow-up tracking methods, SiamFC~\cite{Bertinetto2016FullyConvolutionalSN}, which was also established in the same period, is also recognized as the cornerstone of Siamese tracking.

\vspace{1em}
\textbf{SINT}:
\cite{Tao2016SiameseIS} presents Siamese INstance search Tracker (SINT), to investigate a more acceptable and effective matching mechanism for pairs of data.
Rather than directly modeling the matching function, SINT aims to learn it through sample videos labeled with the correct boxes.
Once the matching function is developed, the initial template is matched with search regions using the learned function to produce the most comparable forecast result.
The two-stream Siamese structure is introduced as the basic network architecture in order to better fit the matching function.
The matching function is used to track absolutely arbitrary, previously unseen targets once it has been trained.
SINT is used without model update, occlusion detection, tracker combination, or geometric matching in order to obtain higher generalization, which may facilitate UAV-based tracking in adapting to complex and changing environments.
Instead of merely identification or classification, the tracking problem is primarily a localization problem that is inherently susceptible to rough discretizations.
Taking into consideration the actual tracking scenario, the appearance of the target will not change significantly, with just minor changes occurring over time that really should be noted.
SINT decides to utilize as few max-pooling layers as possible since it introduces invariance to local deformations, as opposed to the deep CNN architecture~\cite{Krizhevsky2012ImageNetCW,Simonyan2015VeryDC}.
In addition, the region-of-interest pooling layer~\cite{Girshick2015FastR} is introduced for efficient processing of multiple regions in one frame.

\vspace{1em}
\textbf{SiamFC}:
There is no denying that the emergence of SiamFC~\cite{Bertinetto2016FullyConvolutionalSN} has a huge and profound impact on the subsequent Siamese trackers.
Similarity learning can be used to address the issue of visual object tracking.
When it comes to similarity learning, SiamFC, which uses a fully-convolutional Siamese architecture, emphasizes the importance of fully-convolutional networks, which have the advantage of being able to calculate the similarity at all translated sub-windows on a dense grid utilizing a massively larger search image instead of a search image of the same size in a single evaluation.
Not only that, to further illustrate the fully-convolutional architecture, SiamFC~\cite{Bertinetto2016FullyConvolutionalSN} introduces $T_\tau$ to denote the translation operator $(T_{\tau}x)[u] = x[u - \tau]$ and claims that for any translation $\tau$, a fully-convolutional function $h$ is one that translates signals to signals with integer stride $k$, \emph{i.e.}:
\begin{equation}
\label{eq:translation operator}
h(T_{k\tau}x) = T_{\tau}h(x)\ .
\end{equation}

After the same convolutional embedding function $\varphi$ without any paddings, the feature maps $\varphi(z)$ and $\varphi(x)$ can be obtained from the template image $z$ and the search image $x$.
Instead of a single score, a score map comprising location information defined on a finite grid can be effectively produced from the feature maps and employing the cross-correlation layer:
\begin{equation}
\label{eq:cross-correlation}
f(z, x) = \varphi(z) \star \varphi(x) + b\mathbbm{1}\ ,
\end{equation}
where $b\mathbbm{1}$ indicates a signal that takes value $b \in \mathbb{R}$ in every location; $\star$ denotes the cross-correlation operation.
The network is trained on positive and negative pairs, with the logistic loss as the loss function $L$:
\begin{equation}
\label{eq:logistic loss}
L(y, v) = {\rm log}(1 + e^{-y \cdot v})\ ,
\end{equation}
where $v$ is the real-valued score of a single exemplar-candidate pair and $y\in
\{+1, -1\}$ is its ground-truth label.
Furthermore, during the tracking stage, a cosine window is added to the score map to penalize substantial displacements, and the search region is only around four times its prior size.
Similarity learning is at the heart of Siamese tracking methods, however, early research struggled to train trackers effectively due to a lack of massive amounts of labeled datasets~\cite{Fiaz2018HandcraftedAD,Li2018DeepVT}.
In order to take full advantage of similarity learning, \cite{Bertinetto2016FullyConvolutionalSN} concentrates on the newly developed ILSVRC dataset~\cite{Russakovsky2015ImageNetLS}.
Due to its large amount of data and various sceneries and objects, it can be utilized to train a deep tracking model.
As a result, similarity learning can be more effectively applied to Siamese tracking, enabling SiamFC to achieve exceptional tracking results, even though it does not update a model or keep track of previous appearances, nor does it incorporate additional cues or refine the prediction with bounding box regression.

\Remark Noting the success of SiamFC, the enormous potential of Siamese tracking methods has received a lot of attention.
Therefore, cross-correlation and pre-training based on the ILSVRC dataset have been widely employed in the following research.
Furthermore, the advantages of Siamese trackers in terms of running speed can be observed due to the simple and effective structure, laying a solid foundation for subsequent Siamese UAV tracking.

\subsection{Feature Modeling}\label{sec:Feature Modeling}
Feature modeling aims to obtain more effective feature maps for subsequent use.
In order to recognize and utilize feature information contained in images, the backbone networks are deployed to extract features from both template and search regions.
The feature information of the images is embedded in the high-dimensional space as feature maps via the core structure of the Siamese networks, two parameter-sharing branches.
An intermediate structure will be implemented between the backbone networks and cross-correlation operation to explore the more significant information of the extracted feature maps.

\subsubsection{Feature Extraction}\label{sec:Feature Extraction}
In order to extract the features of the object, Siamese trackers typically employ an appropriate structure to extract information from the template and search images to generate feature maps.
Because of the simple and efficient construction, the CNN represented by AlexNet~\cite{Krizhevsky2012ImageNetCW} was widely employed in feature extraction in the earlier Siamese trackers~\cite{Tao2016SiameseIS,Bertinetto2016FullyConvolutionalSN,Guo2017LearningDS}.
However, as research continues deeper, the conventional shallower backbone network may not be able to provide enough information to fulfill the demands of tracking tasks.
In order to further investigate the utilization of image feature information, some trackers attempt to strengthen the backbone networks and seek more effective feature extraction methods.
Although a deeper and wider network~\cite{He2016DeepRL,Howard2017MobileNetsEC,Sandler2018MobileNetV2IR,Howard2019SearchingFM,Szegedy2016RethinkingTI}, such as ResNet, can potentially provide richer object features, the direct deployment in the Siamese structure is problematic.
SiamRPN++~\cite{Li2019SiamRPNEO} and SiamDW~\cite{Zhang2019DeeperAW} investigate the network's structure, analyse causes of the problem, and then modify the deeper and wider backbone to suit the tracking task.
In addition, SE-SiamFC~\cite{Sosnovik2021ScaleEI} focuses on scaling and improves the backbone networks to acquire more diverse feature information.

\vspace{1em}
\textbf{SiamRPN}++:
Early Siamese trackers usually lack the capacity to take advantage of deep features.
B. Li \emph{et al.}~\cite{Li2019SiamRPNEO} analyze that the core reason is the lack of strict translation invariance.
Based on this conclusion, a spatial-aware sampling strategy is proposed to break the restriction and a novel model architecture for layer-wise and depth-wise aggregations is developed.
The use of cross-correlation in the construction of the Siamese tracker entails two inherent constraints: strict translation invariance and structural symmetry.
Although AlexNet~\cite{Krizhevsky2012ImageNetCW} is extensively employed in general Siamese trackers, ResNet~\cite{He2016DeepRL}, a more sophisticated architecture, does not enhance tracking performance since it violates the strict translation invariance.
To address this issue, the conv4 and conv5 block of ResNet-50 are modified by reducing the strides and using the dilated convolutions~\cite{Shelhamer2017FullyCN}.
Besides, the ResNet-50 backbone is trained by the spatial-aware sampling strategy to avoid imposing a significant center bias on objects.
The application of the modified deep backbones facilitates the feature extraction capability of the Siamese tracker to overcome LR and the motion blur caused by FM.

\vspace{1em}
\textbf{SiamDW}:
The most common backbone networks employed in Siamese trackers, like AlexNet~\cite{Krizhevsky2012ImageNetCW}, are relatively shallow and do not fully use the powers of newer deep neural networks.
Z. Zhang \emph{et al.}~\cite{Zhang2019DeeperAW} propose unique residual modules to remove the detrimental effects of padding and develop novel architectures with controllable receptive field size and network stride utilizing these modules.
A range of experiments is carried out to get quantitative and qualitative analyses.
As a result, four guidelines for designing a Siamese network architecture have emerged:
\vspace{-0.5em}
\begin{itemize}
\item{Siamese trackers prefer a rather short network stride.}
\item{The setting of the receptive field of output features should be closely related to the exemplar image size.}
\item{When building a network architecture, the stride, receptive field, and output feature size should all be taken into account.}
\item{The problem of perceptual inconsistency between the two network streams must be addressed in a fully-convolutional Siamese matching network.}
\end{itemize}

\vspace{-0.5em}
Based on these four guidelines, cropping-inside residual (CIR) units are meant to reduce the underlying position bias.
The architecture of the residual unit and the downsampling residual unit in ResNet~\cite{He2016DeepRL} are modified as the CIR and the Downsampling CIR (CIR-D) to reduce the negative impacts caused by padding.
In addition, similar to Inception~\cite{Szegedy2015GoingDW} and ResNeXt~\cite{Xie2017AggregatedRT}, the CIR-Inception and CIR-NeXt modules are created by expanding the CIR unit with multiple features.
Consequently, deeper backbone networks comparable to ResNet~\cite{He2016DeepRL} are built with varied network strides, receptive field sizes, and construction blocks employing CIR and CIR-D units.
On the other hand, CIR-Inception and CIR-NeXt units are used to build two types of wider backbone networks.
Deeper backbone networks can be made suitable for tracking tasks by introducing CIR units and modifying the structure.
For UAV deployment, deeper backbones should be further tuned to improve feature extraction capabilities while meeting real-time requirements at the same time.

\vspace{1em}
\textbf{SE}-\textbf{SiamFC}:
Since rotation and scaling in tracking would interfere with the object's position, I. Sosnovik \emph{et al.}~\cite{Sosnovik2021ScaleEI} establish the theory for scale-equivariant Siamese trackers and present SE-SiamFC, a scale-equivariant variant of SiamFC~\cite{Bertinetto2016FullyConvolutionalSN}.
In particular, regular structures are substituted by scale-equivariant convolutions and scale-pooling~\cite{Sosnovik2020ScaleEquivariantSN} in the backbones to record extra scale correlations between features of multiple scales.
After that, non-parametric scale-convolution is used to replace the connecting correlation.
Furthermore, an extra scale-pooling is introduced to adjust the dimension of the heatmap in order to conform to the structure of SiamFC.
Capturing object scale information is beneficial for complicated UAV-based tracking.
However, when designing a Siamese tracker for UAVs, the computational cost must be taken into account firstly.

\subsubsection{Feature Refinement}\label{sec:Feature Refinement}
Feature refinement is the process of modulating the feature maps extracted by the backbone to further mine and refine the valuable information.
RASNet~\cite{Wang2018LearningAR}, SiamAttn~\cite{Yu2020DeformableSA}, SA-Siam~\cite{He2018ATS}, and SiamGAT~\cite{Guo2021GraphAT} introduce attention mechanisms for investigating internal connections in spatial and channel dimensions; SiamBM~\cite{He2018TowardsAB}, DSiam~\cite{Guo2017LearningDS}, UpdateNet~\cite{Zhang2019LearningTM}, and GCT~\cite{Gao2019GraphCT} leverage temporal information and update templates to link current frame and previous frames; GCT~\cite{Gao2019GraphCT}, StructSiam~\cite{Zhang2018StructuredSN}, and SiamGAT~\cite{Guo2021GraphAT} explore the associations between parts to concentrate on local and global information; DaSiamRPN~\cite{Zhu2018DistractorawareSN} is oriented by distractor-aware focusing on the differentiation of distractors; TADT~\cite{Li2019TargetAwareDT} emphasizes the role of the target to construct target-aware features.

\vspace{1em}
\textbf{RASNet}:
Although offline training may effectively balance tracking precision and speed, adapting an offline-trained model to an online-tracked target remains problematic.
By reformulating the correlation filter inside a Siamese tracking framework and integrating several types of attention mechanisms, the Residual Attentional Siamese Network (RASNet)~\cite{Wang2018LearningAR} seeks to adjust the model without requiring online updates.
\cite{Wang2018LearningAR} discusses the limitations of the Siamese tracker, such as the fact that the discriminator part of the Siamese tracker learns from just one sample, the first frame, and that the joint learning of feature representation and discriminator renders the model particularly susceptible to over-fitting.
Given these constraints, the Siamese network in RASNet is reformulated with weighted cross-correlation, which extends Eq. (\ref{eq:cross-correlation}) as:
\begin{equation}
\label{eq:weighted cross-correlation}
f_{p',q'}(z, x) = \sum\limits_{i=0}^{m-1}\sum\limits_{j=0}^{n-1}\sum\limits_{c=0}^{d-1} \rho_{i,j}\beta_c\varphi_{i,j,c}(z) \star \varphi_{p'+i,q'+j,c}(x)\ ,
\end{equation}
where the template image feature map $\varphi(z) \in \mathbb{R}^{m \times n \times d}$, the search image feature map $\varphi(x) \in \mathbb{R}^{p \times q \times d}$ that both are obtained from MatConvNet~\cite{Vedaldi2015MatconvnetCN}, and the response map $f(z, x) \in \mathbb{R}^{p' \times q'}$; $p \geqslant m$, $q \geqslant n$, $p' = p - m + 1$, $q' = q - n + 1$; $\rho$ denotes the dual attention, $\beta$ represents the channel attention.
The dual attention is exploited as general attention overlaid by residual attention because the superposition of estimations can simultaneously capture both the commonalities and differences of targets, and the residual attention can reduce the computing burden by encoding the global information of targets.
Furthermore, since channel attention allows the network to process semantic features for different contexts, it is employed to retain the capacity of a deep network to adapt to the target's appearance variation.
As one of the most significant UAV tracking challenges, VC may cause the appearance variation of the UAV-tracked target over time.
With the addition of multiple attention mechanisms, the capacity of UAV trackers to adapt to target appearance variations could be improved.

\vspace{1em}
\textbf{SiamAttn}:
For most Siamese trackers, the features of the target template and search regions are calculated separately.
Y. Yu \emph{et al.}~\cite{Yu2020DeformableSA} propose Deformable Siamese Attention Networks (SiamAttn), by developing a Siamese attention mechanism that calculates deformable self-attention and cross-attention.
Following SiamRPN++~\cite{Li2019SiamRPNEO}, SiamAttn uses ResNet-50~\cite{He2016DeepRL} as its backbone and includes a deformable Siamese attention (DSA) module that improves the learned discriminative representations of the template and search images.
The DSA module is comprised of two sub-modules: self-attention module and cross-attention module.
The self-attention sub-module is concerned with two aspects: channels and special locations~\cite{Fu2019DualAN}, with channel self-attention and spatial self-attention, computed independently on both the template and search branches.
For the two Siamese branches those are relatively independent, a cross-attention sub-module is proposed in order to learn more significant contextual information encoded from another branch and collaborate more effectively.
Furthermore, a $3 \times 3$ deformable convolution~\cite{Dai2017DeformableCN} is applied to the calculated attentional features to improve the capability for managing difficult geometric transformations, resulting in the final attentional features.
The DSA module improves target-background discriminability in calculated attentional features, with the potential to allow UAV-tracked objects to be more discriminative against background and distractors in intelligent transportation missions.

\vspace{1em}
\textbf{SiamGAT}:
For most cross-correlation-based trackers, the target feature region size must be pre-fixed, which results in either conserving a lot of unfavorable background information or losing several valuable foreground information.
To solve the above issues, D. Guo \emph{et al.}~\cite{Guo2021GraphAT} propose a target-aware Siamese graph attention network (SiamGAT) that establishes a part-to-part correspondence between the target and the search region with a complete bipartite graph and utilizes the graph attention mechanism to propagate target information from the template feature to the search feature.
In addition, to accommodate the size and aspect ratio changes of different objects, a target-aware area selection mechanism is included.
Existing correlations frequently neglect part-level correspondence between the target and search regions, and the global matching method can considerably condense the target information propagating to the search feature.
To solve these problems, A complete bipartite graph is used to construct part-to-part correspondence between the target template and the search region.
Each $1 \times 1 \times c$ grid of the feature map $\varphi(\cdot)$ can be regarded as a node and all nodes can be combined to form a set $V(\cdot)$, where $c$ is the number of feature channels.
Using a complete bipartite graph~\cite{Velickovic2018GraphAN} $G = (V, E)$ the part-level relations between the target and search region can be modeled, where $V = V(z) \cup V(x)$ and $E = \{(u,v)|\forall u \in V(z),\forall v \in V(x)\}$.
Consequently, two sub-graphs of $G$ can be defined as $G(z) = (V(z), \emptyset)$ and $G(x) = (V(x), \emptyset)$.
The linear transformations are performed to the node features to adaptively develop a better representation between the nodes, and then the inner product between transformed feature vectors can be computed as the correlation score.
The aggregated representation for node $i$ can be generated and fused using the attention that passed from all nodes in $G(z)$ to the $i$-th node in $G(x)$ to obtain a more robust feature representation enhanced by the target information.
In order to achieve part-to-part information propagation, SiamGAT leverages the graph attention module (GAM) to connect the Siamese backbone and the tracking head.
Furthermore, GAM activates the target-aware region, which adapts to the various aspect ratios of the tracked object commonly encountered in VC.

\vspace{1em}
\textbf{SA}-\textbf{Siam}:
Since just a bounding box is initialized in the first frame, and the target has inevitable movements, deformations, and appearance changes in successive frames, differentiating the unknown target from the cluttered background is exceptionally difficult, posing major obstacles to tracking success and accuracy.
SA-Siam~\cite{He2018ATS} is a two-fold fully-convolutional Siamese network composed of a semantic branch and an appearance branch.
The two branches are trained individually before being merged to perform visual tracking tasks.
The appearance branch follows the SiamFC network~\cite{Bertinetto2016FullyConvolutionalSN} trained in a similarity learning problem to extract appearance features, while the semantic branch directly exploits a CNN pre-trained in the image classification task~\cite{Russakovsky2015ImageNetLS} and utilizes the features from the last two convolutional layers, \emph{i.e.}, conv4 and conv5, with a channel attention module to achieve target adaptation.
As a result, the appearance branch and semantic branch response maps can be expressed as:
\begin{equation}
\label{eq:appearance cross-correlation}
f_{\rm a}(z, x) = \varphi_{\rm a}(z) \star \varphi_{\rm a}(x)\ ,
\end{equation}
\begin{equation}
\label{eq:semantic cross-correlation}
f_{\rm s}(z^{\rm s}, x) = g(\xi\cdot\varphi_{\rm s}(z)) \star g(\varphi_{\rm s}(x))\ ,
\end{equation}
where $\xi$ denotes the channel attention weighing coefficient generated through max-pooling and multilayer perceptron (MLP), and a fusion module $g(\cdot)$ is used to adjust the semantic features in the semantic branch.
In addition to appearance information, the utilization of semantic information assists the tracker in acquiring the target's feature information more accurately under severe UAV tracking challenges.

\vspace{1em}
\textbf{SiamBM}:
The rotation of the object and background clutter are major sources of interference during the tracking process, which reduces tracking accuracy.
SiamBM~\cite{He2018TowardsAB} introduces two mechanisms, angle estimation and spatial masking, to solve this difficulty, aiming for a better match between the template and search regions.
The purpose of angle estimation is to modify the scale and angle of the target to enhance features effectively.
The features from the last two convolutional layers, conv4 and conv5, are utilized based on the structure of SA-Siam~\cite{He2018ATS}.
In addition, to avoid being distracted by cluttered context information in the background, the channel attention model in the semantic branch of SA-Siam~\cite{He2018ATS} is substituted with the spatial mask due to its superior stability.
A corresponding spatial mask is deployed when the target object's aspect ratio exceeds a preset threshold.
Moreover, template updates are performed in SiamBM.
The feature map $\varphi(z_t)$ of the $t$-th template $z_t$ is defined as:
\begin{equation}
\label{eq:BM update function1}
\varphi(z_t) = \lambda_{\rm S} \varphi(z_0) + (1-\lambda_S) \varphi(z_t^u)\ ,
\end{equation}
\begin{equation}
\label{eq:BM update function2}
\varphi(z_t^u) = \lambda_{\rm U} \varphi(x_{t-1}) + (1-\lambda_U) \varphi(z_{t-1}^u)\ ,
\end{equation}
where $\lambda_{\rm S}$ is the weight of the first frame and $\lambda_{\rm U}$ is the updating rate; $\varphi(x_{t-1})$ is the feature of the target in ($t$-1)-th frame; $\varphi(z_t^u)$ is the moving average of feature maps.
Furthermore, the introduction of low-overhead mechanisms such as angle estimation and spatial mask enhances efficiency, with the potential to make tracker deployment on UAVs easier.

\vspace{1em}
\textbf{DSiam}:
Generally, only the first frame is used as a template.
This has the advantage of assuring that the template is relatively reliable.
However, the potential changes of the object and its circumstances are basically unknown and constantly happen during the tracking process.
This inevitable reality condition necessitates the capacity of the tracker to adapt to target appearance variation, eliminate background information disturbance, and maintain real-time tracking.
To solve this problem, the dynamic Siamese network (DSiam)~\cite{Guo2017LearningDS} is proposed by Q. Guo \emph{et al.} and allows for efficient online learning of target appearance variation and background suppression from preceding images.
In order to accomplish these two aims, DSiam extends the cross-correlation operation Eq. (\ref{eq:cross-correlation}) as:
\begin{equation}
\label{eq:dynamic cross-correlation}
f_t(z, x) = (\mathbf{V}_{t-1}*\varphi(z)) \star (\mathbf{W}_{t-1}*\varphi(x_t))\ ,
\end{equation}
where $\mathbf{V}$ and $\mathbf{W}$ denote the target appearance variation transformation and the background suppression transformation; $*$ denotes circular convolution.
When calculating the response map $f_t$ at the $t$-th frame, the transformations $\mathbf{V}_{t-1}$ and $\mathbf{W}_{is}$ from ($t$-1)-th frame, which are calculated by using the regularized linear regression (RLR)~\cite{Schlkopf2001LearningWK}, are used to update the template image $z$ and the search image $x_t$.
Furthermore, via element-wise multi-layer fusion, DSiam allows spatially-variant integration and its weight can be offline learned.
During the training phase, instead of merely using picture pairs, DSiam uses labeled video sequences to jointly train, allowing the network to completely acquire the rich spatial-temporal information of moving objects and learn all parameters offline.
Based on the dynamic online strategy, making full use of the objects' spatial-temporal information is an effective method against FM.

\vspace{1em}
\textbf{UpdateNet}:
Because appearance changes are frequently significant in tracking, neglecting to update the template might result in the tracker's obvious failure.
Furthermore, a basic linear update is frequently insufficient to fulfill changing updating requirements and to generalize to all possibly emerging circumstances.
To accommodate this problem, L. Zhang \emph{et al.}~\cite{Zhang2019LearningTM} propose to learn the target template update itself and construct a CNN, \emph{i.e.}, UpdateNet.
By summarizing and analyzing the limitations of conventional updates, UpdateNet provides an adaptive update strategy.
The updated template $\widetilde{z}_i$ is computed via a learned function $\psi$: 
\begin{equation}
\label{eq:updated template function}
\widetilde{z}_i = \psi(z_0^{\rm GT},\widetilde{z}_{i-1},z_i)\ ,
\end{equation}
where $z_0^{\rm GT}$ is the initial ground-truth template, $\widetilde{z}_{i-1}$ is the most recent accumulated template, and $z_i$ is the template generated from the current frame's estimated target location.
Furthermore, a skip connection from $z_0^{\rm GT}$ is added to the output of UpdateNet to assure the template's dependability.
The Euclidean distance between the two templates is monitored by a $L2$ loss during the training stage, as follows:
\begin{equation}
\label{eq:update L2 loss}
L_2 = ||\psi(z_0^{\rm GT},\widetilde{z}_{i-1},z_i)-z_{i+1}^{\rm GT}||^2\ .
\end{equation}

\Remark Online updates in the tracker tend to be required, particularly in the case of long-term tracking.
It is a reasonable solution to enhance the target's feature representation by updating the template in a learnable method.
It should be emphasized, however, that while designing a tracker for UAV, a suitable updating mechanism should be adopted to strike a balance between performance and efficiency.

\vspace{1em}
\textbf{GCT}:
The majority of Siamese methods do not fully use spatial-temporal target appearance modeling in various contexts.
J. Gao \emph{et al.}~\cite{Gao2019GraphCT} present the Graph Convolutional Tracking (GCT) jointly incorporated two types of Graph Convolutional Networks (GCNs)~\cite{Kipf2017SemiSupervisedCW}, spatial-temporal GCN and context GCN.
The tracker's sensitivity to substantial appearance changes can typically be reduced by making proper use of the target's appearance and historical information.
As a result, the Spatial-Temporal GCN (ST-GCN) $\Psi_1(\cdot)$ is used to model the structured representation of historical target exemplars, and the ConText GCN (CT-GCN) $\Psi_2(\cdot)$ is used to learn adaptive features for target localization using the context of the current frame.
The message passing between current context information $\phi(x_t)$ and each of the historical $T$ exemplar embeddings $\phi(z_{t-T:t-1})$ can be written as:
\begin{equation}
\label{eq:GCN cross-correlation}
f(z_{t-T:t-1}, x_t) = \Psi_2(\Psi_1(\phi(z_{t-T:t-1}),\phi(x_t))) \star \phi(x_t)\ .
\end{equation}

The joint utilization of spatial-temporal information and context information can support the tracker to correctly distinguish the foreground and background in the complex UAV-based tracking, which is expected to guide the feature adaption to overcome the partial occlusion in OCC.

\vspace{1em}
\textbf{StructSiam}:
While previous trackers generally characterize the appearance of template from a global perspective, resulting in high sensitivity, a structured Siamese network (StructSiam)~\cite{Zhang2018StructuredSN} is provided that considers the template's local patterns and their structural relationships at the same time.
Three components make up the structured Siamese network: a local pattern detector, a context modeling module, and an integration module.
The local pattern detector aims to discern discriminative patterns automatically to focus on local regions of the template and capture more comprehensive information.
Furthermore, the conditional random field (CRF) approximation has been introduced in StructSiam to model the joint probability linkages among local patterns.
In contrast to the correlation in SiamFC, the integration module in StructSiam gathers the local patterns of both template and search regions into a $1 \times 1$ feature map, with each channel responding as an attribute to suggest the presence of a specific pattern independent of its location.
The final matching procedure is based on the target's final structural patterns, allowing it to overcome UAV challenges like VC, FM, and OCC, which contain significant appearance variations.

\vspace{1em}
\textbf{DaSiamRPN}:
The robustness of Siamese trackers is hampered by the fact that Siamese trackers can only distinguish foreground from non-semantic backgrounds, but semantic backgrounds are easily treated as distractors.
To overcome this problem, Z. Zhu \emph{et al.}~\cite{Zhu2018DistractorawareSN} present a distractor-aware Siamese network (DaSiamRPN).
Several strategies are included during the offline training phase to manage the distribution and enable the model to focus on the semantic distractors.
First, the training dataset is enlarged to include COCO~\cite{Lin2014MicrosoftCC}, in addition to ILSVRC~\cite{Russakovsky2015ImageNetLS} and Youtube-BB~\cite{Real2017YouTubeBoundingBoxesAL}.
Additionally, the negative pairings are made up of labeled targets from the same and distinct categories, and multiple data augmentation strategies are customized.
Furthermore, using the Non-Maximum Suppression (NMS), the proposal with the maximum ranking is accepted as the target, while the other high-scoring proposals are handled as distractors.
As a result, the proposals are re-ranked via a distractor-aware objective function.
In response to long-term tracking, DaSiamRPN introduces an iterative local-to-global selection method to re-detect the target by increasing the size of the search region iteratively when failed tracking is indicated.
Additionally, when UAV trackers encounter the full occlusion in OCC, this method can assist the trackers in checking back.

\vspace{1em}
\textbf{TADT}:
For a tracking task, a network is required to distinguish a random or, most of the time, never seen object.
However, pre-trained deep features cannot model arbitrary objects in various forms, and are difficult to distinguish them from the complicated background.
In response to this, X. Li \emph{et al.}~\cite{Li2019TargetAwareDT} develop a Target-Aware Deep Tracking (TADT) model to learn target-aware features that can distinguish targets with large appearance changes.
In addition to generic feature extraction and correlation, TADT includes a target-aware model constructed with a regression loss and a ranking loss to identify target-aware filters containing target-active and scale-sensitive information.
By using regression loss to pick filters that are more active to the target, more relevant information can be acquired.
All the samples $X(i,j)$ in an image
patch are aligned with the target center to a Gaussian label map $Y(i,j)= e^{-\frac{i^2+j^2}{2\sigma^2}}$, where $(i,j)$ is the offset against the target and $\sigma$ is the kernel width.
The ridge regression loss can be written as:
\begin{equation}
\label{eq:ridge regression loss}
L_{\rm reg} = ||Y(i,j)-W*X(i,j)||^2+\lambda||W||^2\ ,
\end{equation}
where $*$ denotes the convolution operation and $W$ is the regressor weight.
When it comes to obtaining scale-sensitive data, the issue can be addressed by identifying the training sample whose size is closest to the target.
The smooth approximated ranking loss~\cite{Li2017ImprovingPR} can be written as:
\begin{equation}
\label{eq:ranking loss}
L_{\rm rank} = {\rm log}\Big(1+\sum\limits_{(x_i,x_j)\in{\Omega}}e^{\mathcal{P}(x_i)-\mathcal{P}(x_j)}\Big)\ ,
\end{equation}
where $(x_i,x_j)$ is a pair-wise training sample in the training pairs set $\Omega$; $\mathcal{P}(\cdot)$ is the prediction model.
Benefiting from the objectness and semantic information pertaining, the generated features contain effective discriminative ability and robustness against appearance variations to address UAV tracking challenges.

\subsection{Target Localization}\label{sec:Target Localization}
Target localization is the fundamental goal of object tracking to estimate the location and scale of the target.
The extracted and refined feature maps can be used to construct the response maps, which represent the similarity between the template and the search region, via cross-correlation.
On this foundation, classification and regression are used to predict the tracking target's bounding box.
Based on their classification and regression styles, Siamese trackers can be divided into two subcategories, \emph{i.e.}, anchor-based and anchor-free.

\subsubsection{Anchor-Based Method}\label{sec:Anchor-Based Method}
\begin{figure}[ht]
\centering
\subfloat[\small Anchor-based method]{\includegraphics[width=3.5in]{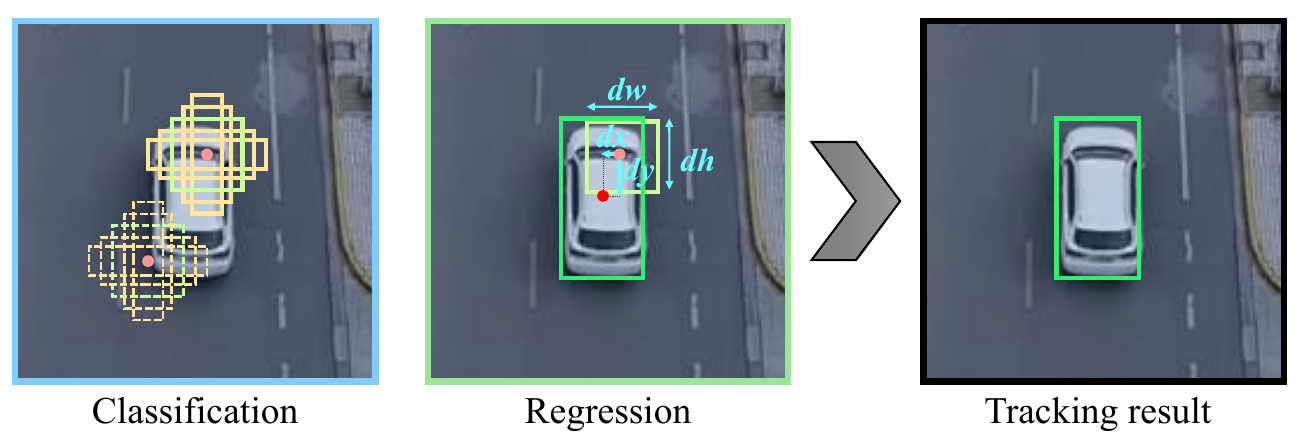}%
\label{fig:anchor-based}}
\vfil
\subfloat[\small Anchor-free method]{\includegraphics[width=3.5in]{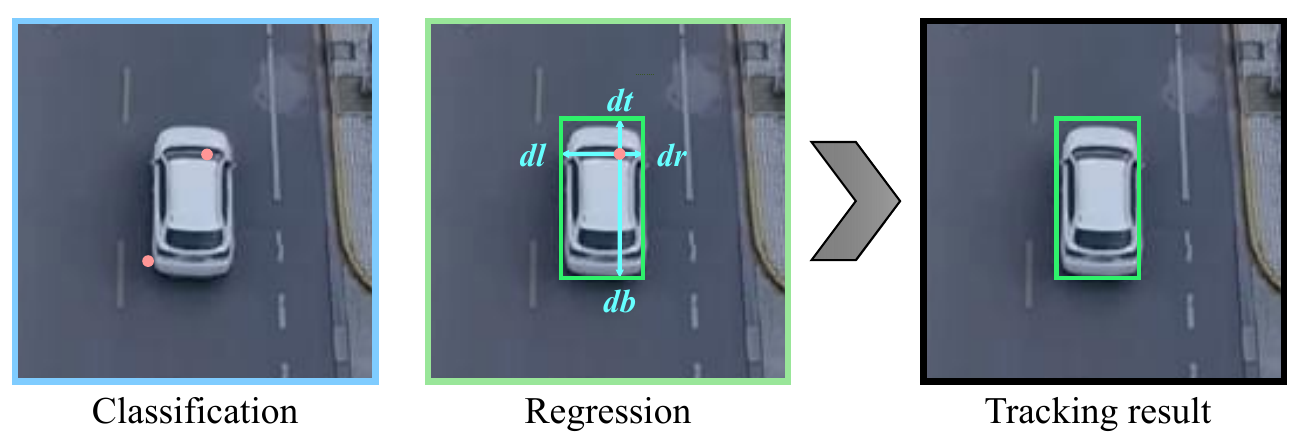}%
\label{fig:anchor-free}}
\caption{Two typical target localization methods.
(a) and (b) represent the process of anchor-based and anchor-free methods, respectively.
In (a), each box represents a pre-defined anchor for each position used to estimate the classification score; $dx$, $dy$, $dw$, and $dh$ represent the adjustment parameters of location and scale obtained by regression, and the green box represents the final result with the highest score.
In (b), each point represents a position s directly used for classification; $dt$, $db$, $dl$, and $dr$ represent the regressed offsets of each direction, and the green box represents the final result with the highest score.
(Sequence courtesy of benchmark UAVTrack112~\cite{Fu2021OnboardRA}.)
}
\label{fig:target location}
\end{figure}

Similar to detection, the anchor-based strategy is commonly employed in visual tracking to produce accurate results.
As shown in Fig.~\ref{fig:anchor-based}, based on pre-defined anchors with varying proportions and sizes at each point, the foreground and background in each anchor are estimated by classification, and regression is used to derive the adjustment parameters of these anchors.
SiamRPN~\cite{Li2018HighPV} introduces the region proposal network (RPN); C-RPN~\cite{Fan2019SiameseCR} suggests a multi-stage cascaded RPN based on RPN; SiamMask~\cite{Wang2019FastOO} leverages class-agnostic binary segmentation; SiamRPN++~\cite{Li2019SiamRPNEO} establishes multi-level RPN; SiamAttn~\cite{Dai2017DeformableCN} makes use of a region refinement module; SiamCAN~\cite{Zhou2021SiamCANRV} adds an additional localization branch.

\vspace{1em}
\textbf{SiamRPN}:
To further improve performance while maintaining desirable real-time performance, B. Li \emph{et al.}~\cite{Li2018HighPV} presents the Siamese region proposal network (SiamRPN) which adopts an end-to-end offline training strategy.
Similar to SiamFC, SiamRPN adopts a fully-convolution Siamese network to extract the features of the template image and the search image, and with shared parameters the two patches can be implicitly encoded by the same transformation.
Since standard proposal extraction strategies are time expensive and cannot satisfy the requirements of real-time processing~\cite{Uijlings2013SelectiveSF}, SiamRPN proposes deploying Region Proposal Network (RPN)~\cite{Ren2015FasterRT} integrated with the Siamese network to perform object tracking.
Based on the typical fully-convolution Siamese network, the feature maps $\varphi(\cdot)$ are extended in the channel dimension and further separated into two branches, $\varphi_{\rm cls}(\cdot)$ and $\varphi_{\rm reg}(\cdot)$, one for classification and the other for regression, via the pair-wise correlation:
\begin{equation}
\label{eq:CLS cross-correlation}
f_{\rm cls}(z, x) = \varphi_{\rm cls}(z) \star \varphi_{\rm cls}(x)\ ,
\end{equation}
\begin{equation}
\label{eq:REG cross-correlation}
f_{\rm reg}(z, x) = \varphi_{\rm reg}(z) \star \varphi_{\rm reg}(x)\ ,
\end{equation}
where $f_{\rm cls}(z, x) \in \mathbb{R}^{w \times h \times 2k}$ and $f_{\rm reg}(z, x) \in \mathbb{R}^{w \times h \times 4k}$; $k$ denotes the number of anchors.
Under the supervision of softmax loss, the classification branch aims to distinguish between negative and positive activation of each anchor at the original map's corresponding position.
Following that, the regression branch returns to the distance $dx$, $dy$, $dw$, and $dh$ to refine the location and scale of each anchor.
The classification loss is the cross-entropy loss, and smooth $L1$ loss with normalized coordinates is exploited for regression, as follows:
\begin{equation}
\label{eq:smooth L1 loss}
{\rm smooth}\ L1(x, \sigma) = \begin{cases}
0.5\sigma^2x^2, & |x|<\frac{1}{\sigma^2} \\
|x|-\frac{1}{2\sigma^2}, & |x|\geqslant\frac{1}{\sigma^2}
\end{cases}\ .
\end{equation}

Furthermore, ~\cite{Li2018HighPV} proposed to formulate the tracking problem as a local one-shot detection challenge~\cite{Bertinetto2016LearningFO}, which may further boost the speed, and presents two strategies to select the proposals.
The template branch of the Siamese feature extraction network is pruned, except for the initial frame, resulting in great efficiency.
While the target patch is being delivered into the template branch, the detection kernel is being pre-computed so that the tracking task can be performed as a one-shot detection in subsequent frames.

\Remark Based on the one-shot strategy, the template branch only generates computations when the first frame is initialized, which decreases the computational burden and enhances the efficiency of the subsequent tracking process, allowing the Siamese tracker to be further adapted to the deployment of UAVs.

Moreover, two proposal selection strategies are presented to enable the one-shot detection framework appropriate for tracking: excluding the bounding boxes that are situated too distant from the center and re-ranking the proposals using the cosine window and scale change penalty.

\vspace{1em}
\textbf{C}-\textbf{RPN}:
Since SiamRPN~\cite{Li2018HighPV} only has one-stage RPN, which cannot deal with similar distractors and large scale variance, H. Fan \emph{et al.}~\cite{Fan2019SiameseCR} propose a multi-stage tracking framework Siamese Cascaded RPN (C-RPN) to address this issue.
C-RPN consists of the Siamese network and the cascaded RPN.
The features of the template $z$ and the search regions $x$ are extracted via the Siamese network and received by the cascaded RPN for further processing.
To make better use of the multi-level features, a set of $\mathcal{L}$ RPNs are cascaded in the multi-stage tracking framework of the cascaded RPN.
In the $l$-th ($1<l\leqslant{\mathcal{L}}$) stage RPN, the features $\varphi^l(\cdot)$ of the conv-$l$ layer are fused with the high-level fused features $\phi^{l-1}(\cdot)$ from Feature Transfer Block (FTB), as follows:
\begin{equation}
\label{eq:Feature Transfer Block}
\phi^l(\cdot) = {\rm FTB}(\phi^{l-1}(\cdot), \varphi^l(\cdot))\ ,
\end{equation}
while in the first stage RPN, $\phi^1(\cdot) = \varphi^1(\cdot)$.
Therefore, Eq. (\ref{eq:CLS cross-correlation}) and Eq. (\ref{eq:REG cross-correlation}) can be reconstructed as:
\begin{equation}
\label{eq:l CLS cross-correlation}
f_{\rm cls}^l(z, x) = \phi_{\rm cls}^l(z) \star \phi_{\rm cls}^l(x)\ ,
\end{equation}
\begin{equation}
\label{eq:l REG cross-correlation}
f_{\rm reg}^l(z, x) = \phi_{\rm reg}^l(z) \star \phi_{\rm reg}^l(x)\ .
\end{equation}

The anchor set is pre-defined before the first stage, and the anchors can then be filtered and modified after each stage to obtain a more accurate result.
Due to the complexity of UAV tracking circumstances, a suitable multi-stage strategy can assure the highest possible level of tracking accuracy but should consider the impact on efficiency in process of adapting to the real-time requirements of UAV-based tracking.

\vspace{1em}
\textbf{SiamMask}:
Q. Wang \emph{et al.}~\cite{Wang2019FastOO} develop SiamMask, which combines object tracking with segmentation and performs three primary tasks: similarity learning, bounding box regression, and class-agnostic binary segmentation.
As in SiamFC~\cite{Bertinetto2016FullyConvolutionalSN}, similarity learning examines the similarity relation between template and search regions.
As in SiamRPN~\cite{Li2018HighPV}, the goal of bounding box regression is to generate more precise results.
In addition, SiamMask focuses on class-agnostic binary segmentation, which refines object information via creating binary segmentation masks.
SiamMask leverages a modified ResNet-50~\cite{He2016DeepRL} as the backbone and adds an adjustment layer without sharing parameters.
In addition, to create a multi-channel response map, the basic cross-correlation Eq. (\ref{eq:cross-correlation}) is substituted by depth-wise cross-correlation~\cite{Bertinetto2016LearningFO}.
After the critical semantic information is obtained, a two-layer neural network $h(\cdot)$ predicts a binary mask $m^n$ for each spatial element $f^n(z, x)$ using the response map $f(z, x)$ created by cross-correlation, as follows:
\begin{equation}
\label{eq:segmentation network}
m^n = h(f^n(z, x))\ .
\end{equation}

The mask prediction task's loss function $L_{\rm mask}$ is a binary logistic regression loss that is used during training.
By merging with the mask branch, SiamFC~\cite{Bertinetto2016FullyConvolutionalSN} and SiamRPN~\cite{Li2018HighPV} can be respectively derived into two variants with two branches and three branches.
Finally, three strategies are investigated through experiments to transform the binary mask into a bounding box: axis-aligned bounding rectangle (Min-max), rotated minimum bounding rectangle (MBR), and the automatic bounding box generation optimization strategy (Opt)~\cite{Kristan2016TheVO}.
The MBR strategy is adopted to obtain the box, taking into consideration resource consumption and generation accuracy.
The use of segmentation can increase the bounding box's precision and hence the tracking results, however, the impact on calculation speed should be addressed while constructing the UAV tracker.
UAV trackers' performance should be optimized while preserving real-time, or existing approaches should be tweaked to improve efficiency to match UAV-based tracking.

\vspace{1em}
\textbf{SiamRPN}++:
SiamRPN++~\cite{Li2019SiamRPNEO} can further enhance tracking performance by utilizing the extensive hierarchical information of multi-level features after ResNet~\cite{He2016DeepRL} is established for feature extraction.
The extracted multi-level features from the final three residual blocks, conv3, conv4, and conv5, are fed into three Siamese RPN module~\cite{Li2018HighPV} individually.
The overall classification and regression results then can be obtained by integrating all of the outputs via a weighted-fusion layer.
Since SiamRPN's up-channel cross-correlation (UP-Xcorr)~\cite{Li2018HighPV} causes a significant imbalance in parameter distribution, making training optimization difficult, a lightweight cross-correlation layer called depth-wise cross-correlation (DW-XCorr) is proposed to replace it.

\Remark The usage of DW-XCorr reduces calculation costs and redundant parameters, resulting in improved convergence during training.
The increased computing efficiency of the cross-correlation operation boosts tracking speed, allowing more Siamese trackers to satisfy the real-time requirements of UAV-based tracking.

\vspace{1em}
\textbf{SiamAttn}:
Based on the aforementioned Siamese attentional features outputted by the DSA module, SiamAttn~\cite{Dai2017DeformableCN} applies three Siamese RPN blocks~\cite{Li2019SiamRPNEO} to construct three prediction maps and subsequently produce a range of region proposals.
To further refine the precision of the predicted target region, a region refinement module is exploited for bounding box regression and mask prediction.
Depth-wise cross-correlations combine multiple stages of two attentional features to produce multiple correlation maps, which are then input into a fusion block and aligned in both the spatial and channel domains.
Following that, the aligned features are fused with the element-wise sum to predict the final results.
Specifically, the convolutional features from the first two stages are integrated into the fused features, and a deformable RoI pooling is utilized to more precisely calculate target features.
As classification and regression losses, a negative log-likelihood loss and a smooth $L1$ loss are used individually.
A smooth $L1$ loss and a binary cross-entropy (BCE) loss are utilized for bounding box regression and mask segmentation.
The addition of the DSA and region refinement modules enhances tracking performance significantly, albeit at the expense of running speed.
The strict requirements of UAV-based tracking on running speed necessitate that the tracker should boost performance while avoiding the loss of efficiency as much as possible.

\vspace{1em}
\textbf{SiamCAN}:
W. Zhou \emph{et al.}~\cite{Zhou2021SiamCANRV} present a Siamese center-aware network (SiamCAN), consisting of the classification and regression branches, as well as an additional localization branch different from the general anchor-based Siamese structure.
The multi-scale feature maps are extracted and cropped~\cite{Valmadre2017EndtoEndRL} from the ResNet-50~\cite{He2016DeepRL} backbone, and then fed into three branches individually.
Both the classification and regression branches are anchor-based, whereas the localization branch is anchor-free and aims to directly locate the target center to assist the regression branch in producing more precise results.
Furthermore, the global context module~\cite{Cao2019GCNetNN} is added to the localization branch to obtain long-range dependencies for increased robustness to large target displacements, and the atrous spatial pyramid module~\cite{Chen2018DeepLabSI} is designed to capture context information at multiple scales.
The depth-wise convolution operation on the multi-scale features produces the correlation feature maps in both the classification and regression branches.
The correlation feature maps are managed by the global context module and the atrous spatial pyramid module after the multi-scale features are resized and element-wise multiplied in the localization branch.
In addition, a multi-scale learnable attention module is employed to obtain the final predictions, guiding these three branches to leverage discriminative features for greater performance.
The classification and localization branches are supervised using the cross-entropy loss, while the regression branch is supervised using the $L1$ loss.
The introduction of the localization branch, in addition to the classification and regression branches, can enable to better localize the target's center and cope with FM.
Reducing network redundancy is a promising measure for adapting to the deployment of UAVs.

\subsubsection{Anchor-Free Method}\label{sec:Anchor-Free Method}
Anchor-free approaches strive to avoid the impact of pre-defined anchors to increase generalization.
As shown in Fig.~\ref{fig:anchor-free}, different from anchor-based methods, anchor-free approaches directly classify the foreground and background of each location and regress the offsets corresponding to the four directions.
SiamBAN~\cite{Chen2020SiameseBA} proposes the box adaptive network; SiamCAR~\cite{Guo2020SiamCARSF} and SiamGAT~\cite{Guo2021GraphAT} reconsider classification and regression methods; SiamFC++~\cite{Xu2020SiamFCTR} summarizes and employs a set of design guidelines; Ocean~\cite{Zhang2020OceanOA} builds the object-aware anchor-free network; SiamCorners~\cite{Yang2022SiamCornersSC} converts bounding box estimations into corner predictions; SiamRCR~\cite{Peng2021SiamRCRRC} connects classification and regression in a reciprocal way; SiamAPN~\cite{Fu2021SiameseAP} and SiamAPN++~\cite{Cao2021SiamAPNSA} construct a small number of high-quality anchors rather than a large number of pre-defined anchors before performing classification and regression.

\Remark Based on the idea of the anchor-free method, the anchor box selection strategy used by SiamAPN~\cite{Fu2021SiameseAP} and SiamAPN++~\cite{Cao2021SiamAPNSA} avoids the use of pre-defined anchors but generates a small number of high-quality anchors through a two-stage method.
The anchor proposal method facilitates better tracking results while maintaining higher efficiency, making it more suitable for UAV deployment.

\vspace{1em}
\textbf{SiamBAN}:
The multi-scale searching strategy or pre-defined anchors are used by the majority of current trackers.
To address this issue, Z. Chen \emph{et al.}~\cite{Chen2020SiameseBA} propose a Siamese box adaptive network (SiamBAN) that makes use of the expressive power of the fully-convolutional network (FCN).
The visual tracking problem can be thought of as a combination of classification and regression in a unified FCN.
ResNet-50~\cite{He2016DeepRL} is adopted as the backbone network of SiamBAN by omitting the final two convolution blocks' downsampling operations and using atrous convolution~\cite{Chen2018DeepLabSI} with adjusted atrous rates.
Box adaptive head consists of a classification module and a regression module.
The classification module outputs two channels for foreground-background classification, and the regression module outputs four channels for estimation of the bounding box.
Compared to the anchor-based trackers, the number of output variables is smaller.
Based on ResNet-50~\cite{He2016DeepRL} with atrous convolution, the multi-level features are utilized for prediction through multiple box adaptive heads, and the classification maps and the regression maps are adaptively fused.
In particular, the region for classification is labeled based on two ellipses centered on the center of the ground-truth bounding box.
The cross-entropy loss is used for classification and the Intersection over Union (IoU) loss is used for regression.
Similar to GIoU~\cite{Rezatofighi2019GeneralizedIO}, the IoU loss is described as:
\begin{equation}
\label{eq:IoU loss}
L_{\rm IoU} = 1 - IoU\ ,
\end{equation}
where $IoU$ denotes the area ratio of intersection over the union of the estimated bounding box and the ground-truth.
The tracker can better adapt to the scale variations of objects caused by VC without the need for a multi-scale searching mechanism.

\vspace{1em}
\textbf{SiamCAR}:
Decomposing the visual object tracking into two subproblems: pixel category classification and object bounding box regression at this pixel, D. Guo \emph{et al.}~\cite{Guo2020SiamCARSF} propose a fully-convolutional Siamese network named SiamCAR.
SiamCAR consists of two simple subnetworks: the Siamese subnetwork for feature extraction and the classification-regression subnetwork for bounding box prediction.
To obtain improved inference for recognition and discrimination, the features extracted from the final three residual blocks of the backbone are concatenated as a unity $\Phi(\cdot)$ for depth-wise cross-correlation~\cite{Li2019SiamRPNEO}, as follows:
\begin{equation}
\label{eq:feature concatenate}
\Phi(\cdot) = {\rm Cat}(\varphi_3(\cdot),\varphi_4(\cdot),\varphi_5(\cdot))\ ,
\end{equation}
where $\varphi_{i}(\cdot)$ denotes the output from the $i$-th residual blocks of the backbone both the template image $z$ and the search image $x$; Cat represents channel-wise concatenation.
For classification, the cross-entropy loss is utilized, while for regression, the IoU loss is employed.
Using the regression targets ${\zeta}_{(i,j)}$ to denote the the distances between the corresponding location and the bounding box's four sides, \emph{i.e.}, $\tilde{\zeta}_{(i,j)}^{0:3} = \tilde{l},\tilde{t},\tilde{r},\tilde{b}$, an indicator function $\mathbb{I}(\cdot)$ can be defined as:
\begin{equation}
\label{eq:indicator function}
\mathbb{I}(\tilde{\zeta}_{(i,j)}) = \begin{cases}
1, & \tilde{\zeta}_{(i,j)}^k>0, k=0,1,2,3 \\
0, & {\rm otherwise}
\end{cases}\ .
\end{equation}
The regression loss can be calculated as:
\begin{equation}
\label{eq:regression loss}
L_{\rm reg} = \frac{1}{\sum{\mathbb{I}(\tilde{\zeta}_{(i,j)})}}\sum\limits_{i,j}\mathbb{I}(\tilde{\zeta}_{(i,j)})L_{\rm IoU}(f_{\rm reg}(i,j,:),\tilde{\zeta}_{(x,y)})\ ,
\end{equation}
where $L_{\rm IoU}$ is the IoU loss as in~\cite{Yu2016UnitboxAA}.
In addition, a center-ness branch~\cite{Tian2019FcosFC} in parallel with the classification branch to remove the outliers caused by the locations far away from the object center.
For each point value in the $f_{\rm cen}(i,j)$ output from this branch, the center-ness score $C(i,j)$ of the corresponding location can be described as:
\begin{equation}
\label{eq:center-ness score}
C(i,j) = \tilde{\zeta}_{(i,j)}\ast{\sqrt{\frac{{\rm min}(\tilde{l},\tilde{r})}{{\rm max}(\tilde{l},\tilde{r})}\times\frac{{\rm min}(\tilde{t},\tilde{b})}{{\rm max}(\tilde{t},\tilde{b})}}}\ .
\end{equation}

The center-ness loss can be calculated as:
\begin{equation}
\label{eq:center-ness loss}
\begin{split}
L_{\rm cen} = &\frac{-1}{\sum{\mathbb{I}(\tilde{\zeta}_{(i,j)})}}\sum\limits_{\mathbb{I}(\tilde{\zeta}_{(i,j)})==1}(C(i,j)\ast{{\rm log}(f_{\rm cen}(i,j))} \\
&+(1-C(i,j))\ast{{\rm log}(1-f_{\rm cen}(i,j))})\ .
\end{split}
\end{equation}

The combined action of classification and center-ness branches improves the precision of the results' location, which may be effective in dealing with FM.

\vspace{1em}
\textbf{SiamGAT}:
Benefiting from the aforementioned GAM, SiamGAT~\cite{Guo2021GraphAT} can perform part-to-part information propagation.
GoogLeNet~\cite{Szegedy2016RethinkingTI} is utilized as the backbone of SiamGAT since it can learn multi-scale feature representations with fewer parameters and quicker processing speed.
Being connected with GAM, the classification-regression head network from SiamCAR~\cite{Guo2020SiamCARSF} is used as the tracking head, which predicts category information and computes the target bounding box for each position.
By replacing DW-XCorr with GAM featuring a target-aware mechanism, SiamGAT enhances both success and precision.
When designing trackers for UAVs, it is also a decent idea to consider developing more effective approaches to optimize or replace cross-correlation operations in order to increase performance and efficiency.

\vspace{1em}
\textbf{SiamFC}++:
Based on a meticulous analysis of current visual tracking problems, Y. Xu \emph{et al.}~\cite{Xu2020SiamFCTR} propose a set of generic object tracker design guidelines:
\vspace{-0.5em}
\begin{itemize}
\item{The tracker should perform two sub-tasks: classification and state estimation.}
\item{The classification score should directly indicate the confidence score of target existence, in the sub-window of the corresponding pixel.}
\item{Prior knowledge, such as scale/ratio distribution, should not be introduced in tracking methods.}
\item{An estimation quality score independent of classification should be applied.}
\end{itemize}

\vspace{-0.5em}
Following these four guidelines, the fully-convolutional Siamese tracker++ (SiamFC++) is designed based on SiamFC~\cite{Bertinetto2016FullyConvolutionalSN}.
Both classification head and regression head are constructed following the embedding space cross-correlation.
SiamFC++ directly classifies the corresponding image patch and regresses the bounding box of the target at the location.
Therefore, it is free of pre-defined anchor boxes.
A quality assessment branch~\cite{Tian2019FcosFC,Jiang2018AcquisitionOL} is added by appending a $1 \times 1$ convolution layer in parallel with the classification head.
The focal loss~\cite{Lin2020FocalLF} is used for classification result, the BCE loss is used for quality assessment and the IoU loss~\cite{Yu2016UnitboxAA} is used for bounding box result.
Since the UAV tracking scene is frequently complicated and changeable, which necessitates strong tracker generalization, hyperparameters and imbalanced samples should be avoided while developing the UAV tracker.

\vspace{1em}
\textbf{Ocean}:
The improvement of anchor-based Siamese trackers is limited by the delayed tracking robustness since they are just trained on the positive anchor boxes.
To address this issue, an object-aware anchor-free network (Ocean)~\cite{Zhang2020OceanOA} is proposed.
The location and scale of target objects are directly estimated in an anchor-free manner, rather than refining the reference anchor boxes.
The training samples for the anchor-free regression network are all the pixels in the ground-truth bounding box, and the distances between each pixel within the target object and the four sides of the ground-truth bounding box are estimated.
The presented feature alignment module changes the fixed sampling positions of a convolution kernel to align with the estimated bounding box to train an object-aware feature for classification.
The regression network predicts a matching object bounding box for each position in the classification map.
Both the object-aware feature and the regular-region feature are utilized for classification in the object-aware classification network because the former offers a global description of the target and the latter focuses on local regions of images.
To optimize the anchor-free networks, the IoU loss~\cite{Yu2016UnitboxAA} and the BCE loss~\cite{Boer2005ATO} are employed to train the regression and classification networks jointly.
The anchor-free regression loss $L_{\rm af\_reg}$ is defined as:
\begin{equation}
\label{eq:anchor-free regression loss}
L_{\rm af\_reg} = -\sum\limits_{i}{\rm ln}(L_{\rm IoU}(f_{\rm reg},T^*))\ ,
\end{equation}
where $f_{\rm reg}$ denotes the prediction, and $i$ indexes the training samples.
The classification loss consists of the object-aware feature loss $L_{\rm o}$ and the regular-region feature loss $L_{\rm r}$, as follow:
\begin{equation}
\label{eq:object-aware feature loss}
L_{\rm o} = -\sum\limits_{j}p_{\rm o}^*{\rm log}(f_{\rm o})+(1-p_{\rm o}^*){\rm log}(1-f_{\rm o})\ ,
\end{equation}
\begin{equation}
\label{eq:regular-region feature loss}
L_{\rm r} = -\sum\limits_{j}p_{\rm r}^*{\rm log}(f_{\rm r})+(1-p_{\rm r}^*){\rm log}(1-f_{\rm r})\ \ ,
\end{equation}
where $f_{\rm o}$ and $f_{\rm r}$ are the classification score maps for the object-aware feature and the regular-region feature, respectively, $j$ denotes the classification training samples, and $p_{\rm o}^*$ and $p_{\rm r}^*$ denote the ground-truth labels.
Specifically, $p_{\rm o}^*$ is a probabilistic label representing the IoU between the predicted bounding box and ground-truth, and $p_{\rm r}^*$ is a binary label formulated as:
\begin{equation}
\label{eq:regular-region feature label}
p_{\rm r}^*[v] = \begin{cases}
1, & ||v-c||\leqslant{R} \\
0, & {\rm otherwise}
\end{cases}\ .
\end{equation}

The tracking algorithm building upon Ocean~\cite{Zhang2020OceanOA} contains two parts: an offline anchor-free model and an online update model.
The object-aware anchor-free networks are utilized for offline tracking, and it consists of three stages: feature extraction, combination, and target localization.
With the last stage cut off and the fourth stage adjusted, a modified ResNet-50~\cite{He2016DeepRL} is utilized as the backbone of Ocean for feature extraction.
The extracted features are then combined via a depth-wise cross-correlation~\cite{Li2019SiamRPNEO} to generate the corresponding similarity features.
As there is only a single-scale feature extracted from the last stage of the backbone, it is processed through three parallel dilated convolution layers~\cite{Zhang2018ContextEF} and fused using point-wise summation.
After this, the object-aware anchor-free networks are utilized to localize the target.
Furthermore, an online update model is equipped with the offline algorithm to capture the appearance changes of the target during tracking.
The structure and parameters of the online branch are inherited from the first three stages of the backbone, while the fourth stage maintains the same structure as the backbone, but its initial parameters are acquired using the presented pre-training strategy~\cite{Bhat2019LearningDM}.

\Remark As UAV-based tracking typically targets small objects and can be accompanied by LR, utilizing the anchor-free method avoids the inaccuracy caused by pre-defined anchors overlapping when confronted with small targets.

\vspace{1em}
\textbf{SiamCorners}:
Since the anchor-based trackers require the miscellaneous design of anchor boxes, which has an impact on the model's applicability convenience, K. X. Yang \emph{et al.}~\cite{Yang2022SiamCornersSC} propose an anchor-free Siamese corner network (SiamCorners).
Using a modified corner pooling layer, the target's bounding box estimate can be turned into a pair of corner predictions (the bottom-right and the top-left corners).
Furthermore, the corner pooling module can estimate multiple corners for a tracking target in deep networks using a layer-wise feature aggregation strategy.
A new penalty term is implemented to suppress substantial changes in the target's size and ratio to choose an optimal tracking box in these candidate corners.
The template is divided into four boundary images and they are cropped for feature extraction.
The top and left boundary image feature maps are provided to the top-left corner prediction branch, while the bottom and right ones are provided to the bottom-right corner prediction branch.
After that, a modified corner pooling is utilized to estimate corner heatmaps and offsets, and the corner candidates are obtained using a simple decoding process.
The corner network is trained follwing H. Law \emph{et al.}~\cite{Law2020CornerNetLiteEK}, and the training objective is a variant of focal loss $L_{\rm vf}$~\cite{Lin2020FocalLF}:
\begin{equation}
\label{eq:variant focal loss}
\begin{split}
L_{\rm vf} = & \frac{-1}{K}\sum\limits_{xy}{\rm \textbf{W}}\ , \\
{\rm \textbf{W}} = & \begin{cases}
(1-\hat{X}_{xy})^{\alpha}{\rm log}(\hat{X}_{xy}), & X_{xy}=1 \\
(1-X_{xy})^{\beta}(\hat{X}_{xy})^{\alpha}{\rm log}(1-\hat{X}_{xy}), & {\rm otherwise}
\end{cases}\ ,
\end{split}
\end{equation}
where $\alpha$ and $\beta$ indicate the weighting factors that control each corner's contribution; $\hat{X}_{xy}$ denotes the amount of penalty reduction is computed by 2D Gaussian; ${X}_{xy}$ denotes the prediction score at the location $(x, y)$, $K$ is the number of candidate corners.
The smooth $L1$ Loss~\cite{Girshick2015FastR} is employed at corner locations to correct the corner offset before remapping the heatmap.
Converting anchor box estimations into corner predictions is an effective method to improve, but corner pooling increases computational expenses, which may slow down the processing speed somewhat and make it difficult to achieve real-time.

\vspace{1em}
\textbf{SiamRCR}:
Classification and regression are commonly used in Siamese trackers.
However, due to the accuracy misalignment between classification and regression, the accuracy of the tracker may be reduced.
To address this problem, J. Peng \emph{et al.}~\cite{Peng2021SiamRCRRC} construct an anchor-free Siamese tracking algorithm, SiamRCR, by building reciprocal links between classification and regression branches and adding a localization branch to predict the localization accuracy.
The correlated feature maps are extracted from the backbone and fed into the corresponding classification and regression branches.
The built-in reciprocal links dynamically re-weight the samples for computing each loss of the two branches.
In particular, the target boxes’ width and height values and the center offsets at the location are directly regressed by treating locations as training samples following J. Long \emph{et al.}~\cite{Long2015FullyCN}.
To eliminate the imbalance between classification and regression, two assistance links are built between them by adopting the dynamically re-weighted losses.
Specifically, the focal loss~\cite{Lin2020FocalLF} is formulated as the dynamically re-weighted classification loss and the IoU loss~\cite{Yu2016UnitboxAA} is formulated as the dynamically re-weighted regression loss.
Moreover, to predict the localization score given the feature maps for regression, the localization branch is developed from the regression branch and supervised by the BCE loss.
The addition of the localization branch and the construction of reciprocal links allow the tracker to achieve a better balance between regression and classification.

\Remark For the design of UAV trackers, it is more essential to cope successfully with the UAV tracking challenges than the approaches adopted.
Whether using an anchor-free or anchor-based method, each has its own range of benefits, and absorbing their respective advantages to improve trackers' performance will be an effective improvement strategy.

\vspace{1em}
\textbf{SiamAPN}:
Although most Siamese trackers have achieved good results in general tracking, some of them do not take into account the UAV tracking challenges, and some are not efficient enough to be applied to UAV tracking.
Siamese Anchor Proposal Network (SiamAPN)~\cite{Fu2021SiameseAP} is designed for obtaining better performance at a faster speed to meet the requirements of UAV-based tracking.
Different from anchor-based methods and anchor-free methods, SiamAPN is built on the concept of anchor proposal, avoiding pre-defining numerous fixed-sized anchors, and acquiring better accuracy through refinement.
For convenient, this work denotes $\varphi_i(\cdot)$ as the output of $i$-th layer for both template image $z$ and search image $x$.
The APN then utilizes $\varphi_4(z)$ and $\varphi_4(x)$, for proposing adaptive anchors.
The similarity map $f_4$ can be formulated as follows:
\begin{equation}
\label{eq:apn4 cross-correlation}
f_4(z, x) = \mathcal{F}(\varphi_4(z)) \star \mathcal{F}(\varphi_4(x))\ ,
\end{equation}
where the feature maps $\varphi_4(z)$ and $\varphi_4(x)$ are adjusted by different convolution operation $\mathcal{F}(\cdot)$.
For each point in the similarity map, APN produces one corresponding anchor.
To eliminate the negative effect caused by the movement of the adaptive anchors, the feature fusion network is designed to establish the connection between APN and the multi-classification\&regression network.
To take advantage of multi-level features information, the similarity map $f_5$ can be obtained by:
\begin{equation}
\label{eq:apn5 cross-correlation}
f_5(z, x) = \mathcal{F}(\varphi_5(z)) \star \mathcal{F}(\varphi_5(x))\ .
\end{equation}

The feature maps $f_4$ and $f_5$ then fused by the channel-wise concatenation and convolutional operation to obtain the response map $\Phi$, as follows:
\begin{equation}
\label{eq:apn feature concatenate}
\Phi = {\rm Cat}(\mathcal{F}(f_4), \mathcal{F}(f_5))\ ,
\end{equation}
where Cat represents channel-wise concatenation.
The response map $\Phi$ is used as input in the multi-classification and regression branches to further refine the proposed anchors.
The multi-classification branch consists of three sub-branches.
They are used to calculate the IoU of each anchor and ground-truth, select the points on the proposed anchor map that lies inside the ground-truth box, and take into account the center distance between each matching point and the ground-truth, respectively.
The regression branch generates one regression feature map, which is combined with the proposed anchor map to produce a more precise position for each anchor.
The final prediction can be made based on the results of the multi-classification and regression branches.
To supervise the generation of the proposed anchors, $L1$ loss is used as the regression loss of APN.
To supervise the entire classification and regression process, the cross-entropy loss and the BCE loss are used on the first two sub-branches and the last sub-branch of the multi-classification branch, while the smooth $L1$ loss and the IoU loss are used on the regression branch.

\Remark Unlike the classical anchor-based and anchor-free methods, the anchor proposal strategy with the efficient two-stage tracking approach can achieve competitive performance while keeping stable real-time capability, making it suitable for UAV deployment.

\vspace{1em}
\textbf{SiamAPN}++:
Since SiamAPN~\cite{Fu2021SiameseAP} may be susceptible to semantic information variation, Z. Cao \emph{et al.}~\cite{Cao2021SiamAPNSA} proposed SiamAPN++ based on the attentional aggregation network (AAN) to improve the robustness to deal with complex scenes.
The attention mechanism is used by AAN, which consists of self-AAN and cross-AAN, to increase feature map expression and filter noise information.
In addition, based on the anchor proposal network (APN)~\cite{Fu2021SiameseAP}, the anchor proposal network-dual feature (APN-DF) is introduced to raise the robustness of proposing anchors using the dual features.
To figure out how various layers interact inside, the similarity maps of the last two layers are acquired as:
\begin{equation}
\label{eq:adapn4 cross-correlation}
f_4(z, x) = \mathcal{F}(\varphi_4(z) \star \varphi_4(x))\ ,
\end{equation}
\begin{equation}
\label{eq:adapn5 cross-correlation}
f_5(z, x) = \mathcal{F}(\varphi_5(z)) \star \mathcal{F}(\varphi_5(x))\ .
\end{equation}

At the same time, to maintain the similarity of cross-interdependencies, the cross-AAN is utilized in the APN-DF.
The feature maps $\boldsymbol{\rm R}_{\rm A} \in \mathbb{R}^{C \times H \times W}$ can be generated by:
\begin{equation}
\label{eq:DF cross-AAN}
\begin{split}
\boldsymbol{\rm R}_{\rm A} = f_5 &+ \gamma_1{\rm FFN}({\rm GAP}(f_4))f_5 \\
&+ \gamma_2\mathcal{F}({\rm Cat}(f_4, f_5))\ ,
\end{split}
\end{equation}
where $\gamma_1$ and $\gamma_2$ denote the learning weights; FFN represents feedforward neural network; GAP indicates global average pooling; Cat represents channel-wise concatenation.
To further release the expression ability of feature maps, AAN is exploited to replace the feature fusion in~\cite{Fu2021SiameseAP}.
The self-AAN aggregates the interdependencies of single feature maps by spatial and channel dimensions.
To explore more abundant semantic information, $f'_5$ can be obtained in the same way as $f_5$ with different convolution layers.
With three different convolution layers, $f_5$ can generate three new feature maps denoted as $\{\boldsymbol{\rm R}^q,\boldsymbol{\rm R}^k\} \in \mathbb{R}^{C \times (H \times W)}$ and $\boldsymbol{\rm R}^v \in \mathbb{R}^{C \times H \times W}$.
Using $\boldsymbol{\rm R}^q$ and $\boldsymbol{\rm R}^k$, the spatial attention map $\boldsymbol{\rm R}^a \in \mathbb{R}^{(H \times W) \times (H \times W)}$ can be acquired.
Therefore, the spatial attention $\boldsymbol{\rm R}^s \in \mathbb{R}^{C \times H \times W}$ calculated as:
\begin{equation}
\label{eq:spatial attention}
\boldsymbol{\rm R}^s = \gamma_3\boldsymbol{\rm R}^v \times \boldsymbol{\rm R}^a + f'_5\ ,
\end{equation}
where $\gamma_3$ denotes the learning weight; $\times$ represents the matrix multiplication.
In addition to spatial attention, the channel attention is also built as:
\begin{equation}
\label{eq:channel attention}
\begin{split}
\boldsymbol{\rm R}_{\rm C} = & \boldsymbol{\rm R}^s + \gamma_4 {\rm Sigmoid}({\rm \textbf{W}}) \boldsymbol{\rm R}^s\ , \\
{\rm \textbf{W}} = & {\rm FFN}({\rm GAP}(\boldsymbol{\rm R}^s)) + {\rm FFN}({\rm GMP}(\boldsymbol{\rm R}^s))\ ,
\end{split}
\end{equation}
where $\gamma_4$ denotes the learning weight; FFN represents feedforward neural network; GMP indicates global max pooling.
The cross-AAN aggregates the cross-interdependencies of different semantic features including the anchor position information.
The cross-AAN includes two cross paths: one explicitly aggregates channel interdependencies by creating a matching weight for each channel of $\boldsymbol{\rm R}_{\rm C}$, while the other implicitly highlights the interdependencies of $\boldsymbol{\rm R}_{\rm A}$ and $\boldsymbol{\rm R}_{\rm C}$.
The final feature maps $\boldsymbol{\rm R} \in \mathbb{R}^{C \times H \times W}$ can be calculated as:
\begin{equation}
\label{eq:final feature maps}
\begin{split}
\boldsymbol{\rm R} = \boldsymbol{\rm R}_{\rm C} &+ \gamma_5 {\rm FFN}({\rm GAP}(\boldsymbol{\rm R}_{\rm A})) \boldsymbol{\rm R}_{\rm C} \\
&+ \gamma_6 \mathcal{F}({\rm Cat}(\boldsymbol{\rm R}_{\rm A}, \boldsymbol{\rm R}_{\rm C}))\ .
\end{split}
\end{equation}

Based on APN-DF and AAN, the classification and regression networks use a SiamAPN-like structure~\cite{Fu2021SiameseAP} to predict the bounding box.

\Remark The two-stage SiamAPN~\cite{Fu2021SiameseAP} and SiamAPN++~\cite{Cao2021SiamAPNSA} use the anchors generated in the first stage to improve the tracking performance of small targets and high-speed targets, the avoidance of pre-defined anchors improves robustness and generalization.
As a result, these trackers are capable of successfully resolving the UAV tracking challenges while running at high speeds.
Designed for UAV-based tracking, they are instructive for future Siamese tracker development.

\subsection{Extra Types}\label{sec:Extra Types}
Some trackers look for additional methods to improve their performance, such as altering the loss function~\cite{Dong2018TripletLI}, examining sample generation strategies~\cite{Wang2018SINTRV}, lightening the model structure~\cite{Yan2021LightTrackFL}, employing re-detection to enhance accuracy~\cite{Voigtlaender2020SiamRV}, and performing long-term tracking of the complete image area~\cite{Huang2020GlobalTrackAS}.

\vspace{1em}
\textbf{SiamTri}:
Logistic loss is frequently used by general Siamese trackers, such as SiamFC, to assess the distance between the exemplar and positive or negative instance.
A triplet loss~\cite{Dong2018TripletLI} is proposed to further explore the probable connection among them.
Based on Eq. (\ref{eq:logistic loss}), the pair-wise logistic loss between exemplar and instance can be written as:
\begin{equation}
\label{eq:pair-wise logistic loss}
L_{\rm l}(Y, V) = \sum\limits_{x_i \in X}w_i{\rm log}(1-e^{-y_i \cdot v_i})\ ,
\end{equation}
where $Y$, $V$ and $X$ denote the sets of ground-truth label, similarity score and
instance input, respectively; $w_i$ is the weight for an instance $x_i$.
Considering the relation between positive and negative instances, the formulation of this matching probability ${\rm prob}(vp_i,vn_j)$ can be described as:
\begin{equation}
\label{eq:matching probability}
{\rm prob}(vp_i,vn_j)=\frac{e^{vp_i}}{e^{vp_i}+e^{vn_j}}\ .
\end{equation}

The triplet loss, then, can be formulated as:
\begin{equation}
\label{eq:triplet loss}
L_{\rm t}(V_{\rm p}, V_{\rm n}) = -\frac{1}{MN}\sum\limits_{M}^{i}\sum\limits_{N}^{j}{\rm log}({\rm prob}(vp_i,vn_j))\ ,
\end{equation}
where the balance weight $\frac{1}{MN}$ is used to keep with the scale of the loss.
More potential information can be acquired by combining $L_{\rm l}$ and $L_{\rm t}$, resulting in a more effective representation with little extra computation during training phase.
In some cases, an appropriate loss function can lower the demand for training samples, which is beneficial for UAV tracking that may lack data for some special scenarios.

\vspace{1em}
\textbf{SINT}++:
Observing that trackers are difficult to have better robustness owing to numerous restrictions, such as a lack of diversity since adopting the dense sampling strategy to generate positive examples, and the inadequate training data for specific issues during the training stage, X. Wang \emph{et al.}~\cite{Wang2018SINTRV} propose the positive samples generation network (PSGN) and the hard positive transformation network (HPTN).
PSGN is used to create a manifold of each tracking video sequence using a variational auto-encoder (VAE)~\cite{Kingma2014AutoEncodingVB}, from which significant positive samples can be extracted.
Using image patches extracted from the background, HPTN can produce occlusions on the target objects.
By combining PSGN and HPTN with a two-streaming Siamese network, SINT++ compensates for the lack of unique training data to improve tracking robustness.
The sample generation strategy can also be used to solve a lack of UAV tracking samples, as well as specialized training to tackle UAV tracking challenges.

\vspace{1em}
\textbf{LightTrack}:
In addition to excellent performance, lightweight design is also important for object tracking.
LightTrack~\cite{Yan2021LightTrackFL} is designed to achieve better performance with fewer model Flops and parameters via taking full advantage of a one-shot neural architecture search (NAS).
Current prevailing object trackers usually use ImageNet for pre-training, which is time-consuming and computation-intensive.
Based on a one-shot NAS, the weight-sharing strategy is introduced to eschew pre-training each candidate from scratch.
The search space of backbone architectures can be encoded into a supernet, and it only needs to be pre-trained once on ImageNet, so that its weights can be shared across different backbone architectures which are subnets of the one-shot model.
In addition to a backbone for feature extraction, Siamese architectures also include a head network for target localization.
They need to be searched as a whole to get fit for the tracking task.
Under the supervision of the BCE loss and IoU loss~\cite{Yu2016UnitboxAA}, the optimizer updates one random path sampled from the backbone and head supernets in each search iteration.
Based on the trained supernet, paths in the supernet are picked and evaluated under the direction of the evolutionary controller.
To search for efficient neural architectures, depth-wise separable convolutions (DSConv)~\cite{Chollet2017XceptionDL} and mobile inverted bottleneck (MBConv)~\cite{Sandler2018MobileNetV2IR} with squeeze-excitation module~\cite{Howard2019SearchingFM,Hu2020SqueezeandExcitationN} are used to construct a new search space.
The efficiency of the tracker has been increased by neural architecture search, allowing for the promotion of more trackers whose running speed does not approach real-time to satisfy the needs of UAV deployment.

\vspace{1em}
\textbf{SiamR}-\textbf{CNN}:
To fully use the capabilities of two-stage object detection methods for visual object tracking, P. Voigtlaender \emph{et al.}~\cite{Voigtlaender2020SiamRV} presents a Siamese re-detection architecture (SiamR-CNN) with a tracklet-based dynamic programming algorithm, to model the full history of both the target and potential distractors by re-detecting both the first-frame template and previous-frame predictions.
A hard example mining technique for increasing the discriminative ability of the re-detection head is presented to counteract the detrimental impacts of similar objects.
Instead of obtaining general hard examples for detection, hard examples are found for re-detection conditioned on the reference object by retrieving objects from other videos.
Based on a Faster R-CNN~\cite{Ren2015FasterRT} network, SiamR-CNN consists of a backbone feature extractor, a category-agnostic RPN, and a category-specific detection head.
Among them, the weights of the backbone and the RPN are fixed while a re-detection head is used in place of the category-specific detection head.
The re-detection head's input features are proposed by the RPN using RoI Align~\cite{He2020MaskR} and are concatenated with the RoI Aligned deep features of the initialization bounding box.
After that, these joint features are given into the re-detection head using a three-stage cascade~\cite{Cai2018CascadeRD} without shared weights.
Using spatio-temporal information, the Tracklet Dynamic Programming Algorithm (TDPA) keeps track of both the object of interest and any possible distractions.
It then selects the most matching tracklets for the template from the first and current frames using a dynamic programming-based scoring system.
Additionally, an off-the-shelf bounding-box-to-segmentation (Box2Seg)~\cite{Luiten2018PReMVOSPR} network is used to produce segmentation masks.
The performance of SiamR-CNN has achieved an incredible breakthrough after being introduced in various strategies, however, the enormous sacrifice of running speed is regrettable.

\vspace{1em}
\textbf{GlobalTrack}:
To fill the void of a solid baseline for global instance search that can withstand target absences or tracking failures, L. Huang \emph{et al.}~\cite{Huang2020GlobalTrackAS} propose GlobalTrack.
GlobalTrack is built on two-stage object detectors that allow it to perform full-image and multi-scale searches of random instances using only a single query as a guide.
Moreover, a cross-query loss is proposed to improve the robustness of this method against distractors.
Similar to Faster-RCNN~\cite{Ren2015FasterRT}, the architecture of GlobalTrack consists of two submodules: Query-Guided RPN (QG-RPN) and Query-Guided RCNN (QG-RCNN): the former is for generating query-specific proposals, and the latter is for discriminating the proposals and delivering the final predictions.
The goal of QG-RPN is to encode query information in backbone features via correlation.
The BCE loss and the smooth $L1$ loss are used as the classification and localization losses~\cite{Girshick2015FastR} for training QG-RPN.
Based on proposals generated by QG-RPN, QG-RCNN is used for the classification and bounding box refinement of these proposals.
Note that similar losses are used for training QG-RCNN.
To raise GlobalTrack's awareness of the connection between queries and prediction outputs, the same image can be searched using different queries, and then their prediction losses are averaged to balance both QG-RPN and QG-RCNN.
The cross-query loss can be written as:
\begin{equation}
\label{eq:cross-query loss}
L_{\rm cql} = \frac{1}{M}\sum\limits^M(L_{\rm qg\_rpn}+L_{\rm qg\_rcnn})\ ,
\end{equation}
where $L_{\rm qg\_rpn}$ and $L_{\rm qg\_rcnn}$ denote the losses of QG-RPN and QG-RCNN respectively, and $M$ query-search image pairs are constructed to calculating the loss.
Long-term tracking and back detection after the target have vanished are feasible with the full-image search strategy, but persistently high computing costs may not be practical for UAV deployment.

\Remark When designing a UAV tracker, performance is vital, but finding the correct balance between high performance and great efficiency is even more important.

\section{Experimental Evaluation}\label{sec:Experimental Evaluation}
After having a comprehensive understanding of the Siamese tracker, this work conducts a range of experiments for UAV-based deployment and present relevant analyses.
To begin with, implementation information is detailed in Sec.~\ref{sec:Implementation Information}, including an introduction to three frequently used evaluation metrics, a description of six authoritative public UAV benchmarks, as well as the specifications of the experiment platform and parameter settings.
Moreover, Sec.~\ref{sec:Overall Performance} presents an overall evaluation of the trackers' precision, success, and speed.
Based on the results, the trackers that can match the UAV's real-time requirements are selected for further analysis.
Faced with the challenges of UAV-based tracking deployment, Sec.~\ref{sec:Attribute-Based Evaluation} proposes and examines the performance of these trackers in response to re-defined UAV tracking attributes.
To better cope with severe UAV tracking challenges in future research, failure cases are analyzed in Sec.~\ref{sec:Failure Cases}.
Based on the aforementioned analysis, the two most outstanding trackers are chosen for the onboard tests to validate the feasibility and efficiency of Siamese trackers for real-world UAV deployment in Sec.~\ref{sec:Onboard Tests}.
Finally, in allusion to demonstrate the performance of Siamese trackers in low-illumination environments, Sec.~\ref{sec:Low-Illumination Analysis} performs a low-illumination evaluation and discusses the impact of low-illumination scenarios on Siamese UAV tracking.

\subsection{Implementation Information}\label{sec:Implementation Information}
Before the experimental evaluation, the details of the implementation are first explained.
To precisely and effectively evaluate the performance of the trackers, three evaluation metrics, \emph{i.e.}, success rate, precision, and normalized precision~\cite{Mller2018TrackingNetAL}, are introduced based on the one-pass evaluation (OPE)~\cite{Wu2015ObjectTB}.
Furthermore, the experimental evaluation is based on six authoritative public UAV benchmarks, \emph{i.e.}, UAV123@10fps~\cite{Mueller2016ABA}, UAV20L~\cite{Mueller2016ABA}, DTB70~\cite{Li2017VisualOT}, UAVDT~\cite{Du2018TheUA}, VisDrone-SOT2020-test~\cite{Fan2020VisDroneSOT2020TV}, and UAVTrack112~\cite{Fu2021OnboardRA}.
The experiment platform and parameter settings of the implemented code are described in detail.

\subsubsection{Evaluation Metrics}\label{sec:Evaluation Metrics}
Based on the OPE~\cite{Wu2015ObjectTB}, the overlap score (OS) and center-location error (CLE) are commonly used to evaluate the success rate and precision of object tracking.
An OPE stands for initializing the tracker with the object's location and scale in the first frame's ground-truth, so that the tracker can calculate and predict the bounding boxes in subsequent frames.
Based on the prediction, the precision and success rate plots can be portrayed to illustrate the trackers' performance.
Considering that the precision metric can be easily affected by the image resolution and the bounding box scale, the normalized precision metric ~\cite{Mller2018TrackingNetAL} is also included in the experimental evaluation.

\begin{figure}[ht]
\centering
\includegraphics[width=3.5in]{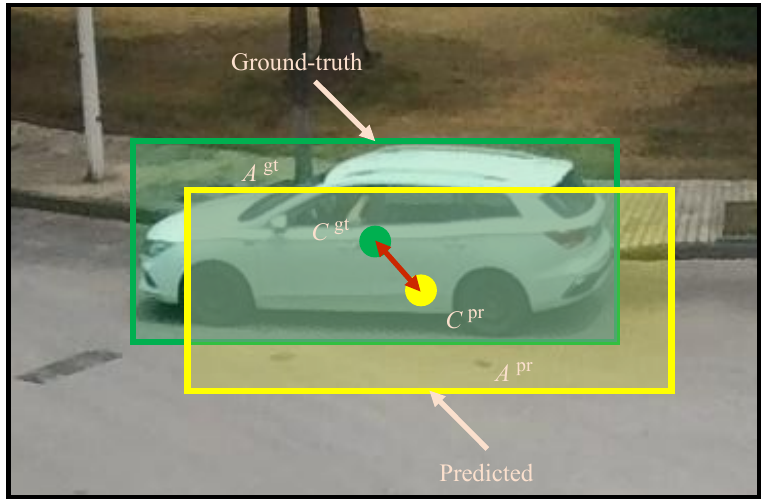}%
\caption{The area and location center of the ground-truth and predicted bounding boxes.
They are used in three evaluation metrics for tracking.
(Sequence courtesy of benchmark UAVTrack112~\cite{Fu2021OnboardRA}.)
}
\label{fig:metrics}
\end{figure}

For the success rate plot, as shown in Fig.~\ref{fig:metrics}, the OS is first calculated in every frame as the Intersection over Union (IoU) of the area in the ground-truth bounding boxes ($A^{\rm gt}$) and the one predicted by the tracker ($A^{\rm pr}$), as follows:
\begin{equation}
\label{eq:OS}
{\rm OS} = \frac{|A^{\rm gt} \cap A^{\rm pr}|}{|A^{\rm gt} \cup A^{\rm pr}|}\ ,
\end{equation}
where $|\cdot|$ denotes the number of pixels in the area.
By comparing the OS of a frame to a set threshold, it is possible to determine whether or not this frame is successful.
The ratio of total successful frames to all frames represents the success rate within one threshold.
By consecutively arranging thresholds from zero to one, the corresponding success rate can be formed into a curve.
In the general evaluation, the area under the curve (AUC) is calculated to place trackers in a ranking.

For the precision plot, as shown in Fig.~\ref{fig:metrics}, the CLE, which is defined by the distance between the center point in the ground-truth ($C^{\rm gt}$) and the center point of the bounding box predicted by the tracking method ($C^{\rm pr}$), should be determined in every frame, as follows:
\begin{equation}
\label{eq:CLE}
{\rm CLE} = ||C^{\rm gt}-C^{\rm pr}||\ ,
\end{equation}
where $||\cdot||$ represents the Euclidean distances between two points.
By computing the CLE of each video frame, the precision can be represented as the ratio of the number of frames above a specified threshold to the total number of frames.
A precision curve plot, similar to the success rate curve plot, can be generated based on different pixel distance thresholds.
During the general evaluation, the threshold is set at 20 pixels for the ranking of all trackers, \emph{i.e.}, distance precision (DP) at a CLE = 20 pixels.

Since the precision metric can be easily affected by the image resolution and the bounding box scale, \cite{Mller2018TrackingNetAL} suggests the normalized precision metric to evaluate the precision.
The normalized precision plot is similar to the precision plot, except the precision is normalized over the ground-truth bounding box scale.
Therefore, the $CLE_{\rm norm}$ can be calculated by reformulating Eq. (\ref{eq:CLE}) as:
\begin{equation}
\label{eq:normalized CLE}
\begin{split}
{\rm CLE}_{\rm norm} =\  & ||{\rm \textbf{W}}(C^{\rm gt}-C^{\rm pr})||\ , \\
{\rm \textbf{W}} =\  & {\rm diag}(A^{\rm gt}_x, A^{\rm gt}_y)\ .
\end{split}
\end{equation}

Based on the normalized precision curve plot, the AUC between 0 and 0.5 is then used to rank the trackers.

\Remark In this experimental evaluation, this work compares the performance of trackers using all three evaluation metrics.

\subsubsection{Benchmarks}\label{sec:Benchmarks}
The experimental evaluation is based on six authoritative public UAV benchmarks, \emph{i.e.}, UAV123@10fps~\cite{Mueller2016ABA}, UAV20L~\cite{Mueller2016ABA}, DTB70~\cite{Li2017VisualOT}, UAVDT~\cite{Du2018TheUA}, VisDrone-SOT2020-test~\cite{Fan2020VisDroneSOT2020TV}, and UAVTrack112~\cite{Fu2021OnboardRA}.
The details and characteristics of each benchmark are introduced briefly.

\begin{table*}[hb]
\caption{Details of each UAV tracking benchmark.
They include numbers of sequences, minimum, maximum, average frames in each sequence, and total frames of the six UAV tracking benchmarks, \emph{i.e.}, UAV123@10fps, UAV20L, DTB70, UAVDT, VisDrone-SOT2020-test, and UAVTrack112.
The first two places are emphasized with \underline{underline}.
}
\centering
\renewcommand{\arraystretch}{1.3}
\newcommand{\tabincell}[2]{\begin{tabular}{@{}#1@{}}#2\end{tabular}}
\begin{tabular}{ c c c c c c }
\toprule
\textbf{Benchmark} & \textbf{Sequences} & \textbf{Minimum Frames} & \textbf{Maximum Frames} & \textbf{Average Frames} & \textbf{Total Frames} \\
\midrule
UAV123@10fps~\cite{Mueller2016ABA} & \underline{123} & 37 & 1,029 & 308 & 37,885 \\
UAV20L~\cite{Mueller2016ABA} & 20 & \underline{1,717} & \underline{5,527} & \underline{2934} & \underline{58,670} \\
DTB70~\cite{Li2017VisualOT} & 70 & 68 & 699 & 225 & 15,777 \\
UAVDT~\cite{Du2018TheUA} & 50 & 82 & \underline{2,969} & 742 & 37,084 \\
VisDrone-SOT2020-test~\cite{Fan2020VisDroneSOT2020TV} & 35 & \underline{90} & 2,783 & \underline{941} & 32,922 \\
UAVTrack112~\cite{Fu2021OnboardRA} & \underline{112} & 19 & 2,624 & 896 & \underline{100,313} \\
\bottomrule
\end{tabular}
\label{tab:benchmarks}
\end{table*}

\textbf{UAV123@10fps}:
123 completely annotated HD video sequences captured from low-altitude UAVs are included in UAV123~\cite{Mueller2016ABA}.
There are 112,578 frames in total in these sequences, which span a wide range of scenes and objects from an aerial perspective.
To explore the influence of speed on performance and imitate discontinuous aerial video sampling, UAV123@10fps is downsampled from its 30 frames per second (FPS) version.

\textbf{UAV20L}:
The object's location changes between frames get bigger and more irregular as the frame interval is longer, rendering object tracking more challenging.
Since long-term tracking is another issue that UAV tracking frequently encounters, UAV20L~\cite{Mueller2016ABA} comprises the 20 longest sequences formed by merging short subsequences.

\textbf{DTB70}:
DTB70~\cite{Li2017VisualOT} consists of 70 sequences totaling 15,777 frames that aim to address the difficult subject of significant camera motion in various extreme conditions.

\textbf{UAVDT}:
UAVDT~\cite{Du2018TheUA} is designed to focus on complicated circumstances like weather, altitude, camera view, vehicle type, and occlusion.
The single object tracking (SOT) portion, includes 50 sequences and 37,084 frames.

\textbf{VisDrone-SOT2020-test}:
VisDrone-SOT2020~\cite{Fan2020VisDroneSOT2020TV} is based on the Vision Meets Drone Single-Object Tracking challenge.
The dataset is collected in a variety of real-world circumstances using various drone platforms in different weather and lighting conditions, which is worthwhile to evaluate the tracker's performance in real-world scenarios.
The testing set, \emph{i.e.}, VisDrone-SOT2020-test, is applied in this experimental evaluation.

\textbf{UAVTrack112}:
UAVTrack112~\cite{Fu2021OnboardRA} is designed for aerial tracking, contains 112 sequences with representative scenes and objects, in particular, a vast amount of cityscape scenarios.
These sequences are collected and annotated from the real-world with aerial-specific challenges.

As shown in TABLE~\ref{tab:benchmarks}, details are displayed including numbers of sequences, minimum, maximum, average frames in each sequence, and total frames of the six UAV tracking benchmarks~\cite{Mueller2016ABA,Li2017VisualOT,Du2018TheUA,Fan2020VisDroneSOT2020TV,Fu2021OnboardRA}.

\subsubsection{Experimental Platform}\label{sec:Experimental Platform}
The large-scale evaluation experiments are conducted on an NVIDIA Jetson AGX Xavier with an 8-core Carmel ARM v8.2 64-bit CPU, 512-core Volta GPU, and 32 GB of random-access memory (RAM) using MAXN nvpmodel.

\subsubsection{Parameter Settings}\label{Parameter Settings}
To assure the experiment's fairness and objectivity, as well as the result's reliability, all of the trackers evaluated have maintained their official models and parameters from their open-source codes.

\Remark For the trackers that use different backbone networks, such as in SiamRPN++~\cite{Li2019SiamRPNEO}, A, R, and M represent AlexNet~\cite{Krizhevsky2012ImageNetCW}, ResNet-50~\cite{He2016DeepRL}, and MobileNetV2~\cite{Sandler2018MobileNetV2IR}, respectively; in SiamDW~\cite{Zhang2019DeeperAW}, CR, CI, and CX represent CIResNet-22, CIResIncep.-22, and CIResNeXt-22, respectively.

\subsection{Overall Performance}\label{sec:Overall Performance}

\begin{figure*}[!t]
\centering
\subfloat[]{\includegraphics[width=6.8in]{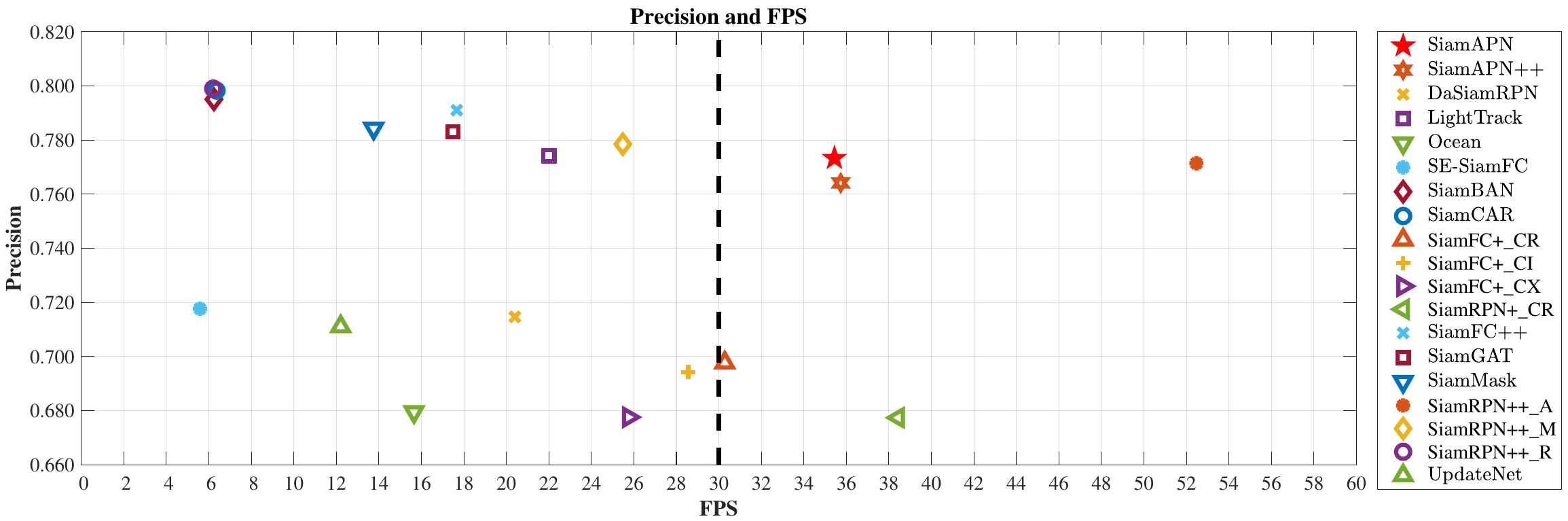}%
\label{fig:star_prec}}
\hfil
\subfloat[]{\includegraphics[width=6.8in]{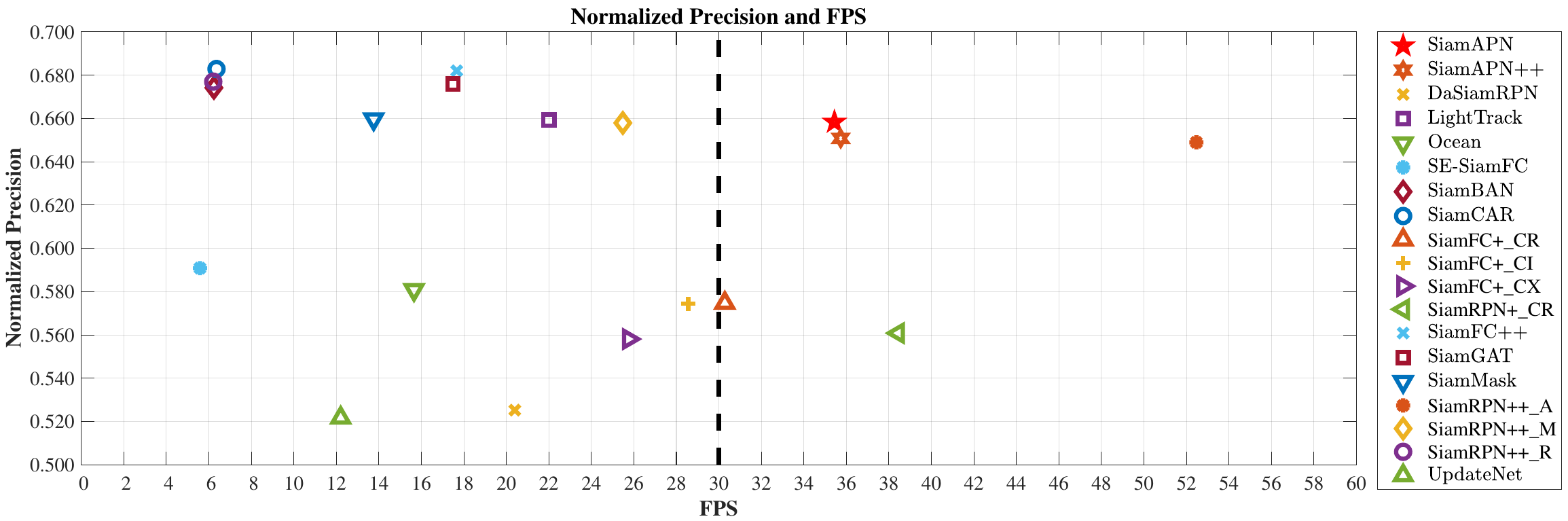}%
\label{fig:star_np}}
\hfil
\subfloat[]{\includegraphics[width=6.8in]{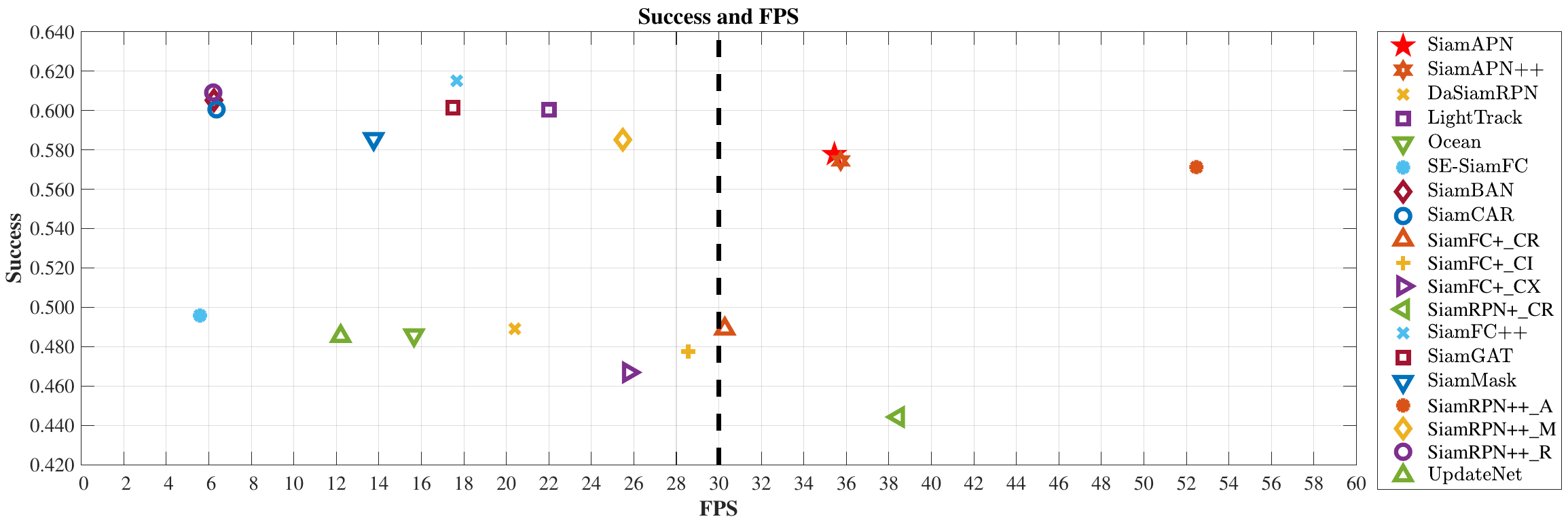}%
\label{fig:star_succ}}
\caption{Comparison of the average performance of famous and recent Siamese trackers under all the six authoritative UAV tracking benchmarks~\cite{Mueller2016ABA,Fan2020VisDroneSOT2020TV,Li2017VisualOT,Du2018TheUA,Fu2021OnboardRA}.
When the tracking speed reaches the dotted line (30 FPS) on the typical UAV onboard processor, it meets the requirements of UAV real-time tracking.
}
\label{fig:star}
\end{figure*}

To analyze Siamese trackers' performance in UAV tracking circumstances, 19 famous and recent Siamese trackers with different backbone networks are evaluated, \emph{i.e.}, DaSiamRPN~\cite{Zhu2018DistractorawareSN}, LightTrack~\cite{Yan2021LightTrackFL}, Ocean~\cite{Zhang2020OceanOA}, SE-SiamFC~\cite{Sosnovik2021ScaleEI}, SiamAPN~\cite{Fu2021SiameseAP}, SiamAPN++~\cite{Cao2021SiamAPNSA}, SiamBAN~\cite{Chen2020SiameseBA}, SiamCAR~\cite{Guo2020SiamCARSF}, SiamDW (SiamFC+\_CR, SiamFC+\_CI, SiamFC+\_CX, and SiamRPN+\_CR)~\cite{Zhang2019DeeperAW}, SiamFC++~\cite{Xu2020SiamFCTR}, SiamGAT~\cite{Guo2021GraphAT}, SiamMask~\cite{Wang2019FastOO}, SiamRPN++ (SiamRPN++\_A, SiamRPN++\_M, and SiamRPN++\_R)~\cite{Li2019SiamRPNEO}, and UpdateNet~\cite{Zhang2019LearningTM}.
Their tracking results are obtained on a typical UAV onboard processor using the six authoritative benchmarks ~\cite{Mueller2016ABA,Fan2020VisDroneSOT2020TV,Li2017VisualOT,Du2018TheUA,Fu2021OnboardRA}.

As shown in Fig.~\ref{fig:star}, each tracker's average overall performance differs somewhat but follows similar trends throughout the three evaluation metrics.
SiamAPN~\cite{Fu2021SiameseAP} provides remarkable overall performance in precision, normalized precision, and success rate while satisfying UAV speed requirements, \emph{i.e.}, 30 FPS, followed by SiamAPN++~\cite{Cao2021SiamAPNSA} and SiamRPN++\_A~\cite{Li2019SiamRPNEO}.
As can be observed, the tracking performance of the Siamese trackers has been steadily improving in recent years due to their rapid development.
However, meeting the requirements of UAV deployment in terms of speed is generally challenging.

A desirable advance in the performance of trackers with a deeper backbone has been achieved but at the expense of speed.
When faced with relatively limited computer resources on UAVs, their computing efficiency is not adequate, even though they can reluctantly fulfill real-time needs with good performance on computing devices.
To be specific, SiamRPN++\_R~\cite{Li2019SiamRPNEO}, SiamBAN~\cite{Chen2020SiameseBA}, and SiamCAR~\cite{Guo2020SiamCARSF} reach fewer than 10 FPS, SiamMask~\cite{Wang2019FastOO}, SiamFC++~\cite{Xu2020SiamFCTR}, and SiamGAT~\cite{Guo2021GraphAT} get less than 20 FPS, while LightTrack~\cite{Yan2021LightTrackFL} and SiamRPN++\_M~\cite{Li2019SiamRPNEO} accomplish nearly 30 FPS.
With a relatively shallow backbone, SiamAPN~\cite{Fu2021SiameseAP} and SiamAPN++~\cite{Cao2021SiamAPNSA} have made appropriate trade-offs in the balance of speed and performance, making them a solid choice for UAV tracking.
The topic of how to make the most of the backbone's feature extraction capabilities while boosting the tracker's efficiency will be one to ponder.

TABLEs~\ref{tab:overall1}~and~\ref{tab:overall2} compare the detailed performance and speed of the majority of the SOTA Siamese trackers.
From the results, the performance of different trackers on different datasets can be observed and compared.
On the UAV20L~\cite{Mueller2016ABA}, which is mainly based on long sequences, SiamGAT~\cite{Guo2021GraphAT} and LightTrack~\cite{Yan2021LightTrackFL} have made good breakthroughs.
Based on the deeper backbone, ResNet-50~\cite{He2016DeepRL}, Siamese trackers, \emph{e.g.}, SiamRPN++\_R~\cite{Li2019SiamRPNEO}, SiamBAN~\cite{Chen2020SiameseBA}, SiamCAR~\cite{Guo2020SiamCARSF}, and SiamMask~\cite{Wang2019FastOO}, perform admirably across a variety of UAV datasets.
With a more effective bounding box prediction strategy, SiamAPN~\cite{Fu2021SiameseAP} and SiamAPN++~\cite{Cao2021SiamAPNSA} are particularly suited to UAV-based tracking.

Speed is critical for UAV-based tracking, and optimal performance should be determined by the desired operating speed.
Although some trackers can obtain higher performance, their processing speeds are not ideal.
The running speed of 30 FPS can be recognized as the speed that meets the real-time requirements~\cite{Mueller2016ABA,Fu2022CorrelationFF}.
As shown in Fig.~\ref{fig:star}, five Siamese trackers, \emph{i.e.}, SiamAPN~\cite{Fu2021SiameseAP}, SiamAPN++~\cite{Cao2021SiamAPNSA}, SiamFC+\_CR~\cite{Zhang2019DeeperAW}, SiamRPN+\_CR~\cite{Zhang2019DeeperAW}, and SiamRPN++\_A~\cite{Li2019SiamRPNEO}, can run faster than 30 FPS on NVIDIA Jetson AGX Xavier and fulfill real-time requirements.

\begin{table*}[!hb]
\caption{FPS (frames per second), DP (distance precision at CLE = 20 pixels), NDP (normalized distance precision), AUC (area under curve) comparison of the Siamese trackers on UAV123@10fps, UAV20L, DTB70.
The first, second and third places, respectively, are denoted by \textbf{\color[RGB]{220,0,0} red}, \textbf{\color[RGB]{0,220,0} green}, and \textbf{\color[RGB]{0,0,220} blue}.
Real-time trackers are displayed at the bottom of the table.
Note that all the trackers maintained their official parameters during the evaluation.
}
\centering
\renewcommand{\arraystretch}{1.3}
\newcommand{\tabincell}[2]{\begin{tabular}{@{}#1@{}}#2\end{tabular}}
\begin{tabular}{ l | c c c c | c c c c | c c c c }
\toprule
\multirow{2}{*}{\diagbox{\textbf{Tracker}}{\textbf{Benchmark}}} & \multicolumn{4}{c|}{\textbf{UAV123@10fps}~\cite{Mueller2016ABA}} & \multicolumn{4}{c|}{\textbf{UAV20L}~\cite{Mueller2016ABA}} & \multicolumn{4}{c}{\textbf{DTB70}~\cite{Li2017VisualOT}} \\
 & \textbf{FPS} & \textbf{DP} & \textbf{NDP} & \textbf{AUC} & \textbf{FPS} & \textbf{DP} & \textbf{NDP} & \textbf{AUC} & \textbf{FPS} & \textbf{DP} & \textbf{NDP} & \textbf{AUC} \\
\midrule
DaSiamRPN~\cite{Zhu2018DistractorawareSN} & 20.6 & 0.692 & 0.528 & 0.483 & 20.6 & 0.631 & 0.492 & 0.442 & 20.4 & 0.705 & 0.502 & 0.474 \\
LightTrack~\cite{Yan2021LightTrackFL} & 22.7 & 0.776 & 0.669 & \textbf{\color[RGB]{0,220,0} 0.598} & 20.9 & \textbf{\color[RGB]{0,220,0} 0.791} & \textbf{\color[RGB]{0,220,0} 0.707} & \textbf{\color[RGB]{0,220,0} 0.620} & 21.7 & 0.761 & 0.638 & 0.587 \\
Ocean~\cite{Zhang2020OceanOA} & 16.7 & 0.657 & 0.573 & 0.461 & 17.6 & 0.630 & 0.574 & 0.444 & 16.2 & 0.634 & 0.533 & 0.455 \\
SE-SiamFC~\cite{Sosnovik2021ScaleEI} & 5.5 & 0.717 & 0.617 & 0.512 & 5.5 & 0.648 & 0.593 & 0.453 & 5.7 & 0.730 & 0.573 & 0.490 \\
SiamBAN~\cite{Chen2020SiameseBA} & 6.2 & 0.770 & 0.665 & 0.585 & 6.2 & 0.736 & 0.662 & 0.564 & 6.2 & \textbf{\color[RGB]{220,0,0} 0.832} & \textbf{\color[RGB]{0,220,0} 0.696} & \textbf{\color[RGB]{220,0,0} 0.643} \\
SiamCAR~\cite{Guo2020SiamCARSF} & 6.4 & \textbf{\color[RGB]{220,0,0} 0.789} & \textbf{\color[RGB]{0,220,0} 0.682} & \textbf{\color[RGB]{0,0,220} 0.596} & 6.2 & 0.687 & 0.619 & 0.523 & 6.3 & \textbf{\color[RGB]{0,220,0} 0.831} & \textbf{\color[RGB]{220,0,0} 0.700} & 0.603 \\
SiamFC+\_CI~\cite{Zhang2019DeeperAW} & 28.5 & 0.684 & 0.588 & 0.482 & 28.3 & 0.632 & 0.587 & 0.407 & 28.1 & 0.694 & 0.547 & 0.472 \\
SiamFC+\_CX~\cite{Zhang2019DeeperAW} & 25.7 & 0.665 & 0.570 & 0.470 & 25.8 & 0.583 & 0.543 & 0.388 & 25.5 & 0.681 & 0.535 & 0.454 \\
SiamFC++~\cite{Xu2020SiamFCTR} & 17.7 & 0.759 & 0.665 & 0.589 & 17.7 & 0.742 & 0.668 & 0.575 & 17.5 & \textbf{\color[RGB]{0,0,220} 0.812} & \textbf{\color[RGB]{0,0,220} 0.689} & \textbf{\color[RGB]{0,220,0} 0.637} \\
SiamGAT~\cite{Guo2021GraphAT} & 17.5 & \textbf{\color[RGB]{0,220,0} 0.788} & \textbf{\color[RGB]{220,0,0} 0.688} & \textbf{\color[RGB]{220,0,0} 0.602} & 17.4 & \textbf{\color[RGB]{220,0,0} 0.796} & \textbf{\color[RGB]{220,0,0} 0.719} & \textbf{\color[RGB]{220,0,0} 0.620} & 17.3 & 0.752 & 0.643 & 0.583 \\
SiamMask~\cite{Wang2019FastOO} & 13.7 & \textbf{\color[RGB]{0,0,220} 0.788} & \textbf{\color[RGB]{0,0,220} 0.672} & 0.590 & 13.6 & 0.679 & 0.612 & 0.514 & 13.5 & 0.775 & 0.636 & 0.575 \\
SiamRPN++\_M~\cite{Li2019SiamRPNEO} & 25.7 & 0.771 & 0.661 & 0.578 & 25.4 & 0.723 & 0.652 & 0.547 & 24.7 & 0.785 & 0.643 & 0.593 \\
SiamRPN++\_R~\cite{Li2019SiamRPNEO} & 6.2 & 0.784 & 0.671 & 0.594 & 6.1 & \textbf{\color[RGB]{0,0,220} 0.758} & \textbf{\color[RGB]{0,0,220} 0.679} & \textbf{\color[RGB]{0,0,220} 0.579} & 6.2 & 0.798 & 0.661 & \textbf{\color[RGB]{0,0,220} 0.614} \\
UpdateNet~\cite{Zhang2019LearningTM} & 12.3 & 0.682 & 0.519 & 0.475 & 12.3 & 0.634 & 0.493 & 0.442 & 12.2 & 0.711 & 0.505 & 0.477 \\
\hline
SiamAPN~\cite{Fu2021SiameseAP} & 35.3 & 0.760 & 0.660 & 0.571 & 34.5 & 0.692 & 0.624 & 0.518 & 35.0 & 0.784 & 0.654 & 0.586 \\
SiamAPN++~\cite{Cao2021SiamAPNSA} & \textbf{\color[RGB]{0,0,220} 35.7} & 0.764 & 0.669 & 0.581 & \textbf{\color[RGB]{0,0,220} 34.9} & 0.703 & 0.639 & 0.533 & \textbf{\color[RGB]{0,0,220} 35.1} & 0.790 & 0.655 & 0.594 \\
SiamFC+\_CR~\cite{Zhang2019DeeperAW} & 30.4 & 0.676 & 0.582 & 0.491 & 30.2 & 0.585 & 0.541 & 0.408 & 30.0 & 0.726 & 0.568 & 0.492 \\
SiamRPN+\_CR~\cite{Zhang2019DeeperAW} & \textbf{\color[RGB]{0,220,0} 38.6} & 0.642 & 0.545 & 0.425 & \textbf{\color[RGB]{0,220,0} 38.2} & 0.542 & 0.505 & 0.382 & \textbf{\color[RGB]{0,220,0} 37.3} & 0.709 & 0.570 & 0.453 \\
SiamRPN++\_A~\cite{Li2019SiamRPNEO} & \textbf{\color[RGB]{220,0,0} 52.0} & 0.737 & 0.633 & 0.551 & \textbf{\color[RGB]{220,0,0} 51.5} & 0.701 & 0.634 & 0.533 & \textbf{\color[RGB]{220,0,0} 51.6} & 0.793 & 0.657 & 0.586 \\
\bottomrule
\end{tabular}
\label{tab:overall1}
\end{table*}

\begin{table*}[!hb]
\caption{FPS (frames per second), DP (distance precision at CLE = 20 pixels), NDP (normalized distance precision), AUC (area under curve) comparison of the Siamese trackers on UAVDT, VisDrone-SOT2020-test, UAVTrack112.
The first, second and third places, respectively, are denoted by \textbf{\color[RGB]{220,0,0} red}, \textbf{\color[RGB]{0,220,0} green}, and \textbf{\color[RGB]{0,0,220} blue}.
Real-time trackers are displayed at the bottom of the table.
Note that all the trackers maintained their official parameters during the evaluation.
}
\centering
\renewcommand{\arraystretch}{1.3}
\newcommand{\tabincell}[2]{\begin{tabular}{@{}#1@{}}#2\end{tabular}}
\begin{tabular}{ l | c c c c | c c c c | c c c c }
\toprule
\multirow{2}{*}{\diagbox{\textbf{Tracker}}{\textbf{Benchmark}}} & \multicolumn{4}{c|}{\textbf{UAVDT}~\cite{Du2018TheUA}} & \multicolumn{4}{c|}{\textbf{VisDrone-SOT2020-test}~\cite{Fan2020VisDroneSOT2020TV}} & \multicolumn{4}{c}{\textbf{UAVTrack112}~\cite{Fu2021OnboardRA}} \\
 & \textbf{FPS} & \textbf{DP} & \textbf{NDP} & \textbf{AUC} & \textbf{FPS} & \textbf{DP} & \textbf{NDP} & \textbf{AUC} & \textbf{FPS} & \textbf{DP} & \textbf{NDP} & \textbf{AUC} \\
\midrule
DaSiamRPN~\cite{Zhu2018DistractorawareSN} & 20.4 & 0.794 & 0.515 & 0.498 & 20.2 & 0.763 & 0.574 & 0.536 & 20.2 & 0.710 & 0.532 & 0.495 \\
LightTrack~\cite{Yan2021LightTrackFL} & 23.7 & 0.776 & 0.627 & 0.59 & 19.3 & 0.754 & 0.629 & 0.579 & 21.7 & 0.783 & 0.677 & 0.619 \\
Ocean~\cite{Zhang2020OceanOA} & 14.3 & 0.725 & 0.595 & 0.523 & 15.4 & 0.706 & 0.565 & 0.500 & 14.5 & 0.713 & 0.619 & 0.519 \\
SE-SiamFC~\cite{Sosnovik2021ScaleEI} & 5.6 & 0.626 & 0.454 & 0.405 & 5.6 & 0.734 & 0.610 & 0.544 & 5.6 & 0.759 & 0.628 & 0.515 \\
SiamBAN~\cite{Chen2020SiameseBA} & 6.3 & \textbf{\color[RGB]{220,0,0} 0.806} & \textbf{\color[RGB]{0,220,0} 0.650} & \textbf{\color[RGB]{220,0,0} 0.601} & 6.2 & 0.770 & 0.631 & 0.567 & 6.3 & 0.813 & 0.697 & \textbf{\color[RGB]{0,0,220} 0.625} \\
SiamCAR~\cite{Guo2020SiamCARSF} & 6.3 & \textbf{\color[RGB]{0,220,0} 0.804} & \textbf{\color[RGB]{220,0,0} 0.660} & \textbf{\color[RGB]{0,0,220} 0.598} & 6.4 & \textbf{\color[RGB]{220,0,0} 0.838} & \textbf{\color[RGB]{220,0,0} 0.704} & \textbf{\color[RGB]{220,0,0} 0.630} & 6.4 & 0.793 & 0.688 & 0.610 \\
SiamFC+\_CI~\cite{Zhang2019DeeperAW} & 28.8 & 0.639 & 0.486 & 0.422 & 28.7 & 0.744 & 0.605 & 0.540 & 28.8 & 0.726 & 0.604 & 0.494 \\
SiamFC+\_CX~\cite{Zhang2019DeeperAW} & 25.9 & 0.641 & 0.463 & 0.416 & 25.8 & 0.708 & 0.579 & 0.523 & 25.9 & 0.713 & 0.598 & 0.491 \\
SiamFC++~\cite{Xu2020SiamFCTR} & 17.7 & \textbf{\color[RGB]{0,0,220} 0.802} & \textbf{\color[RGB]{0,0,220} 0.648} & \textbf{\color[RGB]{0,220,0} 0.600} & 17.7 & 0.788 & \textbf{\color[RGB]{0,0,220} 0.666} & \textbf{\color[RGB]{0,220,0} 0.609} & 17.7 & \textbf{\color[RGB]{0,220,0} 0.818} & \textbf{\color[RGB]{220,0,0} 0.719} & \textbf{\color[RGB]{220,0,0} 0.646} \\
SiamGAT~\cite{Guo2021GraphAT} & 17.7 & 0.754 & 0.622 & 0.574 & 17.5 & \textbf{\color[RGB]{0,220,0} 0.811} & \textbf{\color[RGB]{0,220,0} 0.687} & \textbf{\color[RGB]{0,0,220} 0.606} & 17.5 & 0.799 & 0.696 & 0.620 \\
SiamMask~\cite{Wang2019FastOO} & 14.1 & 0.782 & 0.632 & 0.580 & 13.6 & \textbf{\color[RGB]{0,0,220} 0.806} & 0.659 & 0.588 & 13.9 & 0.799 & 0.682 & 0.602 \\
SiamRPN++\_M~\cite{Li2019SiamRPNEO} & 26.2 & 0.757 & 0.612 & 0.556 & 24.9 & 0.795 & 0.659 & 0.598 & 25.6 & 0.797 & 0.685 & 0.604 \\
SiamRPN++\_R~\cite{Li2019SiamRPNEO} & 6.3 & 0.801 & 0.642 & 0.594 & 6.2 & 0.788 & 0.654 & 0.596 & 6.2 & \textbf{\color[RGB]{220,0,0} 0.826} & \textbf{\color[RGB]{0,220,0} 0.716} & \textbf{\color[RGB]{0,220,0} 0.639} \\
UpdateNet~\cite{Zhang2019LearningTM} & 12.3 & 0.790 & 0.504 & 0.487 & 11.7 & 0.790 & 0.605 & 0.562 & 12.2 & 0.697 & 0.522 & 0.485 \\
\hline
SiamAPN~\cite{Fu2021SiameseAP} & 36.2 & 0.710 & 0.571 & 0.516 & 35.4 & 0.802 & 0.659 & 0.575 & 35.7 & \textbf{\color[RGB]{0,0,220} 0.815} & \textbf{\color[RGB]{0,0,220} 0.704} & 0.619 \\
SiamAPN++~\cite{Cao2021SiamAPNSA} & \textbf{\color[RGB]{0,0,220} 36.3} & 0.758 & 0.600 & 0.549 & \textbf{\color[RGB]{0,0,220} 35.6} & 0.738 & 0.612 & 0.535 & \textbf{\color[RGB]{0,0,220} 36.1} & 0.770 & 0.665 & 0.586 \\
SiamFC+\_CR~\cite{Zhang2019DeeperAW} & 30.5 & 0.649 & 0.481 & 0.427 & 30.4 & 0.754 & 0.625 & 0.570 & 30.2 & 0.728 & 0.603 & 0.502 \\
SiamRPN+\_CR~\cite{Zhang2019DeeperAW} & \textbf{\color[RGB]{0,220,0} 39.3} & 0.748 & 0.581 & 0.477 & \textbf{\color[RGB]{0,220,0} 38.2} & 0.742 & 0.594 & 0.497 & \textbf{\color[RGB]{0,220,0} 38.7} & 0.669 & 0.563 & 0.440 \\
SiamRPN++\_A~\cite{Li2019SiamRPNEO} & \textbf{\color[RGB]{220,0,0} 54.4} & 0.774 & 0.612 & 0.557 & \textbf{\color[RGB]{220,0,0} 52.6} & 0.804 & 0.651 & 0.579 & \textbf{\color[RGB]{220,0,0} 52.8} & 0.797 & 0.680 & 0.595 \\
\bottomrule
\end{tabular}
\label{tab:overall2}
\end{table*}

\begin{figure*}[ht]
\centering
\includegraphics[scale=0.13]{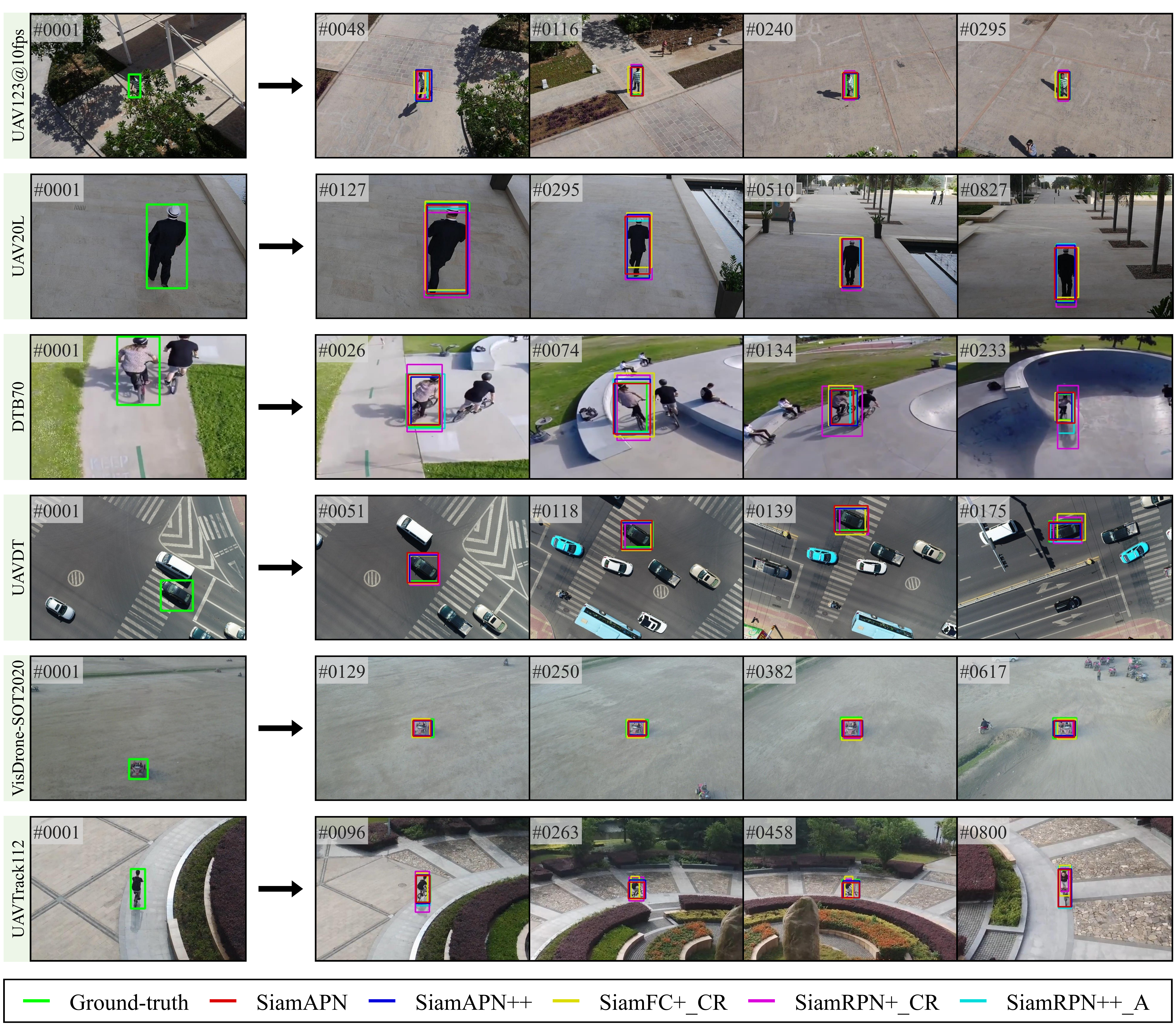}
\caption{Qualitative comparison of five real-time trackers.
The sequences and the corresponding benchmarks are (a) person13 at UAV123@10fps~\cite{Mueller2016ABA}, (b) person19 at UAV20L~\cite{Mueller2016ABA}, (c) BMX4 at DTB70~\cite{Li2017VisualOT}, (d) S0309 at UAVDT~\cite{Du2018TheUA}, (e) uav0000367\_04137\_s at VisDrone-SOT2020-test~\cite{Fan2020VisDroneSOT2020TV}, and (f) bike2 at UAVTrack112~\cite{Fu2021OnboardRA} (from the first row to the last).
The ground-truth bounding boxes and the trackers’ predicted boxes are colored differently.
After the same initialization, the closer the bounding box predicted by the tracker is to the ground-truth, the better the tracking result is.
}
\label{fig:success}
\end{figure*}

To more intuitively demonstrate the effectiveness of these five trackers, Fig.~\ref{fig:success} presents a series of tracking results on six benchmarks as a qualitative comparison.
As shown in Fig.~\ref{fig:success}, tracking results of different scenarios from different benchmarks are displayed.
These five real-time trackers all performed admirably.
Among them, the results of SiamAPN~\cite{Fu2021SiameseAP} and SiamAPN++~\cite{Cao2021SiamAPNSA} are closer to the bounding boxes of the ground-truth.

These five trackers run in real-time while maintaining high accuracy and robustness.
Therefore, they are qualified to further evaluate their performance in dealing with UAV challenges.

\subsection{Attribute-Based Evaluation}\label{sec:Attribute-Based Evaluation}
\begin{figure}[ht]
\centering
\includegraphics[scale=0.5]{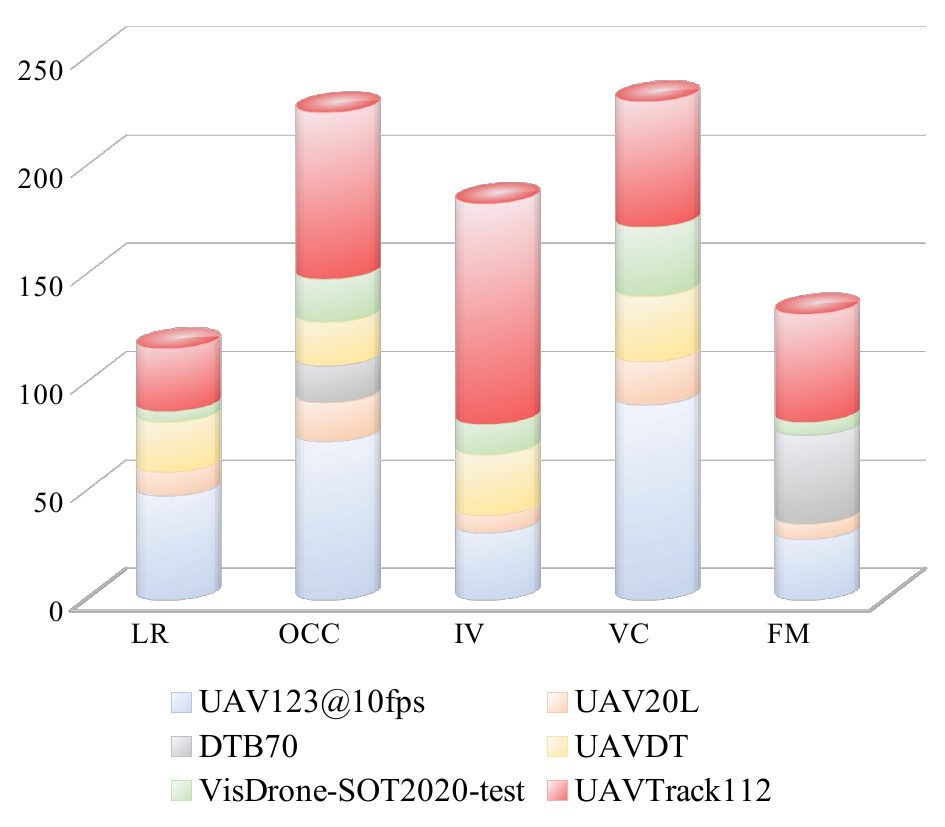}
\caption{The composition of the UAV challenge attributes from six benchmarks~\cite{Mueller2016ABA,Fan2020VisDroneSOT2020TV,Li2017VisualOT,Du2018TheUA,Fu2021OnboardRA}.
The number of attribute sequences for each benchmark is used as the basis for calculating the arithmetic average to analyze the tracker's performance for different UAV tracking challenges.
}
\label{fig:attributes}
\end{figure}

In addition to fulfilling the real-time requirements, it is equally essential to correctly meet the UAV tracking challenges in the deployment of UAV-based intelligent transportation systems.
To further analyze the feasibility and rationality of the tracker deployment on the UAV, this work evaluates the special attributes of the selected five trackers.
The challenges in different datasets are reclassified into the five UAV challenges mentioned above, \emph{i.e.}, LR, OCC, IV, VC, and FM.
Following~\cite{Fu2022CorrelationFF}, for UAV123@10fps~\cite{Mueller2016ABA}, UAV20L~\cite{Mueller2016ABA}, VisDrone-SOT2020-test~\cite{Fan2020VisDroneSOT2020TV}, and UAVTrack112~\cite{Fu2021OnboardRA}, the UAV challenge attributes LR, IV, and FM are the same as the original attributes low resolution, illumination variation, and fast motion; OCC consists of full occlusion (FOC) and partial occlusion (POC); VC consists of viewpoint change and camera motion (CM).
For DTB70~\cite{Li2017VisualOT}, the original attributes fast camera motion (FCM) is reclassified as VC, and occlusion is reclassified as OCC.
For UAVDT~\cite{Du2018TheUA}, the attributes IV, LR, OCC, and VC are taken from the original attributes of illumination variation, small objects (SO), large occlusion (LO), and camera rotation (CR).
Fig.~\ref{fig:attributes} shows the composition of the UAV challenge attributes from six benchmarks.

\begin{figure*}[ht]\scriptsize
\centering
\includegraphics[scale=0.5]{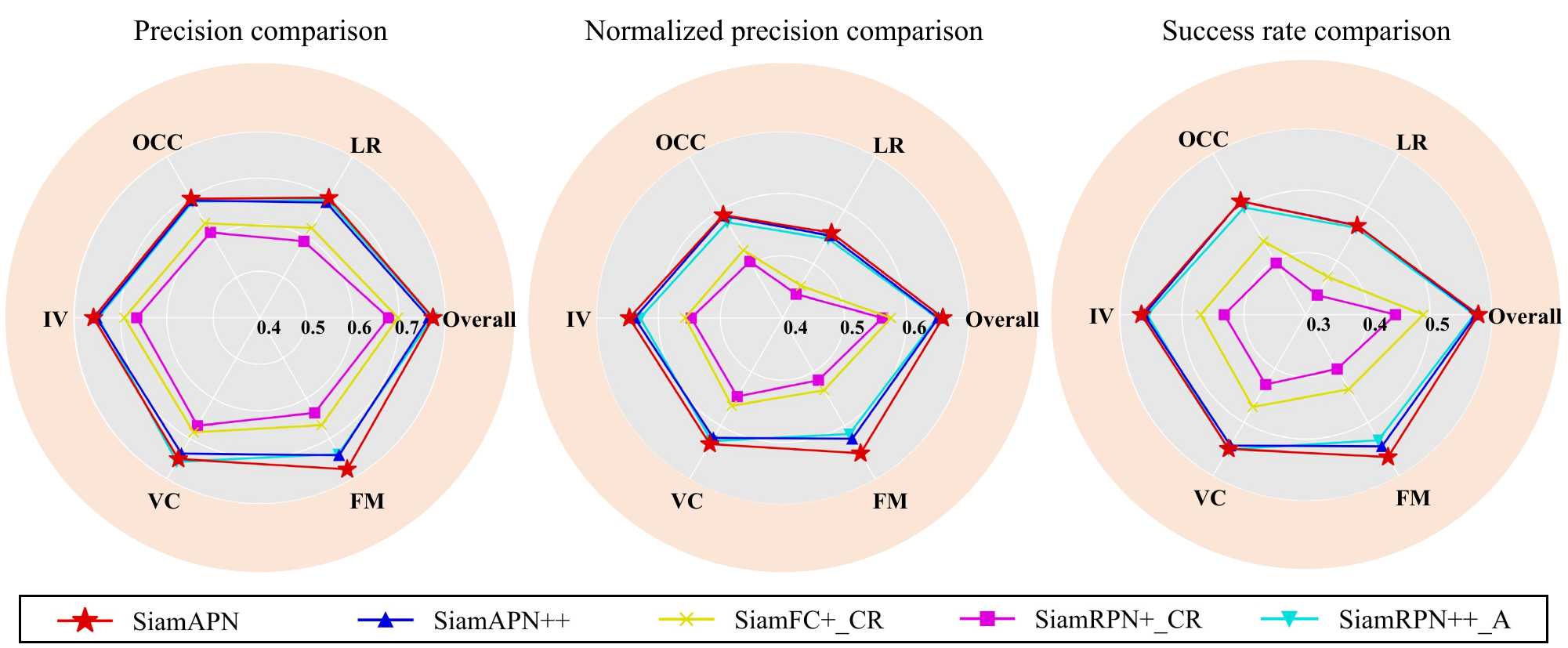}
\caption{The average performance comparison of the five real-time Siamese trackers under all the six authoritative UAV tracking benchmarks~\cite{Mueller2016ABA,Fan2020VisDroneSOT2020TV,Li2017VisualOT,Du2018TheUA,Fu2021OnboardRA} to demonstrate their ability under various attributes.
The specific values are in TABLEs~\ref{tab:precision},~\ref{tab:normalized precision}~and~\ref{tab:success rate}.
The two trackers in the outermost circle are SiamAPN~\cite{Fu2021SiameseAP} and SiamAPN++~\cite{Cao2021SiamAPNSA}, which have better performance to deal with UAV tracking challenges.
}
\label{fig:radar}
\end{figure*}

Based on the tracking results of each tracker under the UAV challenge attributes, the attribute performance comparison of the five real-time Siamese trackers under all the six UAV tracking benchmarks can be obtained in Fig.~\ref{fig:radar} and specific values are listed in TABLEs~\ref{tab:precision},~\ref{tab:normalized precision}~and~\ref{tab:success rate}.
Note that the final result is calculated by the arithmetic average of the results from the sequences containing the corresponding UAV challenge attributes in all six benchmarks.

As shown in Fig.~\ref{fig:radar}, the overall performance of trackers is slightly better than their performance under specific attributes, and each tracker has its specialties when it comes to coping with different UAV challenges.
Overall, the SiamAPN tracker~\cite{Fu2021SiameseAP} performs best, and the SiamAPN++ tracker~\cite{Cao2021SiamAPNSA} beats the SiamRPN++\_A tracker~\cite{Li2019SiamRPNEO} in normalized precision and success rate, although the SiamRPN++\_A tracker~\cite{Li2019SiamRPNEO} has some advantages in precision, but the SiamFC+\_CR tracker~\cite{Zhang2019DeeperAW} and the SiamRPN+\_CR tracker~\cite{Zhang2019DeeperAW} are relatively backward.

TABLEs~\ref{tab:precision},~\ref{tab:normalized precision}~and~\ref{tab:success rate} highlight the two best-performing trackers in addressing each UAV challenges.
Overall, the top two trackers that perform best are SiamAPN~\cite{Fu2021SiameseAP} and SiamAPN++~\cite{Cao2021SiamAPNSA}.
The SiamAPN tracker~\cite{Fu2021SiameseAP} has obvious advantages when it comes to FM and is marginally ahead when it comes to IV.
The SiamAPN++ tracker~\cite{Cao2021SiamAPNSA} is more balanced when dealing with various UAV challenges.
The SiamRPN++\_A tracker~\cite{Li2019SiamRPNEO} shows advantages in dealing with VC but is not good enough in OCC.
Furthermore, each tracker had evident flaws when it came to dealing with LR.

\begin{table*}[!hp]\scriptsize
\caption{The performance comparison of the five real-time Siamese trackers faced with specific UAV challenges.
This table records the average precision of the trackers when dealing with a certain type of challenge and the overall performance, corresponding 'Precision comparison' in Fig.~\ref{fig:radar}.
Note that the overall data is the average value of all the sequences in the six benchmarks.
The first two places are emphasized with \underline{underline}.
}
\centering
\renewcommand{\arraystretch}{1.3}
\newcommand{\tabincell}[2]{\begin{tabular}{@{}#1@{}}#2\end{tabular}}
\begin{tabular}{ p{3cm} p{1.5cm}<{\centering} p{1.5cm}<{\centering} p{1.5cm}<{\centering} p{1.5cm}<{\centering} p{1.5cm}<{\centering} p{1.5cm}<{\centering} }
\toprule
\textbf{Tracker} & \textbf{Overall} & \textbf{LR} & \textbf{OCC} & \textbf{IV} & \textbf{VC} & \textbf{FM} \\
\midrule
SiamAPN~\cite{Fu2021SiameseAP} & \underline{0.773} & \underline{0.697} & \underline{0.696} & \underline{0.758} & \underline{0.750} & \underline{0.776} \\
SiamAPN++~\cite{Cao2021SiamAPNSA} & 0.764 & 0.686 & \underline{0.691} & \underline{0.752} & 0.737 & \underline{0.741} \\
SiamFC+\_CR~\cite{Zhang2019DeeperAW} & 0.698 & 0.623 & 0.635 & 0.692 & 0.684 & 0.666 \\
SiamRPN+\_CR~\cite{Zhang2019DeeperAW} & 0.677 & 0.590 & 0.612 & 0.665 & 0.668 & 0.636 \\
SiamRPN++\_A~\cite{Li2019SiamRPNEO} & \underline{0.771} & \underline{0.692} & 0.688 & 0.748 & \underline{0.757} & 0.738 \\
\bottomrule
\end{tabular}
\label{tab:precision}
\end{table*}

\begin{table*}[!hbp]\scriptsize
\caption{The performance comparison of the five real-time trackers faced with specific UAV challenges.
This table records the average normalized precision of the trackers when dealing with a certain type of challenge and the overall performance, corresponding 'Normalized precision comparison' in Fig.~\ref{fig:radar}.
Note that the overall data is the average value of all the sequences in the six benchmarks.
The first two places are emphasized with \underline{underline}.
}
\centering
\renewcommand{\arraystretch}{1.3}
\newcommand{\tabincell}[2]{\begin{tabular}{@{}#1@{}}#2\end{tabular}}
\begin{tabular}{ p{3cm} p{1.5cm}<{\centering} p{1.5cm}<{\centering} p{1.5cm}<{\centering} p{1.5cm}<{\centering} p{1.5cm}<{\centering} p{1.5cm}<{\centering} }
\toprule
\textbf{Tracker} & \textbf{Overall} & \textbf{LR} & \textbf{OCC} & \textbf{IV} & \textbf{VC} & \textbf{FM} \\
\midrule
SiamAPN~\cite{Fu2021SiameseAP} & \underline{0.658} & \underline{0.558} & \underline{0.591} & \underline{0.647} & \underline{0.634} & \underline{0.651} \\
SiamAPN++~\cite{Cao2021SiamAPNSA} & \underline{0.651} & \underline{0.553} & \underline{0.589} & \underline{0.638} & 0.623 & \underline{0.625} \\
SiamFC+\_CR~\cite{Zhang2019DeeperAW} & 0.575 & 0.460 & 0.526 & 0.557 & 0.563 & 0.534 \\
SiamRPN+\_CR~\cite{Zhang2019DeeperAW} & 0.561 & 0.444 & 0.505 & 0.547 & 0.546 & 0.516 \\
SiamRPN++\_A~\cite{Li2019SiamRPNEO} & 0.649 & 0.547 & 0.578 & 0.629 & \underline{0.628} & 0.616 \\
\bottomrule
\end{tabular}
\label{tab:normalized precision}
\end{table*}

\begin{table*}[!hbp]
\caption{The performance comparison of the five real-time trackers faced with specific UAV challenges.
This table records the average success rate of the trackers when dealing with a certain type of challenge and the overall performance, corresponding 'Success rate comparison' in Fig.~\ref{fig:radar}.
Note that the overall data is the average value of all the sequences in the six benchmarks.
The first two places are emphasized with \underline{underline}.
}
\centering
\renewcommand{\arraystretch}{1.3}
\newcommand{\tabincell}[2]{\begin{tabular}{@{}#1@{}}#2\end{tabular}}
\begin{tabular}{ p{3cm} p{1.5cm}<{\centering} p{1.5cm}<{\centering} p{1.5cm}<{\centering} p{1.5cm}<{\centering} p{1.5cm}<{\centering} p{1.5cm}<{\centering} }
\toprule
\textbf{Tracker} & \textbf{Overall} & \textbf{LR} & \textbf{OCC} & \textbf{IV} & \textbf{VC} & \textbf{FM} \\
\midrule
SiamAPN~\cite{Fu2021SiameseAP} & \underline{0.578} & \underline{0.465} & \underline{0.511} & \underline{0.565} & \underline{0.549} & \underline{0.564} \\
SiamAPN++~\cite{Cao2021SiamAPNSA} & \underline{0.574} & \underline{0.466} & \underline{0.511} & \underline{0.561} & 0.543 & \underline{0.545} \\
SiamFC+\_CR~\cite{Zhang2019DeeperAW} & 0.489 & 0.370 & 0.436 & 0.470 & 0.471 & 0.438 \\
SiamRPN+\_CR~\cite{Zhang2019DeeperAW} & 0.444 & 0.336 & 0.396 & 0.432 & 0.430 & 0.401 \\
SiamRPN++\_A~\cite{Li2019SiamRPNEO} & 0.571 & 0.461 & 0.499 & 0.556 & \underline{0.550} & 0.533 \\
\bottomrule
\end{tabular}
\label{tab:success rate}
\end{table*}

\subsection{Failure Cases}\label{sec:Failure Cases}
\begin{figure}[ht]
\centering
\includegraphics[scale=0.11]{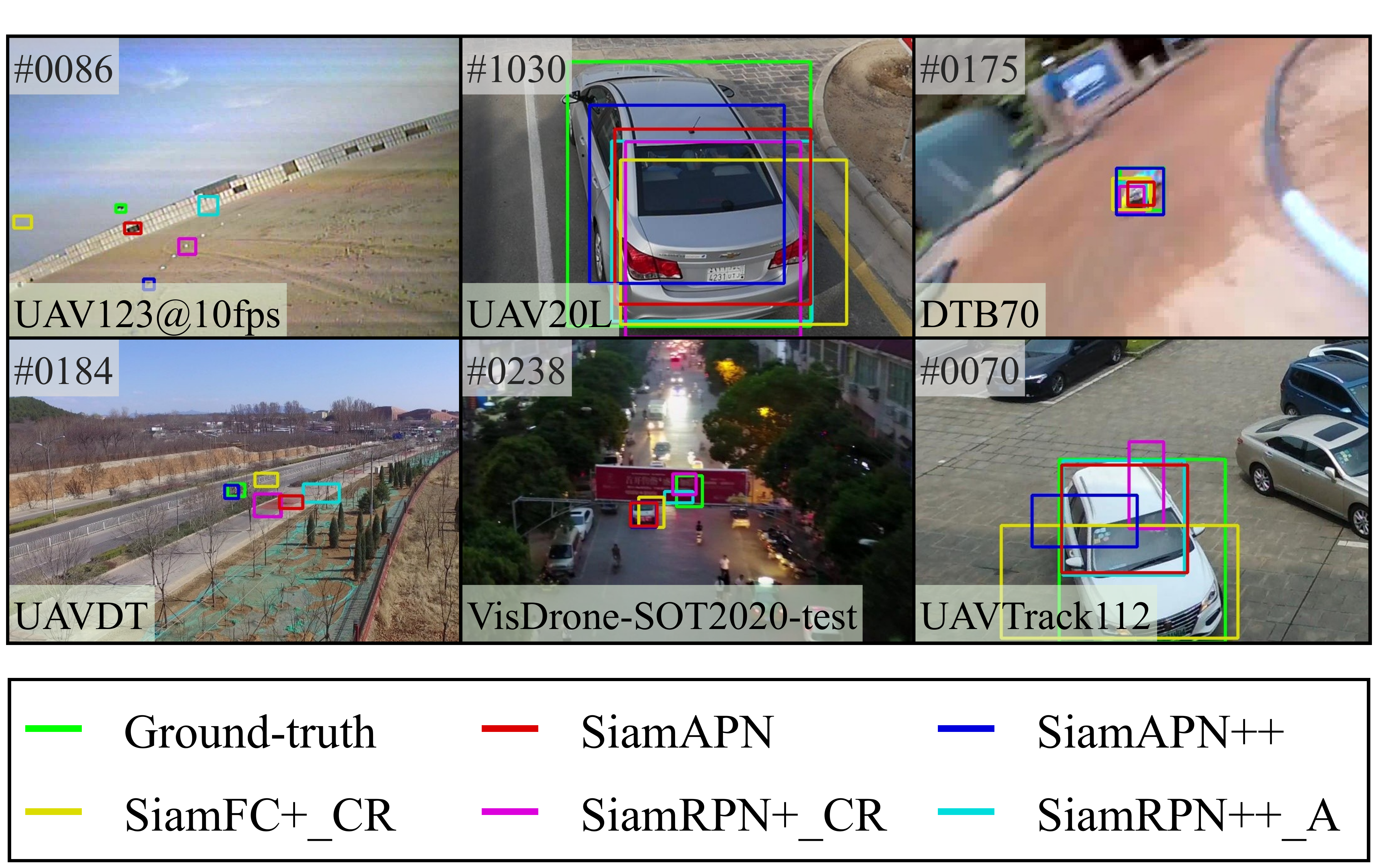}
\caption{Six representative tracking failure cases in each UAV tracking benchmarks.
The sequences and the corresponding benchmarks are (a) uav3 at UAV123@10fps~\cite{Mueller2016ABA}, (b) car6 at UAV20L~\cite{Mueller2016ABA}, (c) RcCar9 at DTB70~\cite{Li2017VisualOT}, (d) S1101 at UAVDT~\cite{Du2018TheUA}, (e) uav0000116\_00503\_s at VisDrone-SOT2020-test~\cite{Fan2020VisDroneSOT2020TV}, and (f) car16\_2 at UAVTrack112~\cite{Fu2021OnboardRA} (from left to right, top to bottom).
The ground-truth bounding boxes and the trackers’ predicted boxes are colored accordingly.
}
\label{fig:failure}
\end{figure}

Despite the fact the above experimental evaluations demonstrated the high performance of Siamese trackers, there are still hurdles to overcome.
This section investigates and analyzes several representative UAV tracking failure cases.
The performance of five trackers, \emph{i.e.}, SiamAPN~\cite{Fu2021SiameseAP}, SiamAPN++~\cite{Cao2021SiamAPNSA}, SiamFC+\_CR~\cite{Zhang2019DeeperAW}, SiamRPN+\_CR~\cite{Zhang2019DeeperAW}, and SiamRPN++\_A~\cite{Li2019SiamRPNEO}, on distinct sequences from each dataset is used to examine the current problems in UAV-based tracking, as shown in Fig.~\ref{fig:failure}.
\begin{enumerate}
\item{The tracker can be harmed by the low resolution of the sample, both for the target and the background. As shown in the first and third cases of Fig.~\ref{fig:failure}, inadequate sampling will have a direct impact on feature extraction, making it difficult for the Siamese trackers to differentiate between foreground and background effectively. Furthermore, small objectives and similar objectives will be harder to handle appropriately in this situation.}
\item{One of the issues that hampers UAV tracking is occlusion. Partial occlusion results in incomplete feature extraction of the target and the loss of useful feature information, and full occlusion means that the target vanishes from the field of view, putting the tracker's back-check performance to the test. As shown in the fourth and fifth cases of Fig.~\ref{fig:failure}, the presence of occlusion leads to incorrect tracking results.}
\item{The changes of viewpoint, as well as the relative motion between the camera and the object, cause deformation of the object, making it difficult for trackers to accurately identify the target. As shown in the second and sixth cases of Fig.~\ref{fig:failure}, accumulated deformation may lead to erroneous results.}
\end{enumerate}

\Remark Multiple UAV challenges frequently coexist in real-world aerial tracking applications, requiring trackers to employ more efficient and effective countermeasures.
Although existing trackers provide a favorable framework for UAV-based tracking, there are still issues that need to be addressed.

\subsection{Onboard Tests}\label{sec:Onboard Tests}

\begin{figure*}[ht]
\centering
\subfloat[]{\includegraphics[width=2.2in]{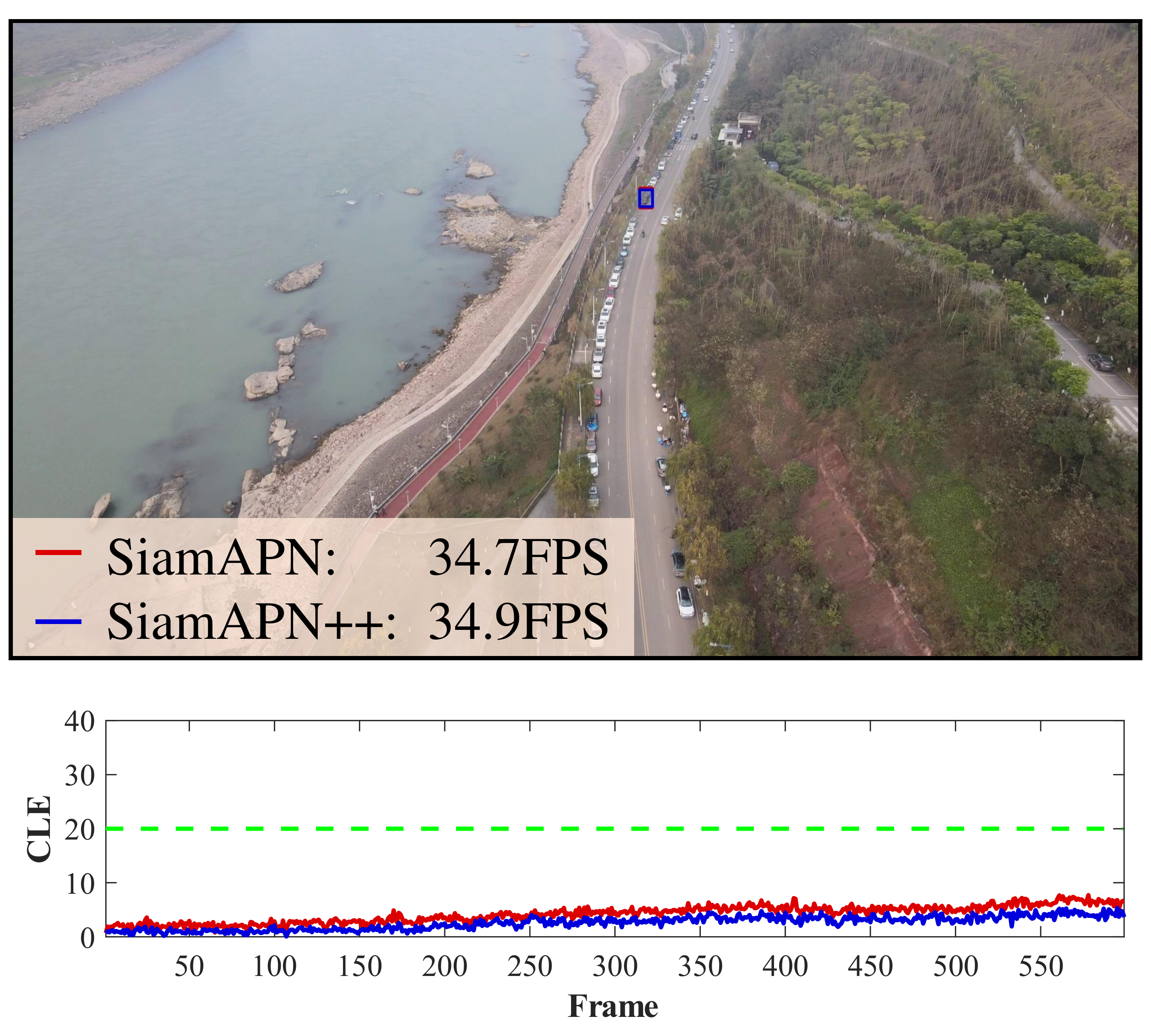}%
\label{fig:onboard(a)}}
\hfil
\subfloat[]{\includegraphics[width=2.2in]{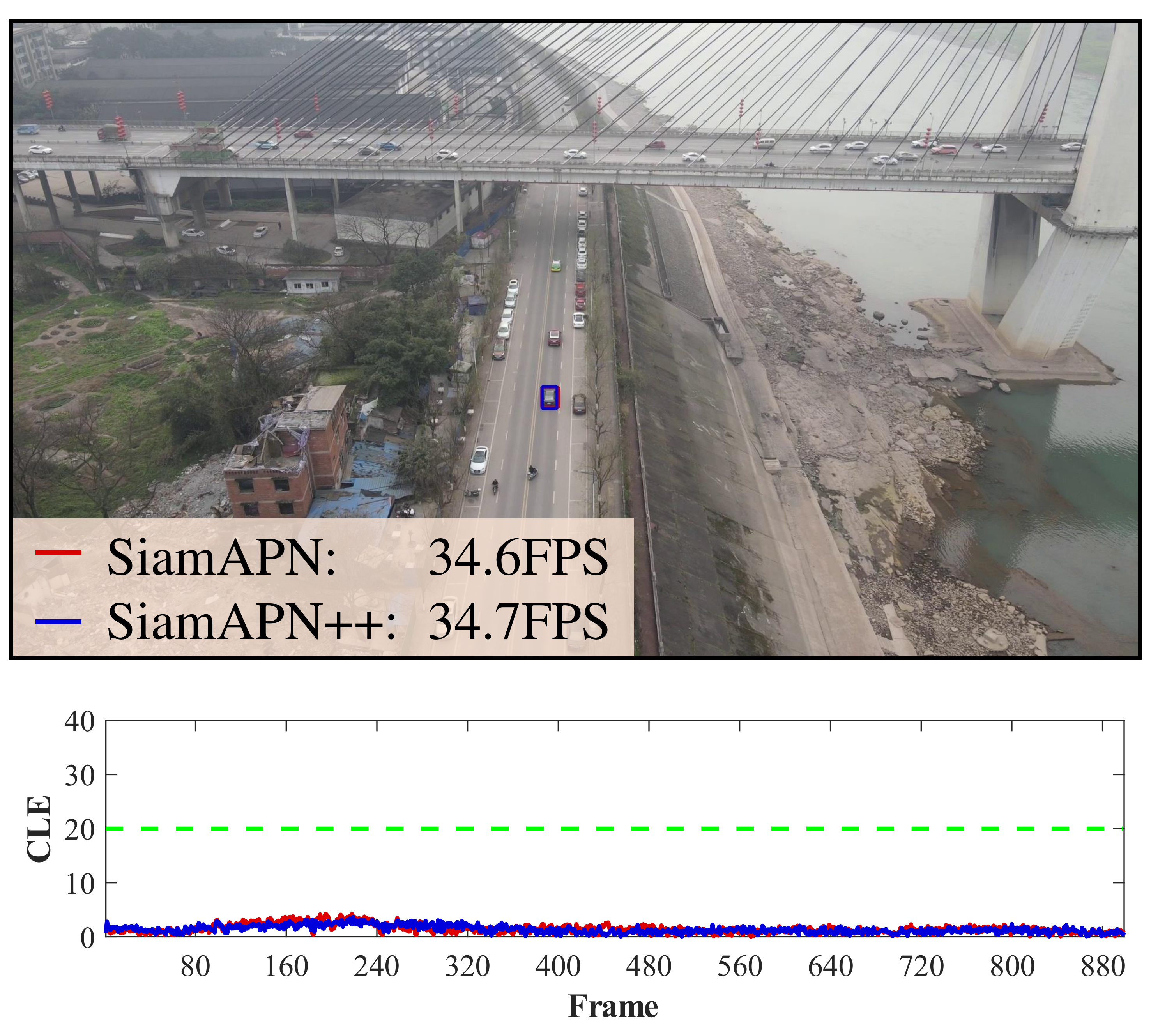}%
\label{fig:onboard(b)}}
\hfil
\subfloat[]{\includegraphics[width=2.2in]{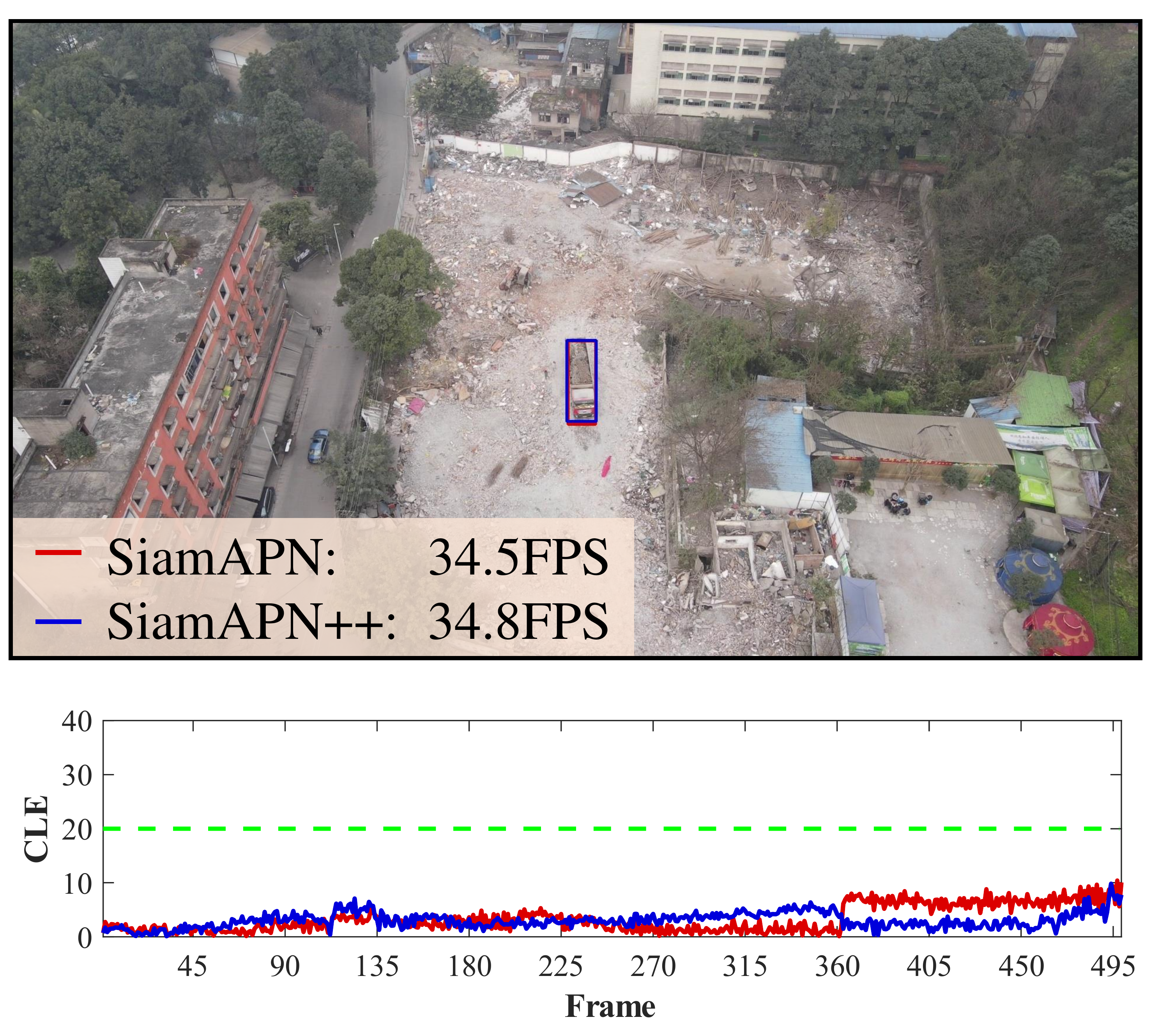}%
\label{fig:onboard(c)}}
\vfil
\subfloat[]{\includegraphics[width=2.2in]{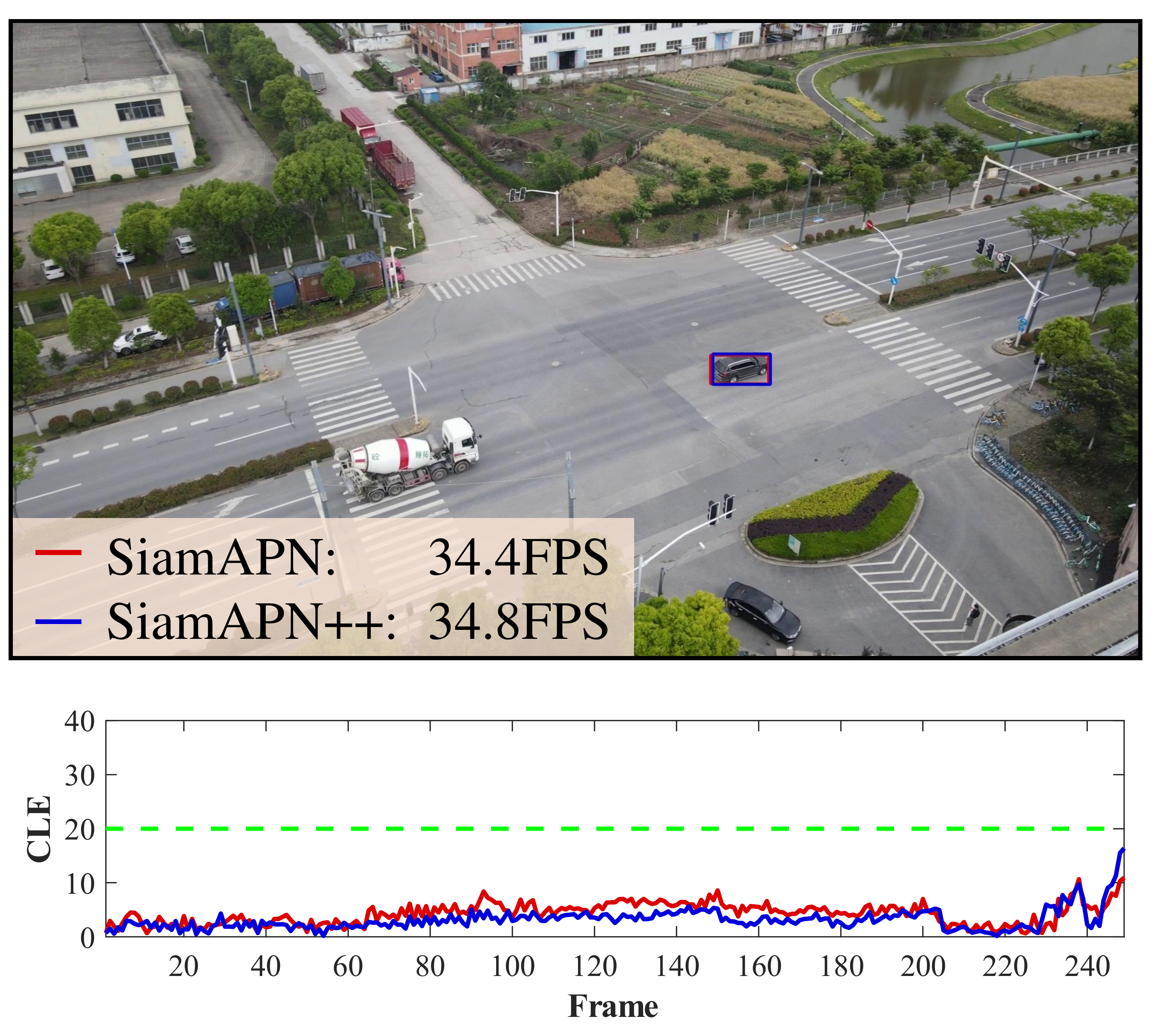}%
\label{fig:onboard(d)}}
\hfil
\subfloat[]{\includegraphics[width=2.2in]{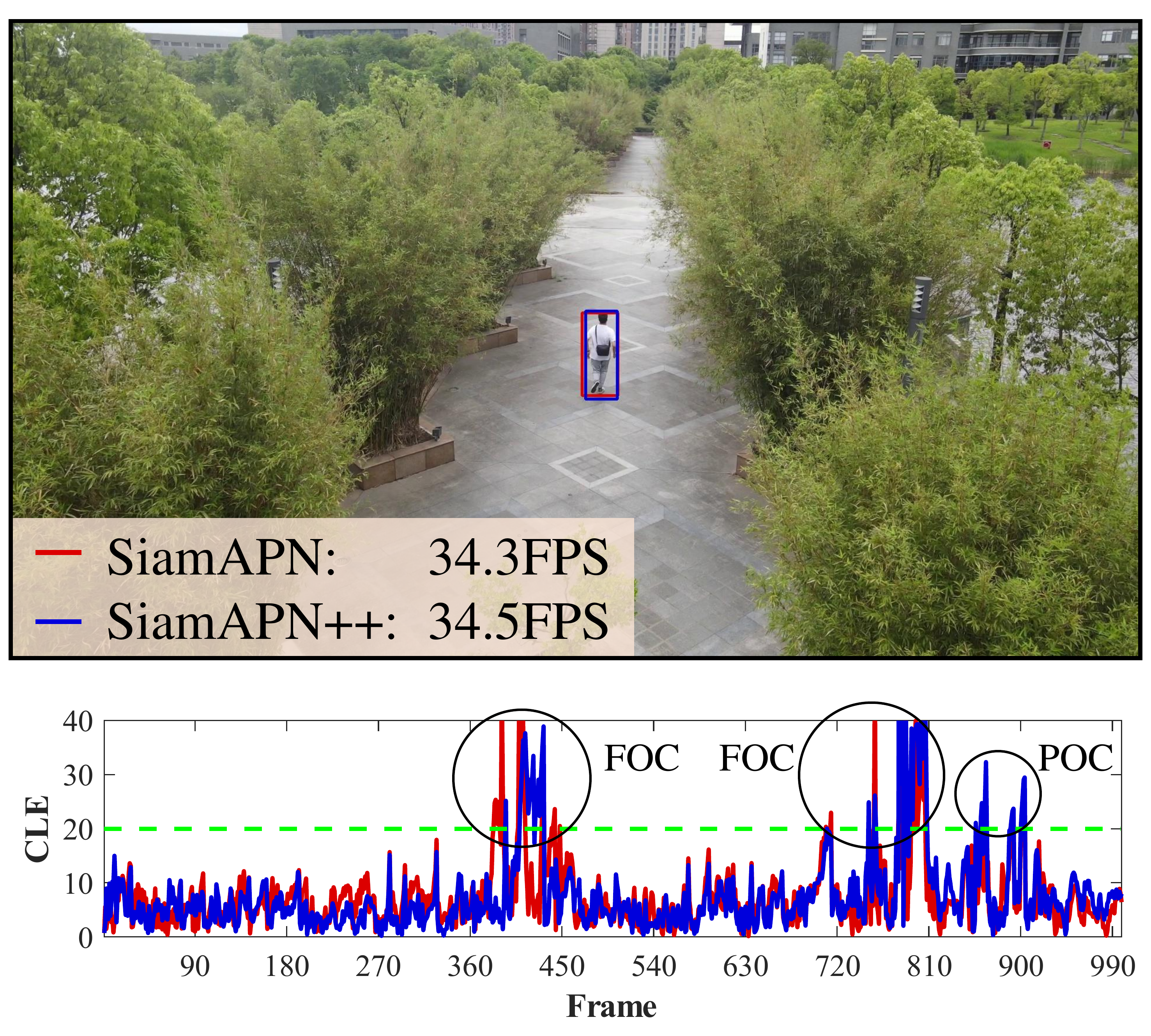}%
\label{fig:onboard(e)}}
\hfil
\subfloat[]{\includegraphics[width=2.2in]{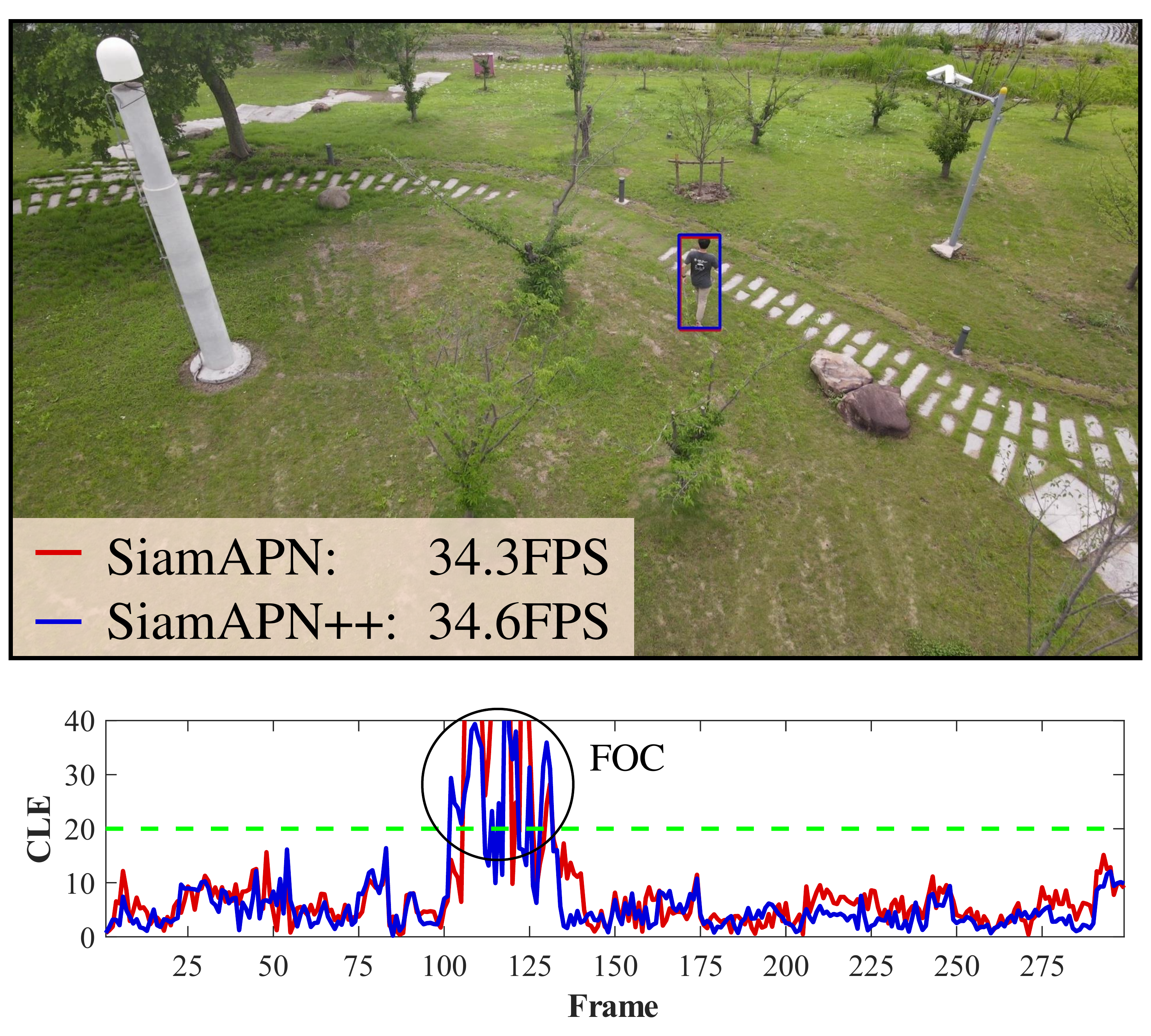}%
\label{fig:onboard(f)}}
\caption{The onboard tracking results and the performance in terms of the CLE.
The {\color[RGB]{220,0,0} red} lines denote the SiamAPN~\cite{Fu2021SiameseAP} tracker and the {\color[RGB]{0,0,220} blue} lines denote the SiamAPN++~\cite{Cao2021SiamAPNSA} tracker.
The {\color[RGB]{0,220,0} green} dotted lines represent the threshold of CLE = 20 pixels.
Typical occlusions appear in tests (e) and (f), where FOC denotes full occlusion and POC denotes partial occlusion.
With the great performance, SiamAPN~\cite{Fu2021SiameseAP} and SiamAPN++~\cite{Cao2021SiamAPNSA} demonstrate their capacity to handle the requirements of UAV tracking tasks.
}
\label{fig:onboard}
\end{figure*}

Besides the above large-scale experimental evaluation, this work also extended a series of onboard tests to further validate the real-time capabilities and robustness of the exceptional Siamese trackers~\cite{Fu2021SiameseAP,Cao2021SiamAPNSA}.
The onboard processor for these onboard tests is an NVIDIA Jetson AGX Xavier using MAXN nvpmodel.
After receiving images from the camera, a serial is used to communicate between the onboard processor and a Pixhawk 2.4.8\footnote{https://docs.px4.io/master/en/flight\_controller/pixhawk.html .}.
Besides, QGroundControl\footnote{http://qgroundcontrol.com/ .} serves as the ground station.

Onboard tracking performance of six tests with the SiamAPN~\cite{Fu2021SiameseAP} and SiamAPN++~\cite{Cao2021SiamAPNSA} trackers have been shown in Fig.~\ref{fig:onboard}.
The six tests, contain the five common UAV tracking challenges described above, \emph{i.e.}, LR, OCC, IV, VC, and FM.
The CLE curves of the two trackers throughout the tracking process of 6 tests are all recorded in Fig.~\ref{fig:onboard}, where they are all under a threshold of 20 pixels except when occlusion occurs in (e) and (f).
It can be seen that the appearance of occlusion has a significant impact on the performance of the trackers.
Nevertheless, the trackers can keep up with the target after the occlusion has passed.
In addition, the running speed of the two trackers in each test is also shown in Fig.~\ref{fig:onboard}, and all meet the real-time requirements of 30 FPS.

\Remark Both superior trackers are up to the challenge of UAV deployment in real-world tracking in terms of speed and performance.
Based on the balance of robustness and real-time performance, and the ability to confront various UAV challenges, SiamAPN~\cite{Fu2021SiameseAP} and SiamAPN++~\cite{Cao2021SiamAPNSA} show the suitability of combining with UAV-based tracking.

\subsection{Low-Illumination Analysis}\label{sec:Low-Illumination Analysis}
\begin{figure*}[!htp]
\centering
\includegraphics[scale=0.13]{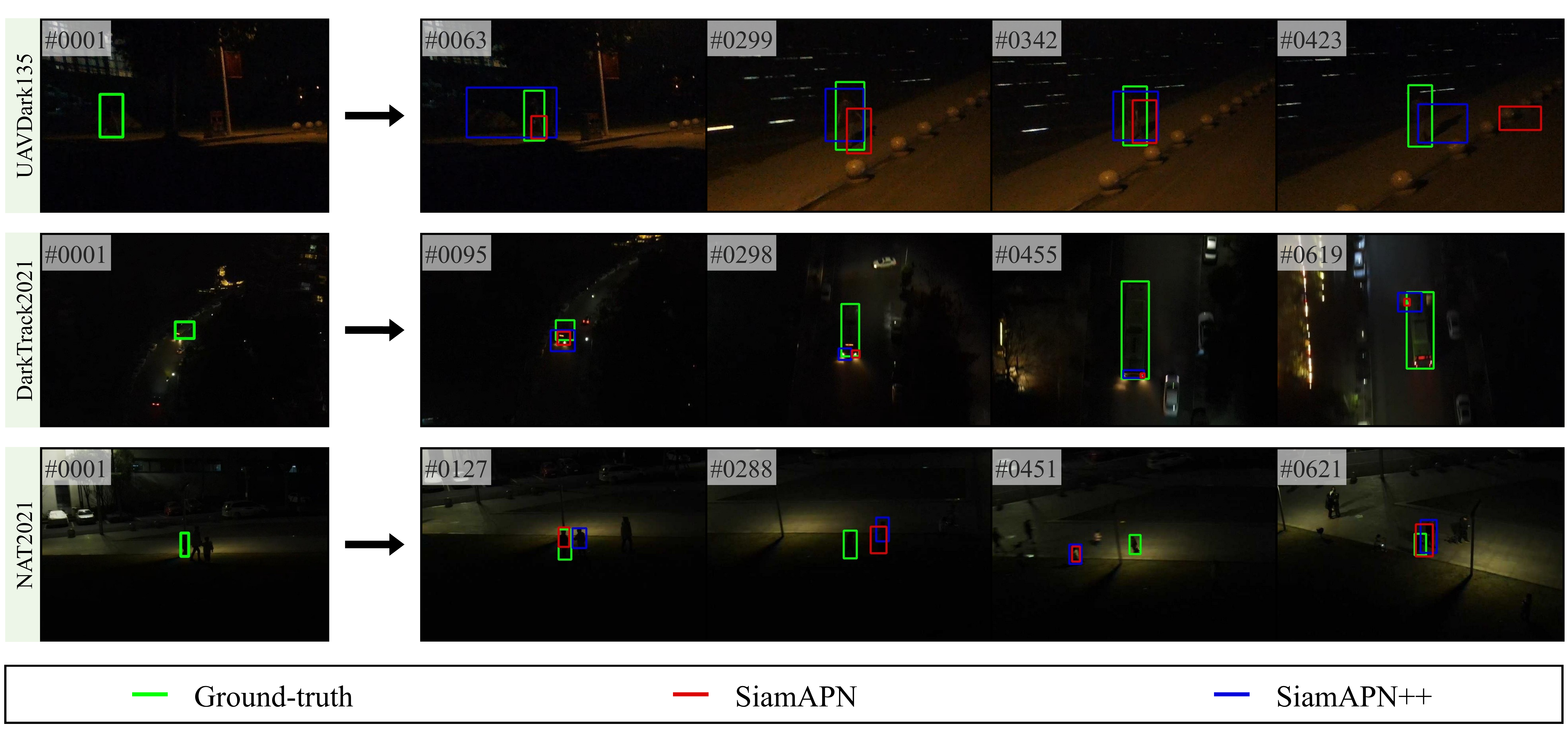}
\caption{The tracking results of two outstanding Siamese trackers in three low-illumination UAV benchmarks.
The sequences and the corresponding benchmarks are (a) person7 at UAVDark135~\cite{Li2022AllDayOT}, (b) bus\_6 at DarkTrack2021~\cite{Ye2022TrackerMN}, and (c) N04010 at NAT2021~\cite{Ye2022UnsupervisedDA} (from the first row to the last).
The ground-truth bounding boxes and the trackers’ predicted boxes are colored differently.
Even the most outstanding Siamese trackers~\cite{Fu2021SiameseAP,Cao2021SiamAPNSA} struggle to keep track of objects in adverse environments.
}
\label{fig:night}
\end{figure*}

An extra evaluation of tracking in the low light environment is conducted to further explore trackers' performance in more complex real-world UAV tracking scenarios.
low-illumination tracking situations, in addition to these basic issues mentioned above, are severe circumstances that wreak havoc on UAV-based tracking.
In order to do so, this work uses three low-illumination UAV benchmarks, \emph{i.e.}, UAVDark135~\cite{Li2022AllDayOT}, DarkTrack2021~\cite{Ye2022TrackerMN} and NAT2021~\cite{Ye2022UnsupervisedDA}, to test two outstanding Siamese trackers.
As shown in TABLE~\ref{tab:dark}, for the outstanding real-time trackers, \emph{i.e.}, SiamAPN~\cite{Fu2021SiameseAP} and SiamAPN++~\cite{Cao2021SiamAPNSA}, their performance all decline dramatically on the nighttime benchmarks when compared to the average performance under the six UAV tracking benchmarks~\cite{Mueller2016ABA,Li2017VisualOT,Du2018TheUA,Fan2020VisDroneSOT2020TV,Fu2021OnboardRA} dominated by the daylight environment.
As shown in Fig.~\ref{fig:night}, even the best UAV trackers can hardly overcome the challenges of low-illumination tracking.
The target cannot acquire enough light due to the loose and narrow distribution of light sources at night, and it is accompanied by shadows, which have a serious impact on trackers.

\Remark In low-illumination UAV tracking scenarios, other UAV challenges are often accompanied, \emph{e.g.}, illumination variation, occlusion, similar object, \emph{etc}.
Therefore, the indistinguishable foreground and background hinder the feature extraction ability of the network.
Moreover, since the sensor in the camera may introduce noise, the tracker will be disturbed, failing to track.

To have a more comprehensive application scenario, UAV-based tracking in low-illumination environments requires to be addressed.

\begin{table*}[hbp]
\caption{The performance comparison of the two real-time trackers under nine UAV benchmarks.
It includes the average performance on the six UAV tracking benchmarks and the performance on the three low-illumination UAV benchmarks respectively.
Obviously, the tracker's performance in low-illumination conditions is significantly lower than in daytime.
}
\centering
\renewcommand{\arraystretch}{1.3}
\newcommand{\tabincell}[2]{\begin{tabular}{@{}#1@{}}#2\end{tabular}}
\begin{tabular}{ l l p{2cm}<{\centering} p{3cm}<{\centering} p{3cm}<{\centering}}
\toprule
\multicolumn{2}{c}{\textbf{Benchmark}} & \textbf{Metric} & \textbf{SiamAPN}~\cite{Fu2021SiameseAP} & \textbf{SiamAPN++}~\cite{Cao2021SiamAPNSA} \\
\midrule
\multirow{3}{*}{Daytime} & \multirow{3}{*}{Average~\cite{Mueller2016ABA,Fan2020VisDroneSOT2020TV,Li2017VisualOT,Du2018TheUA,Fu2021OnboardRA}} & DP & 0.773 & 0.764 \\
 & & NDP & 0.658 & 0.650 \\
 & & AUC & 0.578 & 0.574 \\
\cline{2-5}
\hline
\rowcolor{gray!10}
& & DP & 0.426 & 0.431 \\
\rowcolor{gray!10}
& & NDP & 0.401 & 0.397 \\
\rowcolor{gray!10}
& \multirow{-3}{*}{UAVDark135~\cite{Li2022AllDayOT}} & AUC & 0.300 & 0.330 \\
\cline{2-5}
\rowcolor{gray!10}
 & & DP & 0.430 & 0.494 \\
\rowcolor{gray!10}
& & NDP & 0.389 & 0.446 \\
\rowcolor{gray!10}
& \multirow{-3}{*}{DarkTrack2021~\cite{Ye2022TrackerMN}} & AUC & 0.312 & 0.375 \\
\cline{2-5}
\rowcolor{gray!10}
 & & DP & 0.558 & 0.602 \\
\rowcolor{gray!10}
& & NDP & 0.418 & 0.481 \\
\rowcolor{gray!10}
\multirow{-9}{*}{Nighttime} & \multirow{-3}{*}{NAT2021~\cite{Ye2022UnsupervisedDA}} & AUC & 0.337 & 0.405 \\
\bottomrule
\end{tabular}
\label{tab:dark}
\end{table*}

\section{Prospect}\label{sec:Prospect}
In this section, this work discusses the prospect of future direction based on the current development of Siamese tracker and UAV, as well as the existing issues.
\begin{itemize}
\item{\textbf{Boost processing speed:}
Achieving real-time operating speed is a prerequisite for UAV-based tracking.
Although some trackers can meet the real-time requirements of UAV tracking, most Siamese trackers still face the dilemma of running speed.
It is imperative to boost the tracker's processing speed to solve the problem that even the finest trackers struggle to achieve real-time performance.
In terms of hardware, onboard equipment suited for UAV deployment should be further upgraded to surpass its performance.
From the perspective of algorithm development, the model's redundancy may be minimized without affecting accuracy, especially when using a deep backbone network.
Besides, various existing backbone networks provide a wide range of options for Siamese tracking.
For example, ShuffleNet~\cite{Zhang2018ShufflenetAE,Ma2018ShufflenetVP} developed for mobile devices may be able to meet the deployment of UAV with its high efficiency.
In addition, with the help of a software development kit (SDK), such as TensorRT\footnote{https://developer.nvidia.com/tensorrt .}, the computing efficiency of the network can also be improved to some extent.
}
\item{\textbf{Improve aerial tracking performance:}
Although the existing Siamese tracker has some performance advantages, there is still potential to meet the diverse deployment of UAV-based intelligent transportation systems.
The accuracy and success rate are still inadequate in the face of some challenging UAV circumstances and long-term tracking.
Making full use of the spatial, temporal, and multi-scale information of the object when designing trackers will help to deal with difficult UAV challenges.
Introducing a suitable update strategy makes sense for tracking, especially real-world UAV tracking.
For online tracking, an acceptable and effective updating method is still required.
At the same time, the design of the update mechanism should adapt to the UAV tracking task as efficiently as possible.
With the advent of Transformer~\cite{Vaswani2017AttentionIA}, vision Transformer~\cite{Vaswani2017AttentionIA,Dosovitskiy2021AnII} is gaining popularity in computer vision, combining it with Siamese trackers can be an noteworthy idea.
Due to the ability to extract global information and the utilization of the attention mechanism, the Transformer has achieved success in object detection~\cite{Carion2020EndTE,Dai2021UpDU} and segmentation~\cite{Wang2021MaxDE,Wang2021EndTE}.
In addition, there are already researches~\cite{Cao2021HiftHF,Yan2021LearningST,Chen2021TransformerT,Cao2022TCTrackTC} using Transformer for visual tracking.
Combining the advantages of Transformer and CNN, it may be able to achieve the purpose of improving tracking performance.
The purpose of developing an appropriate UAV tracker is to strike a decent balance between performance and running speed.
Attempting to overcome the UAV tracking challenge on the basis of obtaining and maintaining real-time performance is the orientation that Siamese UAV tracking should strive for in the future.
}
\item{\textbf{Cope with adverse environments:}
In a real-world intelligent transportation application, UAV-based tracking frequently encounters a variety of adverse circumstances.
However, current UAV tracking methods are typically trained using data acquired in normal conditions, which may be insufficient for harsh circumstances in UAV working situations, \emph{e.g.}, low-illumination environments, adverse weather, \emph{etc}.
In low-illumination scenarios, for example, the acquired images have low contrast, brightness, and signal-to-noise ratio, resulting in a significant domain discrepancy between the normal lighting and the low-illumination environment.
Such an issue limits the further extensive application of UAV tracking.
To enrich the training stage, one intuitive solution is to collect a huge number of harsh environment images, but this contains security risks and requires a lot of effort, time, and money for both collecting and labeling.
On the other hand, unsupervised domain adaptation strategies~\cite{Ye2022UnsupervisedDA}, can improve the domain discrepancy with unlabeled images.
This provides a new promising solution for UAV-based tracking in adverse environments represented by low-illumination.
}
\item{\textbf{Raise safety and robustness:}
Due to the vulnerability of CNN, adding perturbations in the image can cause the deep model to output a wrong prediction with high confidence, which is called an adversarial attack.
This characteristic can lead to potential malicious manipulation and pose significant dangers.
As a vision technology with broad application prospects, UAV tracking based on Siamese networks deserves our attention in particular for its safety and robustness.
Recently, there are some works~\cite{Chen2020OneSA,Yan2020CoolingSA} focusing on the adversarial attack for visual object tracking tasks.
\cite{Fu2022Ad2AttackAA} proposes an adaptive attack method based on image-resampling, especially for UAV-based tracking, considering the low computation capability of UAVs.
It can be seen that Siamese trackers on UAVs may collapse and lose the target in the following frames under the attack of the imperceptible perturbations.
The improvement of adversarial robustness can be obtained through the reasonable design of the structure and appropriate training strategies.
From the design of the tracker’s architecture, paying more attention to the low-frequency component and unrobust feature of the image can make the deep tracker insensitive to the adversarial attack.
Moreover, from the training and fine-tune phase, introducing adversarial examples can make targeted robust improvements to the model.
As UAV tracking continues to expand its range of applications, the adversarial robustness of the CNN-based tracker will be a significant evaluation metric besides the accuracy, especially in safety-critical scenarios.
}
\item{\textbf{Compensate tracking latency:}
In a real-world setting, the onboard latency of the trackers will inevitably lead to the mismatch between the estimated object state and its real state~\cite{Li2020PredictiveVT,Li2020TowardsSP}.
Even for the real-time trackers~\cite{Fu2021SiameseAP,Cao2021SiamAPNSA}, such latency can result in significant performance descent.
Specifically, since the tracker can only process the latest frame, there may exist skipped frames for slow trackers.
Besides, the output of the tracker will always lag behind the real object state due to latency.
One possible solution is to integrate predictor into the original tracking pipeline, termed predictive visual tracking (PVT)~\cite{Li2020PredictiveVT}.
Given the object's history motion trajectory and appearance change, a predictive tracker can forecast its future state, which is thus free of latency and more suitable for real-world onboard tracking.
\cite{Li2020PredictiveVT} proposes a baseline approach for PVT, introducing a pre-forecaster and a post-forecaster based on Kalman Filter.
Learning based PVT as well as joint training of predictor and tracker can be expected to improve the tracking effect.
}
\item{\textbf{Enrich the evaluation for UAV-based tracking:}
Despite the existence of various UAV datasets, there are still very few UAV tracking datasets for evaluation, particularly UAV tracking challenge-specific ones.
Expanding the distribution of datasets and the variety of the UAV tracking challenge will facilitate the real-world deployment of UAV-based tracking.
High-quality adverse environment datasets can effectively promote UAV-based tracking to adapt to real-world conditions.
Taking the low-illumination environment as an example, enhancing the data distribution under variable lighting will further boost research on trackers in low-illumination environments to counteract the negative effects of illumination variance and insufficient lighting.
Besides, more effective evaluation metrics are beneficial in measuring the tracker's performance more intuitively and effectively.
According to the particularity of UAV-based tracking, more specialized evaluation approaches can be adopted to more precisely analyze and measure the tracker's performance.
In addition to fundamental performance measurements such as accuracy and success rate, it is also required to analyze the real-time performance of UAVs, such as an evaluation benchmark for real-time UAV tracking proposed by ~\cite{Li2020PredictiveVT}.
Furthermore, evaluation metrics for specific UAV challenges should be proposed to focus on the performance of trackers on UAV tracking challenges.
}
\item{\textbf{Broaden application fields:}
The combination of Siamese networks and UAV tracking has been realized, and the widespread application of UAVs has spawned plenty of intelligent transportation systems.
Among them, the unmanned aerial manipulator (UAM), which is a kind of UAV-based aerial robot, can execute operations utilizing an onboard manipulator and specialized devices.
The UAM exhibits remarkable advantages in high efficiency, stability, cost-effectiveness, and safety in aerial work by physically interacting with the environment, which greatly decreases labor costs and dangers in related tasks.
Even though Siamese tracking for general UAVs has gotten a lot of attention and research, Siamese tracking for UAMs still requires to be improved.
Because the aerial interaction of the UAMs is heavily reliant on the object's visual perception, especially when the UAMs progress to the third generation~\cite{Ollero2021PastPA}, which relies heavily on free visual perception to reach a high level of automation.
Siamese trackers for UAV-derived intelligent flight platforms like UAMs also confront new obstacles, such as more severe scale variation~\cite{Zheng2022SiameseOT}.
In the future, many applications in industrialization and application markets are promising to benefit from more effective Siamese tracking for aerial intelligent transportation systems, \emph{e.g.}, life exploration, aerial maintenance of aeronautical equipment, adverse environment exploration, automated logistics and transportation, emergency rescue and disaster relief, \emph{etc}.
}
\end{itemize}

\section{Conclusions}\label{sec:Conclusions}
Focusing on the deployment of UAV-based aerial tracking for plentiful intelligent transportation applications, this work comprehensively reviews the fundamental structure and concept of leading-edge Siamese trackers.
These Siamese trackers are classified into categories based on their primary innovations and contributions.
Furthermore, based on six UAV tracking benchmarks, an exhaustive experimental evaluation is conducted on the typical UAV onboard processor to qualitatively and quantitatively compare SOTA Siamese trackers for UAV-based intelligent transportation systems.
Moreover, their performances, such as speed, accuracy, and success rate, are examined, as well as the performance of the trackers in response to UAV-specific challenges.
On this basis, the possibility of Siamese trackers' actual combination with UAVs is explored with onboard tests to verify the rationality and feasibility.
This work also examines the constraints that Siamese trackers and UAV-based tracking confront according to the current difficulties and challenges.
Furthermore, this work further explores the performance of SOTA trackers in low-illumination UAV tracking conditions to arose further research.
Finally, this work presents a perspective, discussing the direction of future development based on the existing challenges of UAV tracking and Siamese networks.
In light of the remarkable impact and promising future of Siamese UAV tracking, this work attempts to summarize the current status from both theoretical and practical perspectives to benefit the UAV-based intelligent transportation community.
$\hfill\blacksquare$ 

\section*{Acknowledgement}
This work is supported by the National Natural Science Foundation of China (No. 62173249) and the Natural Science Foundation of Shanghai (No. 20ZR1460100).

\bibliographystyle{IEEETranS.bst}

\bibliography{SNOT}

\end{document}